\documentclass[sigconf]{acmart}
\AtBeginDocument{%
  \providecommand\BibTeX{{%
    \normalfont B\kern-0.5em{\scshape i\kern-0.25em b}\kern-0.8em\TeX}}}

\usepackage{booktabs}
\usepackage{float}
\usepackage{multirow}
\usepackage{textcomp}
\usepackage{subfigure} 
\usepackage{caption}
\usepackage{subcaption}
\usepackage{bm}
\usepackage{appendix}
\usepackage{algorithm}
\usepackage{algorithmic}
\usepackage{makecell}
\usepackage{url}
\usepackage[labelformat=simple]{subcaption}
\usepackage{CJKutf8}
\usepackage{latexsym}
\usepackage{amsfonts}
\usepackage{amsmath}
\usepackage{multicol}
\usepackage{multirow}
\usepackage{xspace}
\usepackage{booktabs}
\usepackage{array}
\usepackage{threeparttable}
\usepackage{tcolorbox}
\usepackage{tabularx}
\usepackage{enumitem}
\usepackage{xcolor,colortbl}
\usepackage{color}
\usepackage{setspace}
\usepackage{makecell}
\usepackage{listings}
\usepackage{xltabular}
\usepackage{tcolorbox}
\tcbuselibrary{breakable}
\usepackage{xcolor}
\usepackage{arydshln}
\definecolor{deepred}{RGB}{160, 0, 0}
\usepackage{marvosym}
\usepackage{fontawesome}
\hypersetup{
    colorlinks=true,
    linkcolor=blue,
    citecolor=blue,
}

\copyrightyear{2026}
\acmYear{2026}
\setcopyright{acmlicensed}\acmConference[WWW '26]{Proceedings of the ACM Web Conference 2026}{April 13--17, 2026}{Dubai, UAE}
\acmBooktitle{Proceedings of the ACM Web Conference 2026 (WWW '26), April 13--17, 2026, Dubai, UAE}
\acmDOI{xxxxxxx.xxxxxxx}

\begin{document}
\title{Toward Generalized Web Agent Training: A Deep Dive into Entropy-Balanced Reinforcement Learning}

\title{Agentic Entropy-Balanced Policy Optimization}

 \author{
 \large
 \textbf{Guanting Dong}\textsuperscript{1}\textsuperscript{$\mathsection$},
\textbf{Licheng Bao}\textsuperscript{2}\textsuperscript{$\mathsection$},
\textbf{Zhongyuan Wang}\textsuperscript{2}\textsuperscript{$\mathsection$},
\textbf{Kangzhi Zhao}\textsuperscript{2},
\textbf{Xiaoxi Li}\textsuperscript{1},
\textbf{Jiajie Jin}\textsuperscript{1},
\textbf{Jinghan Yang}\textsuperscript{2}\textsuperscript{$\mathsection$},}

\author{
 \large
\textbf{Hangyu Mao}\textsuperscript{2},
\textbf{Fuzheng Zhang}\textsuperscript{2},
\textbf{Kun Gai}\textsuperscript{2},
\textbf{Guorui Zhou}\textsuperscript{2}$^{\textrm{\Letter}}$,
\textbf{Yutao Zhu}\textsuperscript{1},
\textbf{Ji-Rong Wen}\textsuperscript{1},
\textbf{Zhicheng Dou}\textsuperscript{1}$^{\textrm{\Letter}}$
}

\affiliation{%
\large
  \institution{\textsuperscript{1}Renmin University of China \ \ \textsuperscript{2}Kuaishou Technology}
  \city{}
  \country{}
}
\affiliation{%
\large
  \institution{\href{mailto:dongguanting@ruc.edu.cn}{\{dongguanting}, \href{mailto:dou@ruc.edu.cn}{dou\}@ruc.edu.cn}}
  \city{}
  \country{}
}

\thanks{$\mathsection$ Work done during internship at Kuaishou. }
\thanks{${\textrm{\Letter}}$ Corresponding author.}

\affiliation{
\institution{
\large
\faGithub\ GitHub: \href{https://github.com/dongguanting/ARPO}{\texttt{\textcolor{cyan}{https://github.com/dongguanting/ARPO}}}}
  \country{}
}

\renewcommand{\shortauthors}{Dong et al.}


\begin{abstract}
  Recently, Agentic Reinforcement Learning (Agentic RL) has made significant progress in incentivizing the multi-turn, long-horizon tool-use capabilities of web agents. While mainstream agentic RL algorithms autonomously explore high-uncertainty tool-call steps under the guidance of entropy, excessive reliance on entropy signals can impose further constraints, leading to the training collapse.
  In this paper, we delve into the challenges caused by entropy and propose the Agentic Entropy-Balanced Policy Optimization (AEPO), an agentic RL algorithm designed to balance entropy in both the rollout and policy update phases. AEPO comprises two core components: \textbf{(1)} a dynamic entropy-balanced rollout mechanism that adaptively allocate global and branch sampling budget through entropy pre-monitoring, while imposing a branch penalty on consecutive high-entropy tool-call steps to prevent over-branching issues; and \textbf{(2)} Entropy-Balanced Policy Optimization that inserts a stop-gradient operation into the high-entropy clipping term to preserve and properly rescale gradients on high-entropy tokens, while incorporating entropy-aware advantage estimation to prioritize learning on high-uncertainty tokens. Results across 14 challenging datasets show that AEPO consistently outperforms 7 mainstream RL algorithms. 
  With just $1K$ RL samples, Qwen3-14B with AEPO achieves impressive results: \textbf{47.6\% on GAIA, 11.2\% on Humanity’s Last Exam, and 43.0\% on WebWalkerQA for Pass@1; 65.0\% on GAIA, 26.0\% on Humanity’s Last Exam, and 70.0\% on WebWalkerQA for Pass@5.} Further analysis reveals that AEPO improves rollout sampling diversity while maintaining stable policy entropy, facilitating scalable web agent training. 
\end{abstract}

\keywords{Agentic Reinforcement Learning, Agentic Search, Web Agent, Tool Learning, Large Language Model}

\maketitle

\section{Introduction}
The emergence of large language models (LLMs) have profoundly revolutionized a wide range of  natural language reasoning tasks~\citep{2409_openai_o1,deepseek-r1,team2025kimi,qwen3,MiniMax-M1,GLM-4.5,team2025kimi2,team2025longcat}. Despite their impressive capabilities, the static nature of their internal knowledge often leads LLMs to experience hallucinations and information staleness in knowledge-intensive scenarios~\citep{zhang2023sirens}. Retrieval-augmented generation (RAG) addresses these limitations by empowering LLMs to reason with retrieved relevant knowledge, thereby improving the reliability of generated answers~\citep{guu2020realm,lewis2021retrievalaugmented,2411_htmlrag,dparag,retrollm,flashrag,vifrag}. However, with the explosive growth of web information, the static RAG workflow limits effective interaction between LLMs and search engines, resulting in significant bottlenecks in open-domain web exploration. To overcome these challenges, a series of LLM-based web agents have emerged~\citep{searcho1,webthinker,WebEvolver}. These agents perform on-demand web searches during reasoning and strategically interact with external tool environments, achieving reliable, in-depth web information seeking~\citep{searchr1,r1_searcher++,chen2025research,li2025websailor,Web-CogReasoner,2412_AR_MCTS,zhou2025agentfly}.

\begin{figure}
    \centering
    \includegraphics[width=\linewidth]{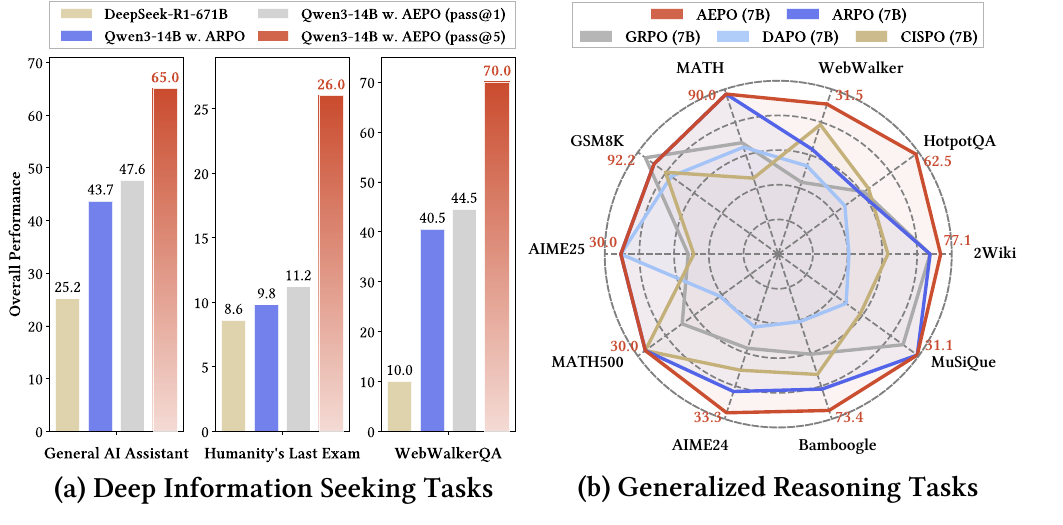}
    \vspace{-2.5em}
    \caption{Performance overview of AEPO algorithm.}
     \vspace{1.5em}
    \label{fig:intro_performance}
\end{figure}

\begin{figure*}[!t]
\centering
\setlength{\abovecaptionskip}{0.05cm}
\setlength{\belowcaptionskip}{0.05cm}
\includegraphics[width=0.985\linewidth]{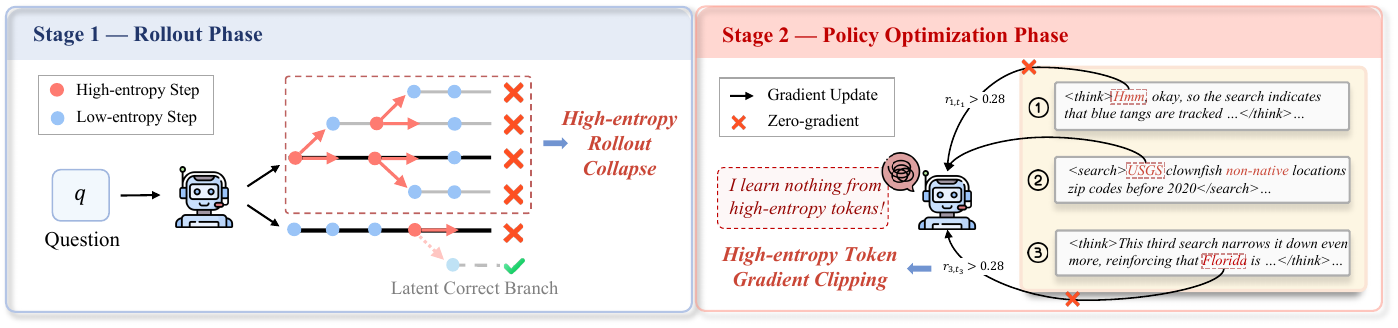}
\caption{Two high-entropy challenges in agentic RL. \textit{(1) High-Entropy Rollout Collapse}: Over-branching at high-entropy steps along specific paths, limiting exploration of other potential correct branches; \textit{(2) High-Entropy Token Gradient Clipping:}  Consistent clipping of high-entropy token gradients during policy updates hinders learning effective exploration behaviors.}
\label{fig:intro} 
\end{figure*}

To strive for efficient training of web agents, early implementations focus on distill tool-use trajectories from stronger models and guide weaker models through supervised fine-tuning (SFT)~\citep{tora,dotamath,siam,CognitiveKernelPro}. However, relying solely on SFT struggles to discover autonomous and generalizable tool-use capability~\citep{sft_memory}. As large-scale reinforcement learning with verifiable rewards (RLVR) demonstrates the potential to unlock frontier LLM capabilities~\citep{deepseek-r1,qwq-32b-preview,team2025kimi2}, several web-search agents adopt trajectory-level RL~\citep{deepseekmath,DAPO,GSPO} combined with carefully designed reward functions to elicit agentic reasoning in LLMs~\citep{wu2025webdancer,dong2025toolstar}. While effective to some extent, this line of work consistently overlooks the multi-turn interactive nature between LLMs and tool environments~\citep{zhang2025landscape}, making it difficult to discover step-level tool-use behaviors during RL training. 
To mitigate this limitation, recent efforts in agentic RL have shown that web agents often display high entropy in their output tokens due to uncertainty about the external tool-call results~\citep{dong2025arpo}. Drawing on this finding, they introduce a tree-structured rollout method that adaptively branches at high-entropy tool-call steps, effectively broadening sampling diversity and coverage~\citep{Group-in-Group,ji2025treesearchllmagent,liu2025ettrl,li2025treepo}.

Although these entropy-driven agentic RL algorithms stimulate exploration of latent tool-use behaviors, such high-entropy signals further pose two extra challenges for web agent training:
\begin{enumerate}[leftmargin=1em]
\item {\textbf{High-entropy Rollput Collapse:}} During the rollout phase, high-entropy tool-call steps often occur consecutively, leading the LLM to over-branch along a single trajectory under high-entropy guidance. This situation depletes the branching budget for other trajectories at high-entropy steps, ultimately limiting the diversity and scope of rollout sampling (see Figure~\ref{fig:intro} (left)).

\item {\textbf{High-entropy Token Gradient Clipping:}} The tree-structured rollout strategy encourages LLMs to explore step-level tool-use behaviors, thus preserving valuable high-entropy tokens.  However, vanilla RL algorithms aggressively clip high-entropy token's gradient during policy update phase, leading to the premature termination of the LLM's exploration (see Figure~\ref{fig:intro} (right)).
\end{enumerate}
Consequently, efficiently balancing entropy in agentic RL remains a fundamental challenge in the pursuit of generalized agent training.

To address these challenges, we propose \textbf{Agentic Entropy-Balanced Policy Optimization (AEPO)}, an entropy-balanced agentic RL algorithm designed for training multi-turn web agents. Unlike traditional entropy-driven RL approaches~\citep{li2025treepobridginggappolicy,dong2025arpo}, AEPO focuses on balancing and rationalizing rollout branching and policy updates under the guidance of high-entropy tool calls, thereby achieving more stable RL training. Specifically, we pioneer the quantification of two inherent entropy-driven challenges on agentic RL.

Building on these insights, AEPO introduces two key algorithmic optimizations: \textbf{(1) Dynamic Entropy-balanced Rollout Mechanism:} To mitigate \textit{``High-entropy Rollout Collapse''} issue, AEPO initially proposes the entropy pre-monitoring to adaptively allocate global and branch sampling budget, ensuring balanced exploration across the tree-structured rollout. Moreover, it incorporates a branch penalty strategy for consecutive high-entropy tool-call steps to effectively address over-branching issues in specific chains. \textbf{(2) Entropy-Balanced Policy Optimization:} Draw inspiration from recent clipping-optimized works~\citep{klear,MiniMax-M1}, we intuitively integrate a \textit{stop-gradient} operation into the high-entropy clipping term during policy updates to tackle the \textit{``High-Entropy Token Gradient Clipping''}. This preserves and properly rescales gradients of high-entropy tokens during backpropagation while leaving the forward pass unchanged. Furthermore, AEPO proposes entropy-aware advantage estimation, integrating entropy advantage into vanilla advantage estimation, enabling the model to prioritize learning on high-uncertainty tokens.


We conduct comprehensive evaluations across 14 datasets covering deep information seeking, knowledge-intensive reasoning, and computational reasoning. As shown in Figure~\ref{fig:intro_performance}, the results show that AEPO consistently outperforms mainstream RL algorithms in generalized reasoning tasks. Remarkably, with only $1k$ RL training samples, Qwen3-14B with AEPO achieves impressive results: \textbf{47.6\% on GAIA, 11.2\% on HLE and 43.0\% on WebWalkerQA for Pass@1; and 65.0\% on GAIA, 26.0\% on Humanity’s Last Exam and 70.0\% on WebWalkerQA for Pass@5.} Further analysis confirms that AEPO effectively broadens sampling diversity during rollouts while maintaining high and stable policy entropy throughout RL training, providing a promising solution for developing general web agents. 

In summary, the key contributions are as follows:
\begin{itemize}[leftmargin=1em]
\item We systematically reveal two entropy-driven issues inherent to agentic RL: {\textit{``High-Entropy Rollout Collapse''}} and {\textit{``High-Entropy Token Gradient Clipping''}}. Through preliminary experiments, we quantify their impact on multi-turn web-agent training, offering empirical evidence for further research into entropy balancing.
\item We propose a \textbf{Dynamic Entropy-Balanced Rollout mechanism}, which adaptively allocates rollout sampling budgets via entropy pre-monitoring, while imposing a branch penalty on consecutive high-entropy steps to prevent over-branching issues.
\item We introduce \textbf{Entropy-Balanced Policy Optimization}, which intuitively integrates a \textit{stop-gradient} operation into the high-entropy clipping term to preserve and rescale gradients on high-entropy tokens, while incorporating entropy-aware advantage estimation to prioritize learning on high-uncertainty tokens.
\item Experiments on 14 challenging benchmarks demonstrate that AEPO consistently outperforms mainstream RL algorithms in web agent training. Quantitative analyses across dimensions such as \textit{Pass@k sampling}, \textit{rollout diversity}, \textit{tool-call efficiency} and \textit{entropy dynamics} verify AEPO's strong scalability and stability, offering valuable insights for developing general web agents.

\end{itemize}

\begin{figure*}[!t]
\centering
\setlength{\abovecaptionskip}{0.1cm}
\setlength{\belowcaptionskip}{0.2cm}
\includegraphics[width=0.985\linewidth]{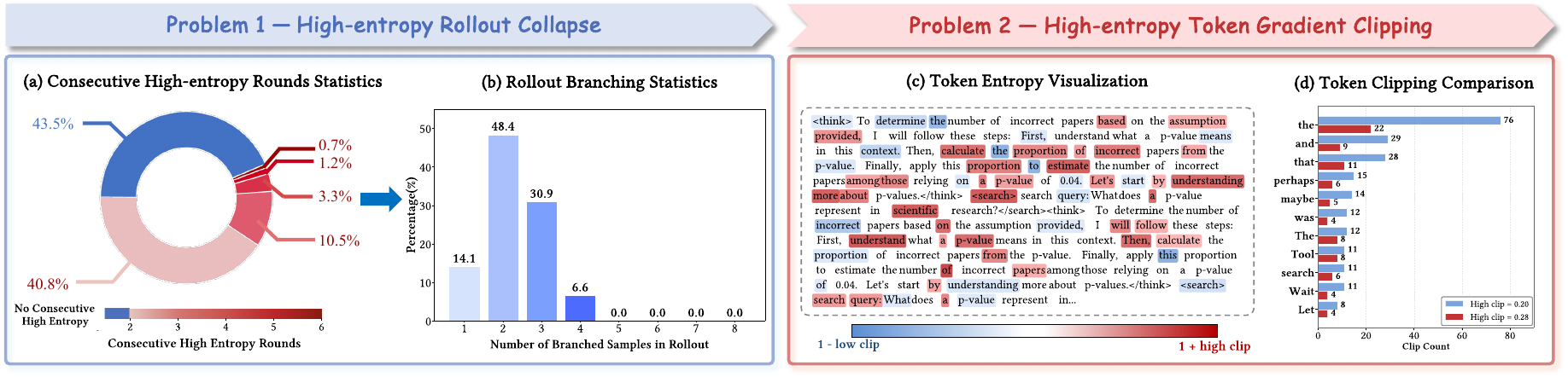}
\caption{
Quantitative statistics of two entropy-based challenges in web agent RL training.}
\label{fig:pre-exp} 
\end{figure*}

\section{Preliminary}

Before introducing the AEPO algorithm, we will briefly outline key task definitions and illustrate preliminary entropy-based experiments to reveal key limitations of web agent RL training.

\subsection{Problem Definition}
\subsubsection{Agentic Reinforcement Learning.}
In this section, we define the training objective for agentic reinforcement learning as follows:
\begingroup\small
\begin{equation}
\small
\max_{\pi_\theta} \mathbb{E}_{x \sim \mathcal{D}, y \sim \pi_\theta(\cdot \mid x; T)} 
\left[ r_\phi(x, y) \right] 
- \beta \, \mathbb{D}_{\text{KL}} \left[ \pi_\theta(y \mid x; T) \,\|\, \pi_{\text{ref}}(y \mid x; T) \right],
\end{equation}
Here, $T$ is the available tool set, $\pi_{\theta}$ and $\pi_{\text{ref}}$ denote the policy and the reference LLM. The symbols $r_{\phi}$ represent the reward functio. The input $x$ is drawn from the dataset $\mathcal{D}$, and $y$ is the corresponding output contain tool-call results.

\subsubsection{Token Entropy Calculation.}
Building on recent studies in entropy-based RL efforts~\citep{20/80,wang2025reinforcement,zhaoxin_entropy,zheng2025first}, we determine the entropy of token generation at step $t$ using the formula:
\begingroup\small
\begin{equation}
\small
H_t = -\sum_{j=1}^{V} p_{t, j} \log p_{t, j}, \quad \text{where } \boldsymbol{p}_t = \pi_\theta\left(\cdot \mid \mathcal{R}_{<t}, x; T\right) = \operatorname{Softmax}\left(\frac{\boldsymbol{z_t}}{\tau}\right).
\end{equation}
In this context, $V$ represents the size of the vocabulary, $\boldsymbol{z_t} \in \mathbb{R}^V$ are the logits before applying the softmax function, and $\tau$ is the temperature parameter for decoding. This entropy quantifies the level of uncertainty in the distribution of token generation.

\subsection{Entropy-based Pilot Experiments}
\label{sec:pre-exp}

In this section, we delves into two high-entropy challenges in web agent training and quantifies their limitations.

\subsubsection{\textbf{Problem-1: High-Entropy Rollout Collapse}} We select ARPO~\citep{dong2025arpo} as the backbone algorithm, a representative entropy-guided agentic RL method, and train with its default $1k$ training samples. We further quantify: {(i)} the steps in each sampled trajectory that exhibit consecutive high-entropy tool usage; {(ii)} within each rollout batch (branching budget is 8), the number and probability of trajectories that contain high-entropy branches. 

\begin{figure*}[!t]
\centering
\setlength{\abovecaptionskip}{0.1cm}
\setlength{\belowcaptionskip}{0.2cm}
\includegraphics[width=0.985\linewidth]{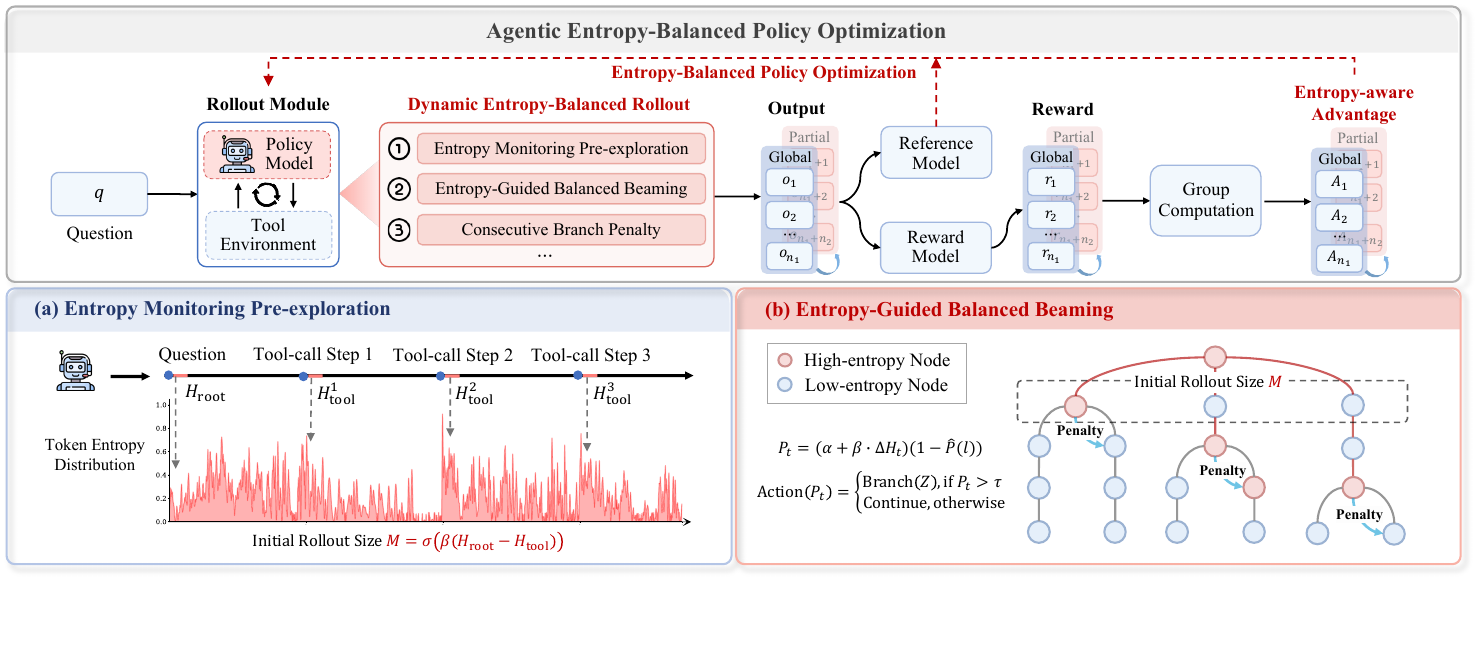}
\vspace{-3em}
\caption{
The overview of Agentic Entropy-Balanced Policy Optimization.
}
\label{fig:method} 
\end{figure*}

As shown in Figure~\ref{fig:pre-exp} (left), our key findings are: \textbf{(i)}~\textbf{High-entropy tool-call turns exhibit transitivity:} the proportion of consecutive high-entropy tool-call turns (56.5\%) exceeds isolated high-entropy turns (43.5\%), with trajectories reaching up to 6 consecutive high-entropy turns. This indicates that high-entropy tool-call rounds often occur consecutively. \textbf{(ii)}~\textbf{Rollout branch collapse:} 93.4\% of branches concentrate on 1–3 trajectories, while the remaining trajectories receive virtually no budget for high-entropy branch sampling. This shows an imbalanced allocation of rollout branching resources.

We argue two observations are tightly coupled: \textbf{Due to an excessive number of consecutive high-entropy rounds in specific samples, the model tends to over-branch on a few trajectories during the rollout phase.} We define this issue as the \textit{``High-Entropy Rollout Collapse''}.

\subsubsection{\textbf{Problem-2: High-Entropy Token Gradient Clipping}} Under the same setup as previous experiment, we further visualize the policy update phase of ARPO, including {(i)}~the importance sampling ratio of tokens in trajectories;\footnote{The token-level importance sampling ratio correlate positively with entropy in RL} {(ii)}~a comparison of the Top-10 gradient-clipped tokens between ARPO and DAPO during a training step, with clipping thresholds of 0.2 and 0.28.

As illustrated in Figure~\ref{fig:pre-exp} (right), we identify the following insights: \textbf{(i)}~Consistent with findings in single-turn RL efforts~\citep{liu2025part,zhaoxin_entropy}, tokens related to logical transitions, connections, and reflections typically exhibit high entropy. Beyond this, specific tool-call tokens also show high entropy. These tokens are highly functional and have low contextual dependency, incentivizing the model to explore diverse reasoning paths and tool-use patterns. \textbf{(ii)}~Vanilla RL method uniformly clip the gradients of high-entropy tokens without distinguishing whether they include valuable exploratory behaviors. Although DAPO adopt clip-higher strategy~\citep{DAPO} to alleviates this by increasing the threshold, the clipping distribution remains similar and the clipped token count is still substantial. 

Moreover, we empirically find that significant gradient clipping emerges in the very first policy update, resulting in \textbf{a lack of gradient support for high-entropy exploratory tokens in early training. This leads to fixed paradigmatic reasoning, hindering the LLM to explore tool-use patterns}. We define this issue as the \textit{``High-entropy Token Gradient Clipping''}.

\subsection{Agentic Tool Design}
In this paper, we focus on exploring entropy-balanced optimization for web agent RL algorithms. To this end, we align with existing work on web agent RL training~\citep{dong2025arpo,li2025websailor,wu2025webdancer} and select three of the most representative tools to evaluate the effectiveness of AEPO: \textbf{(1) Web Search Engine:} Provides retrieved source text and corresponding URL information from the web in response to user queries. \textbf{(2) Web Browser:} Accesses and parses URL information returned by the search engine, then summarizing the content. \textbf{(3) Code Executor:} Executes code generated by models, returning the execution results or error messages.

\section{Methodology}
This section introduces Agentic Entropy-Balanced Policy Optimization~(AEPO), an agentic RL algorithm proposed to balance entropy during both the rollout and policy update phases. As shown in Figure~\ref{fig:method}, AEPO comprises two core components: 

\begin{enumerate}[leftmargin=1em]
\item  \textbf{Dynamic Entropy-Balanced Rollout:} To mitigate the \textit{``High-Entropy Rollout Collapse''} identified in pilot experiments (§\ref{sec:pre-exp}), we adaptively allocate the sampling budget between global and branch sampling via entropy pre-monitoring (§\ref{sec:Entropy Pre-Monitoring}), and penalize consecutive high-entropy tool-call steps during rollout to avoid over-branching (§\ref{sec:Entropy-Balanced Adaptive Rollout}). 

\item  \textbf{Entropy-Balanced Policy Optimization:} To further address \textit{``High-Entropy Token Gradient Clipping''}, we insert a stop-gradient operation into the clipping term to preserve and properly rescale gradients on high-entropy tokens (§\ref{sec:Entropy Clipping-Balanced Mechanism}), while incorporating entropy-aware advantage estimation to prioritize learning on high-uncertainty tokens (§\ref{sec:Entropy-aware Advantage Estimation}). 
\end{enumerate}
Below, we will we delve into the specifics of our approach.

\subsection{Dynamic Entropy-Balanced Rollout}
In this section, we address the \textit{``High-Entropy Rollout Collapse''} by naturally breaking it down into two sub-goals: {(1)} Providing more reasonable resource allocation for global and branch sampling; {(2)} Penalizing continuous high-entropy branch sampling within single trajectories. Consequently, we propose the following two algorithmic solutions.

\subsubsection{\textbf{Entropy Pre-Monitoring.}}
\label{sec:Entropy Pre-Monitoring}
Traditional tree-based rollout empirically allocate resources for global and branch exploration without theoretical support~\citep{li2025treepobridginggappolicy,dong2025arpo,yang2025treerpo,ji2025treesearchllmagent}. Inspired by information bottleneck theory~\citep{tishby2000information}, we advocate the allocation of global and partial branching exploration resources from the perspective of maximizing information gain. Specifically, given a total rollout sampling budget of $k$, which includes $m$ global samples and $k-m$ high-entropy partial branch samples, we simply model the sampling information gain $I_{\text{Gain}}$ per rollout step as:
\begingroup\small
\begin{equation}
\small
I_{\text{Gain}} \  = \ \underbrace{m \cdot I_{\text{root}}}_{\text{Gloabal}} + \underbrace{(k - m) \cdot}_{\text{Partial}} I_{\text{tool}}.
\end{equation}
Here, $I_{\text{root}}$ and $I_{\text{tool}}$ represent the information gain from the input question and external tool-call result. In the autoregressive decoding process of a language model, the information gain of the question is typically measured by the token entropy decoded by the model, with informative questions generally leading to greater uncertainty~\citep{zhu2024information,chen2025revisiting}. Therefore, we derive the following positive correlation:
\begingroup\small
\begin{equation}
\small
\label{eq:info}
I_{\text{Gain}}  \propto \  \underbrace{m \cdot H_{\text{root}}}_{\text{Gloabal}} + \underbrace{(k-m) \cdot H_{\text{tool}}^{\text{avg}}}_{\text{Partial}}, \ \text { where }  H_{\text{tool}}^{\text{avg}} = \frac{1}{N}\sum_{i=1}^{N} H_{\text{tool}}^{i},
\end{equation}
where $H_{\text{root}}$ and $H_{\text{tool}}^i$ represent the entropy of the question and the entropy introduced by the $i$-th tool call, respectively. Based on the formula, we reveal that: \textbf{(1)} When $H_{\text{root}}-H_{\text{tool}}^{avg}>0$, the uncertainty from the initial question surpasses that from the subsequent tools. In this case, we should increase $m$ to enhance global exploration, thereby boosting the information gain $I_{\text{Gain}}$. \textbf{(2)} Conversely, when $H_{\text{root}}-H_{\text{tool}}^{avg}<0$, $m$ should be decreased to allocate more budget to branch exploration via tool calls.

Based on the above theoretical analysis, we propose the entropy pre-monitoring phase. As shown in Figure~\ref{fig:method}(a), we first allow the LLM to generate a complete tool-integrated trajectory for the input $q$. Following ARPO's entropy calculation~\citep{dong2025arpo}, we compute the question and tool average entropies for each trajectory according to Equation~(\ref{eq:info}), forming the entropy matrices $H_{\text{root}} $ and $H_{\text{tool}}^{avg} \in \mathbb{R}^{1 \times k}$. Subsequently, by comparing the $H_{\text{root}}$ and $H_{\text{tool}}$, we dynamically determine the global sampling count $m$ as:
\begingroup\small
\begin{equation}
\label{eq:resource}
\small
m = k \cdot \sigma\left( \beta \cdot (H_{\text{root}} - H_{\text{tool}}^{avg} ) \right),
\end{equation}
where $\sigma(x)$ is the sigmoid function, and $\beta$ controls sensitivity. The value of $m$ is positively correlated with the entropy gap between the question and the tools. As a result, AEPO dynamically allocates rollout sampling resources, thereby enabling efficient sampling.

\subsubsection{\textbf{Entropy-Balanced Adaptive Rollout.}}
\label{sec:Entropy-Balanced Adaptive Rollout}
After entropy pre-monitoring, we introduce the main entropy-balanced adaptive rollout phase to penalize consecutive high-entropy branch sampling, which comprises three core steps:

(1)~\textbf{Entropy Variation Monitoring}: Following resource allocation from the pre-monitoring phase, the LLM generates $m$ global trajectory-level samples for the query $q$, recording the initial entropy matrix $H_{\text{root}}$ for each trajectory. After each tool-call step $t$, the real-time entropy of the model's output is continuously monitored and represented as a step-level entropy matrix $H_t \in \mathbb{R}^{1 \times k}$. The standardized entropy variation relative to the initial entropy is then computed as $\Delta H_t = \text{Normalize}(H_t - H_{\text{root}})$, where the normalization involves dividing the sum of all values in $\Delta H$ by the vocabulary size $V$. 

(2)~\textbf{Entropy-Balanced Beaming}: Unlike traditional entropy-guided branch sampling~\citep{dong2025arpo,zheng2025returnentropyelicitingexplore}, AEPO promotes adaptive exploration that showcases beneficial entropy changes in tool-call steps while also constraining consecutive high-entropy branch sampling in specific chains.  As shown in Figure~\ref{fig:method}(b), we introduce a consecutive branch penalty strategy. Given a tool-call step $t$, the number of consecutive high-entropy branches $l$ prior to step $t$ for each chain is tracked, then defining the branch sampling probability at step $t$ as follows:
\begingroup\small
\begin{equation}
\small
P_t = (\alpha + \gamma \cdot \Delta H_t)(1 - \hat{P}(l)),
\end{equation}
where $\alpha$ is the base sampling probability and $\gamma$ is the entropy stabilization factor. $\hat{P}(l)$ is a linear function related to $l$. $P_t$ decreases as the number of consecutive branching steps $l$ increases, implementing a consecutive branching penalty. \textbf{This design makes the tree-structured rollout sampling more diverse, allowing for a more comprehensive coverage of the problem-solving space.} We then define the rollout branching action at step $t$ as:
\begingroup\small
\begin{equation}
\text{Action}(P_t) = \begin{cases} 
\text{Branch}(Z), & \text{if } P_t > \tau, \\
\text{Continue}, & \text{otherwise}.
\end{cases}
\end{equation}
When $P_t$ exceeds the predetermined threshold $\tau$, $\text{Branch}(Z)$ is initiated, creating $Z$ partial braching reasoning paths from the current node; otherwise, the current trajectory continues.

(3)~\textbf{Termination Conditions:} Finally, our iterative rollout process terminates when one of the following conditions is met: (a)~If the total number of branch paths $Z^*$ reaches the partial sampling budget $k-m$, branching stops and sampling continues until the final answer is generated; (b)~If all paths terminate before reaching $k-m$, then $k-m-Z^*$ additional trajectory-level samples are added to satisfy condition (a).

Through the dynamic entropy-balanced rollout, AEPO ensures the diversity of sampling branches while adaptively allocating exploration resources, thus addressing the \textit{"High-entropy Rollout Collapse"} issue. The algorithm of dynamic entropy-balanced rollout is detailed in Algorithm~\ref{algo:der}.

\subsection{Entropy-Balanced Policy Optimization}

AEPO preserves a considerable number of exploratory tokens via entropy-balanced rollouts, presenting a challenge in effectively updating these tokens' gradients. This section aims to improve targeted learning for these tokens by implementing the following designs:

\subsubsection{\textbf{Entropy Clipping-Balanced Mechanism.}}
\label{sec:Entropy Clipping-Balanced Mechanism}
Unlike traditional methods that entirely discard gradients outside the clipping range~\citep{deepseekmath,dong2025arpo}, AEPO introduces an innovative high-entropy clipping-balanced mechanism. The core idea is to retain high-entropy gradients that exceed the clipping interval, allowing the model to learn valuable exploratory token signals. 

Motivated by GPPO~\citep{klear}, we integrate a \textit{stop-gradient} operation into the high-entropy clipping term of the policy update phrase, decoupling forward and backward propagation. Our mechanism ensures that forward propagation remains unchanged, while protecting the gradient backward of high-entropy tokens from clipping constraints.
For instance, in GRPO~\citep{deepseekmath}, given an input question $x$ and a policy model $y$, GRPO enables the reference policy $\pi_{\text{ref}}$ to generate a group of $G$ outputs $\{y_1, y_2, \ldots, y_G\}$ and optimizes the policy by maximizing:
\begin{equation}
\small
\mathcal{L}=\mathbb{E}_{x \sim \mathcal{D}}\left[\frac{1}{\sum_{j=1}^{G} T_{j}} \sum_{j=1}^{G} \sum_{t=1}^{T_{j}} \min \left(\delta \tilde{A}^{(t)}, \operatorname{clip}\left(\delta, 1-\epsilon_{l}, \frac{1+\epsilon_{h}}{\operatorname{sg}(\delta)} \delta\right) \tilde{A}^{(t)}\right)\right],
\end{equation}
where $\delta = r_t^{(j)}(\theta)$ represents the importance sampling ratio, and $sg(\cdot)$ denotes the \textit{stop-gradient} operation. It is noteworthy that \textbf{the value of the term $\delta \cdot sg(\delta)$ always equals 1, ensuring that AEPO's forward computation remains unchanged}. For the backpropagation, AEPO's gradient update process is formulated as:
\begin{equation}
\small
\begin{array}{c}\nabla_{\theta} \mathcal{L}=\mathbb{E}_{x \sim \mathcal{D}}\left[\frac{1}{\sum_{j=1}^{G} T_{j}} \sum_{j=1}^{G} \sum_{t=1}^{T_{j}} \mathcal{F}_{j, t}(\theta) \cdot \phi_{\theta}\left(a_{j, t}, s_{j, t}\right) \cdot
\tilde{A}^{(t)}\right], \\ \\\text { where } \mathcal{F}_{j, t}(\theta)=\left\{\begin{array}{ll}1+\epsilon_{h}, & \text { if } \delta>1+\epsilon_{h} \text { and } \tilde{A}^{(t)}>0,  \\
0, & \text { if } \delta<1-\epsilon_{h} \text { and } \tilde{A}^{(t)}<0, 
\\\delta, & \text { otherwise }.\end{array}\right.\end{array}
\label{eq:gradient_update}
\end{equation}

During backpropagation, the gradient of a high-entropy token is retained and appropriately rescaled to $1+\epsilon_{h}$ only when $\delta>1+\epsilon_{h}$ and $\tilde{A}^{(t)}>0$. In other cases, the gradient update rule aligns with vanilla clipping mechanism of GRPO. This controlled rescaling ensures that the model learns a balanced exploratory behavior without completely ignoring high-entropy tokens. To more clearly articulate the theoretical aspects of AEPO compared to clipping-optimized RL methods~\citep{MiniMax-M1,su2025gppo}, we discuss their differences in Appendix~\ref{app:discussion}~\footnote{For detailed proof of the gradient form of AEPO, please refer to Appendix~\ref{app:proof}}.


\begin{algorithm}[t]
\caption{Dynamic Entropy-Balanced Rollout}
\label{algo:der}
\begin{algorithmic}[1]
\REQUIRE Reasoning model $\pi_\theta$, external tools $T$,
total rollout size $k$, entropy sensitivity $\beta$, branch penalty slope $\gamma$
\STATE \textbf{Input:} Dataset $D$
\STATE Initialize reference model: $\pi_{\theta}^{\text{old}} \gets \pi_\theta$
\FOR{$i = 1$ to $\mathcal{C}$}
    \STATE Sample mini-batch $D_b \subset D$
    
    \FOR{each query $q \in D_b$}      
        \STATE \textcolor{blue}{// Entropy Pre-Monitoring}
        \STATE Generate 1 complete trajectory $r$ to obtain $H_{\text{root}}$ and $H_{\text{tool}}^{\text{avg}}$
        \STATE Global rollout size $m \gets k\cdot\sigma\!\bigl(\beta(H_{\text{root}}-H_{\text{tool}}^{\text{avg}})\bigr)$
        \STATE Branch rollout size $b \gets k - m$

        \STATE \textcolor{blue}{// Entropy-Balanced Adaptive Rollout}
        \STATE Initialize rollout pool $\mathcal{P}\gets\varnothing$
        \STATE Consecutive-high-entropy counter $l\gets 0$
        \WHILE{$|\mathcal{P}|<m$}
            \STATE Sample trajectory $r$; add to $\mathcal{P}$
        \ENDWHILE
        \WHILE{$b>0$ \textbf{and} $\exists\, r_j \in \mathcal{P}$ not terminated}
            \STATE \textcolor{deepred}{// (1) Entropy Variation Monitoring}
            \STATE Select a trajectory $r\in\mathcal{P}$ at tool-call step $t$
            \STATE $\Delta H_t \gets \text{Normalize}(H_t - H_{\text{initial}})$
            \STATE \textcolor{deepred}{// (2) Entropy-Balanced Beaming}
            \STATE Consecutive penalty $\hat P(l)\gets \gamma \cdot l$
            \STATE Branch probability $P_t\gets (\alpha+\beta\Delta H_t)(1-\hat P(l))$
            \STATE \textcolor{deepred}{// (3) Termination Conditions}
            \IF{$P_t>\tau$}
                \STATE Branch $Z$ sub-trajectories; \;$b\gets b-Z$
            \ELSE
                \STATE $l\gets l+1$ if $\Delta H_t>0$
            \ENDIF
        \ENDWHILE
        \IF{$b>0$}
            \STATE Sample $b$ additional trajectories and add to $\mathcal{P}$
        \ENDIF
    \ENDFOR
\ENDFOR
\STATE \textbf{Output:} rollout trajectory set $\mathcal{P}$
\end{algorithmic}
\end{algorithm}

\subsubsection{\textbf{Entropy-aware Advantage Estimation.}} 
\label{sec:Entropy-aware Advantage Estimation}
Owing to the clipping-balanced mechanism, we retain the gradients of high-entropy tokens. However, a challenge arises in training the model to better distinguish between exploratory and non-exploratory tokens. Traditional outcome-based RL algorithms assign the same advantage to all tokens in a sequence based on the answer correctness, neglecting the model's confidence levels across different tokens~\citep{deepseekmath,hu2025reinforce++}.

To this end, we propose an \textbf{entropy-aware advantage estimation} that incorporates token entropy calculation into advantage shaping. This approach allows the model to assign greater rewards to exploratory tokens that are correct but exhibit high uncertainty. A natural way is to calculate an accuracy-based advantage while integrating an entropy-based advantage term, defined as follows:
\begin{equation}
\small
\tilde{A}^{(t)}_{\text{Acc}}=\frac{r_{t}-\operatorname{mean}\left(\left\{R_{i}\right\}_{i=1}^{G}\right)}{\operatorname{std}\left(\left\{R_{i}\right\}_{i=1}^{G}\right)}, \ \  
\tilde{A}^{(t)}_{\Delta H}=\frac{H_{t}-\operatorname{mean}\left(\left\{H_{t}\right\}_{t=1}^{T}\right)}{\operatorname{std}\left(\left\{H_{t}\right\}_{t=1}^{T}\right)},
\label{eq:Advantage}
\end{equation}
where $H_t$ represents the $t$-th token entropy, and $T$ is the total number of tokens across all trajectories in the group. . We estimate the entropy advantage for each token based on the average token entropy within the same trajectory. Furthermore, we treat the entropy advantage as a regularization term in the advantage estimation to reshape $A_{\text{acc}}$ as:
\begin{equation}
\small
\tilde{A}^{(t)}=\tilde{A}^{(t)}_{\text{Acc}} *\left(1+a \cdot \tilde{A}^{(t)}_{\Delta H}\right).
\end{equation}

This step is computed before the policy update. Notably, our entropy-aware advantage estimation can be seamlessly integrates with existing agentic RL algorithms to further enhance the model's emphasis on learning exploratory tokens during training. The full algorithm of AEPO is detailed in Algorithm~\ref{algo:aepo}.

\begin{table*}[!t]
\centering
\small
\caption{Overall performance on deep information seeking tasks. The best results are indicated in \textbf{bold}, and the second-best results are \underline{underlined}. Results from larger or closed-source models are presented in \textcolor{gray!135}{gray} for reference.}
\vspace{-1em}
\label{tab:deep_search}
\setlength\tabcolsep{0.5pt}
\renewcommand{\arraystretch}{1}
\begin{tabular}{
    p{2.8cm} 
    *{4}{>{\centering\arraybackslash}p{0.9cm}} 
    *{4}{>{\centering\arraybackslash}p{0.9cm}}  
    *{4}{>{\centering\arraybackslash}p{0.9cm}} 
    >{\centering\arraybackslash}p{1.7cm}        
    >{\centering\arraybackslash}p{1.7cm}        
}
\toprule
\multirow{2}{*}{\textbf{Method}}
  & \multicolumn{4}{c}{\textbf{General AI Assistant}}
  & \multicolumn{4}{c}{\textbf{WebWalkerQA}}
  & \multicolumn{4}{c}{\textbf{Humanity's Last Exam}}
  & \textbf{XBench-DR} & \textbf{FRAMES} \\
\cmidrule(lr){2-5} \cmidrule(lr){6-9} \cmidrule(lr){10-13} \cmidrule(lr){14-14} \cmidrule(lr){15-15}
  & Lv.1 & Lv.2 & Lv.3 & Avg.
  & Easy & Med. & Hard & Avg.
  & NS & CE & SF & Avg.
  & Avg. & Avg. \\
\midrule

\multicolumn{14}{l}{\textit{\textbf{Direct Reasoning (>=32B)}}} \\
Qwen3-32B-thinking & 26.2 & 12.1 & 0 & 14.9 & 6.9 & 1.1 & 2.9 & 3.1 & \underline{14.6} & \underline{9.8} & 8.4 & 12.6 & 14.0 & 26.0 \\
DeepSeek-R1-32B    & 21.5 & 13.6 & 0.0 & 14.2 & 7.5 & 1.4 & 4.2 & 3.8 & 6.6 & 5.1 & 6.5 & 6.4 & 10.0 & 23.8\\ 
QwQ-32B            & 30.9 & 6.5  & 5.2 & 18.9 & 7.5 &2.1  &4.6  & 4.3 & 11.5 & 7.3 & 5.2 & 9.6 & 10.7 & 28.8\\
GPT-4o             & \textcolor{gray!135}{23.1} & \textcolor{gray!135}{15.4} & \textcolor{gray!135}{8.3} & \textcolor{gray!135}{17.5} & \textcolor{gray!135}{6.7} &\textcolor{gray!135}{6.0}  &\textcolor{gray!135}{4.2}  & \textcolor{gray!135}{5.5} & \textcolor{gray!135}{2.7} & \textcolor{gray!135}{1.2} & \textcolor{gray!135}{3.2} & \textcolor{gray!135}{2.6} & \textcolor{gray!135}{18.0} & 44.6 \\
DeepSeek-R1-671B   & \textcolor{gray!135}{40.5} & \textcolor{gray!135}{21.2} & \textcolor{gray!135}{5.2} & \textcolor{gray!135}{25.2} & \textcolor{gray!135}{5.0} &\textcolor{gray!135}{11.8}  &\textcolor{gray!135}{11.3}  & \textcolor{gray!135}{10.0} & \textcolor{gray!135}{8.5} & \textcolor{gray!135}{8.1} & \textcolor{gray!135}{9.3} & \textcolor{gray!135}{8.6} & \textcolor{gray!135}{32.7} & 45.6 \\
o1-preview$^\dagger$ & \textcolor{gray!135}{-} & \textcolor{gray!135}{-} & \textcolor{gray!135}{-} & \textcolor{gray!135}{-} &\textcolor{gray!135}{11.9}  & \textcolor{gray!135}{10.4} &\textcolor{gray!135}{7.9}  & \textcolor{gray!135}{9.9} & \textcolor{gray!135}{12.9} & \textcolor{gray!135}{8.1} & \textcolor{gray!135}{6.6} & \textcolor{gray!135}{11.1} & \textcolor{gray!135}{-} & \textcolor{gray!135}{-}\\
\midrule
\multicolumn{14}{l}{\textit{\textbf{Single-Enhanced Method (Qwen3-8B)}}} \\
Vanilla RAG  & 28.2 & 15.4 & \textbf{16.7} & 20.4 &  8.9 & 10.7 & 9.9 & 10.0 & 5.1 & 1.6 & \underline{12.9} & 5.8 & 8.0 & 18.8 \\
Search-o1    & 35.9 & 15.4 & 0.0  & 21.4 &  6.7 & 15.5 & 9.7 & 11.5 & 7.6 & 2.7 & 5.3  & 6.4 & 10.0 & 19.2 \\
WebThinker   & 43.6 & 11.5 & 0.0  & 22.3 & 6.7 & 13.1 & 16.9 & 13.0 & 7.3 & 4.0 & 6.3 & 6.6 & 13.0 & 21.4 \\
ReAct        & 35.9 & 17.3 & \underline{8.3}  & 23.3 &  8.9 & 16.7 & 18.3 & 15.5 & 4.2 & 4.0 & 6.3 & 4.6 & 16.0 & 21.1 \\
\hdashline
\multicolumn{14}{l}{\textit{\textbf{RL-based Method (Qwen3-8B)}}} \\
Qwen3-8B & 28.1 & 15.4 & \textbf{16.7} & 20.4 & 0.0 & 2.4 & 2.8 & 2.0 & 3.9 & 2.7 & 8.4  & 4.6& 9.0 & 19.0 \\
\;\; + GRPO & 48.7 & 25.0 & \underline{8.3} & 32.0 & 28.9 & 32.1 & \underline{28.2} & 30.0 & \underline{7.9} & 4.0 & 10.5  & 7.8 & 20.0& 46.2\\
\;\; + ARPO & \underline{53.9} & \underline{32.7} & \textbf{16.7} & \underline{38.8} & \underline{31.1} & \underline{35.7} & \underline{28.2} & \underline{32.0} & 7.3 & \textbf{6.7} & \textbf{15.8} & \underline{8.8} & \underline{25.0} & \underline{47.8}\\

\rowcolor[HTML]{f0edff}\;\; + AEPO (Ours) & \textbf{61.5} & \textbf{42.3} & \underline{8.3} & \textbf{45.6} & \textbf{40.0} & \textbf{39.3} & \textbf{35.2} & \textbf{38.0} & \textbf{12.1} & \underline{5.3} & 11.6 & \textbf{11.0} & \textbf{28.0} & \textbf{50.2}\\

\midrule
\multicolumn{14}{l}{\textit{\textbf{Single-Enhanced Method (Qwen3-14B)}}} \\
Vanilla RAG  & 38.5 & 19.2 & \underline{8.3}  & 25.2 & 17.8 & 13.1 & 11.3 & 13.5 & 5.5 & 6.3 & 9.4 & 6.0 & 15.0 & 31.4 \\
Search-o1    & 48.7 & 23.1 & 0.0  & 30.1 &  11.1 & 21.4 & 16.9 & 17.5 & 6.4 & 4.0 & 10.5 & 6.8 & 21.0 & 39.8 \\
WebThinker   & 48.7 & 26.9 & \underline{8.3}  & 33.0 &  13.3 & 23.8 & 18.3 & 19.5 & 7.0 & 4.0 & 9.5 & 7.0 & 23.0 & 40.8 \\
ReAct        & 48.7 & 25.0 & \underline{8.3}  & 32.0 &  11.1 & 20.2 & 12.7 & 15.5 & 5.8  & 5.3 & 10.5 & 6.6 & 20.0 & 37.6\\
\hdashline
\multicolumn{14}{l}{\textit{\textbf{RL-based Method (Qwen3-14B)}}} \\
Qwen3-14B & 33.3 & 13.5 & 0.0 & 19.4 & 6.7 & 2.4 & 4.2 & 4.0 & 5.5 & 6.7 & 11.6 & 6.8 & 14.0 & 23.8 \\
\;\; + GRPO & 51.3 & 34.6 & 0.0 & 36.9 & \underline{35.6} & 42.9 & 35.2 & 38.5 & 7.9 & 6.7 & \underline{12.6} & 8.6 & \underline{27.0} & 54.6\\
\;\; + ARPO & \underline{56.4} & \underline{40.4} & \textbf{16.7} & \underline{43.7} & \textbf{40.0} & \underline{44.1} & \underline{36.6} & \underline{40.5} & \underline{10.3} & \underline{10.7} & \textbf{13.7} & \underline{10.0} &\textbf{32.0} & \underline{55.4}\\

\rowcolor[HTML]{f0edff}\;\; + AEPO (Ours) & \textbf{61.5} & \textbf{44.2} & \textbf{16.7} & \textbf{47.6} & \textbf{40.0} & \textbf{50.0} & \textbf{40.9} & \textbf{44.5} & \textbf{10.6} & \textbf{14.7} & 10.5  & \textbf{11.2}& \textbf{35.0} & \textbf{58.8}\\
\bottomrule
\vspace{-3em}
\end{tabular}
\end{table*}

\begin{figure*}[!t]
\centering
\vspace{-0.5em}
\setlength{\abovecaptionskip}{0.1cm}
\setlength{\belowcaptionskip}{0.1cm}
\includegraphics[width=0.98\linewidth]{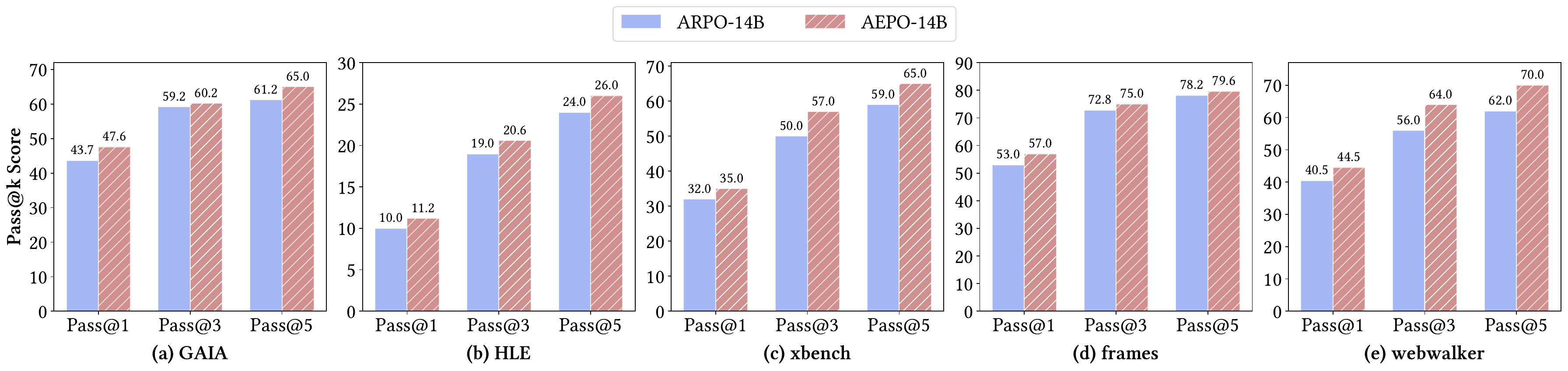}
\caption{The comparison analysis of Qwen3-14B using ARPO and AEPO across Pass@1 to Pass@5 metrics.
}
\label{fig:passk}
\end{figure*}

\section{Experiment Settings}

\label{sec:exp_setup}
\subsection{Datasets}
We assess the effectiveness of AEPO in web agents RL training through three long-term reasoning tasks:
\begin{enumerate}[leftmargin=1em]
\item \textbf{Deep Information Seeking Tasks:} This includes challenging evaluations for web agents: General AI Assistant (GAIA)~\citep{GAIA} and the Human Last Exam (HLE)~\citep{HLE}, as well as deep information seeking: WebWalkerQA~\citep{2501_WebWalker}, XBench~\citep{chen2025xbench}, and Frames~\citep{frames}.
\item \textbf{Knowledge-Intensive Reasoning:} This covers 3 multi-hop complex open-domain question-answering tasks: 2WikiMultihopQA~\citep{2wiki}, Musique~\citep{musique}, and Bamboogle~\citep{bamboogle}, along with the web multi-hop task WebWalkerQA~\citep{wu2025webwalker}.
\item \textbf{Computational Reasoning:} This includes simple math reasoning tasks like GSM8K\citep{cobbe2021gsm8k}, MATH~\citep{MATH}, and competition-level math challenges: MATH500~\citep{math500}, AIME2024, and AIME2025.\footnote{\url{https://huggingface.co/datasets/AI-MO/aimo-validation-aime}}
All dataset splits align with the standard settings established by previous works~\citep{webthinker,dong2025arpo,hira}.
\end{enumerate}

\subsection{Baselines}
We consider the following strong baseline methods:
\begin{enumerate}[leftmargin=1em]
\item \textbf{Advanced RL Algorithms:} We select three categories of RL  algorithms: (1) Vanilla RL: GRPO\citep{deepseekmath} and Reinforce++\citep{hu2025reinforce++}; (2) Clipping-optimized RL: DAPO\citep{DAPO}, CISPO\citep{minimax2025minimaxm1scalingtesttimecompute} and GPPO\citep{klear}; and (3) Agentic RL: GIGPO\citep{GIGPO} and ARPO\citep{dong2025arpo}.
\item \textbf{Advanced Backbone Models:} For challenging reasoning benchmarks, we evaluate the instruction-tuned versions of Qwen2.5\citep{qwen2.5} and Llama3.1\citep{llama3}. For deep information seeking tasks, we also report results for QwQ\citep{qwen_qwq}, DeepSeek-R1\citep{deepseek-r1}, GPT-4o\citep{gpt_4o_system_card}, and o1-preview\citep{gpt_4o_system_card}, using Qwen3-32B\citep{qwen3} as a reference.
\item \textbf{Advanced Web Agents:} We introduce a series of open-source workflow-based search agents as references, including vanilla RAG\citep{rag_lewis}, Search o1\citep{searcho1}, Webthinker\citep{webthinker}, and ReAct\citep{yao2022react}. The detailed introduction of baselines are listed in Appendix~\ref{app:baselines}
\end{enumerate}
\subsection{Evaluation Metric} 
Consistent with previous work, we use the F1 score to evaluate four question-answering tasks that require intensive knowledge reasoning. For other tasks, we employ the VLLM framework to serve Qwen2.5-72B-instruct, using LLM-as-Judge to assess the answers. In all tasks, answers are extracted from the $\text{\\box{}}$ in the respons. By default, the temperature is set to 0.6 and top-p to 0.95, and we evaluate using the Pass@1 score. 


\subsection{Implementation Details}
In the AEPO phase, we implement the AEPO algorithm using the VERL framework~\citep{sheng2024hybridflow}, excluding tool-call results from the loss calculation to avoid bias. Our setup includes a training batch size of 128, a PPO mini-batch size of 16, and a context length of 20K. For AEPO rollout, the global rollout size is 16, with $a, \beta$ set to 0.2. Resource allocation follows Equation~\ref{eq:resource}, with a consecutive branch penalty probability of $P(l)=0.2\cdot l$. Other settings align with ARPO for fair comparison. To stabilize RL training, the KL divergence coefficient in GRPO is set to 0. The RL for reasoning and deep information seeking is 2 and 5 epochs. All experiments use 16 NVIDIA H800 GPUs. 

During training and evaluation, we use the Bing Search API (US-EN region) as the search engine. Following RAG-related work~\citep{webthinker,hira}, we retrieve 10 web pages per query. For reasoning tasks, we use the top 10 snippets; for deep information seeking, we extract up to 6000 tokens per page and use a same size model as a browser agent.

\begin{table*}[t]
\centering
\small
\caption{Overall performance on ten challenging reasoning tasks are presented. The top two outcomes are \textbf{bolded} and \underline{underlined}.}
\vspace{-1em}
\label{tab:general_reasoning}
\setlength\tabcolsep{4pt}
\renewcommand{\arraystretch}{1}
\begin{tabular}{p{2.8cm}ccccccccccc}
\toprule
\multirow{2}[2]{*}{\textbf{Method}} & \multicolumn{5}{c}{\textbf{Mathematical Reasoning}} & \multicolumn{5}{c}{\textbf{Knowledge-Intensive Reasoning}} & \multirow{2}[2]{*}{\textbf{Avg.}} \\
\cmidrule(lr){2-6} \cmidrule(lr){7-11}
& AIME24 & AIME25 & MATH500 & GSM8K & MATH & WebWalker & HotpotQA & 2Wiki. & MuSiQue. & Bamboogle & \\

\midrule
\multicolumn{12}{c}{\textit{\textbf{Backbone Model: Llama3.1-8B-Instruct}}} \\
\midrule

\multicolumn{12}{l}{\textit{\textbf{Classical RL Method}}} \\
\quad + GRPO                 & 13.3 & \underline{13.3} & 62.4 & 87.4 & 79.2 & 26.5 & 57.8 & 71.8 & 31.0 & 68.2 & 51.1 \\
\quad + Reinforce ++         & 13.3 & \textbf{16.7} & 61.4 & 87.0 & 77.2 & 27.5 & 57.1 & 71.6 & 29.9 & 69.1 & 51.1 \\
\hdashline

\multicolumn{12}{l}{\textit{\textbf{Clipping-optimized RL Method}}} \\
\quad + DAPO                 & 16.7 & \underline{13.3} & 61.2 & 87.4 & 76.4 & 25.5 & 56.6 & 70.3 & 29.2 & 67.3 & 50.4 \\

\quad + GPPO & 16.7& 6.7 & 61.8 & 86.6 & 79.4 & 27.5 & 61.8 &72.8 & 29.8 & 71.9 & 51.5 \\
\quad + CISPO & 13.3 & 6.7 & 62.2 & 87.0 & 78.2 & 26.0 & 57.3 & \underline{75.6} & 32.2 & 71.8 & 51.0 \\
\hdashline

\multicolumn{12}{l}{\textit{\textbf{Agentic RL Method}}} \\
\quad + GIGPO &20.0 &13.3 & 62.4 & 87.4& 77.2 & \underline{31.5} & 61.8 & 74.6& 31.8 & 72.1 & 53.2\\
\quad + ARPO                 & \underline{23.3} & \textbf{16.7} & \underline{64.6} & \textbf{88.0} & \underline{80.2} & 30.5 & \textbf{65.4} & 75.5 & \textbf{34.8} & \underline{73.8} & \underline{55.3} \\
 
\rowcolor[HTML]{f0edff}\;\;\;  + AEPO (Ours)                 & \textbf{26.7} & \textbf{16.7 }& \textbf{65.8} & \underline{87.6} & \textbf{80.6} & \textbf{33.5} & \underline{64.7} & \textbf{79.0} & \underline{33.0} & \textbf{75.8} & \textbf{56.3} \\

\midrule
\multicolumn{12}{c}{\textit{\textbf{Backbone Model: Qwen2.5-7B-Instruct}}} \\
\midrule
\multicolumn{12}{l}{\textit{\textbf{Classical RL Method}}} \\
\quad + GRPO                 & 23.3 & \underline{26.7} & 78.0 & \textbf{92.8} & 87.8 & 22.0 & 59.0 & \underline{76.1} & 30.6 & 68.4 & 56.5 \\
\quad + Reinforce ++         & 26.7 & 23.3 & 78.0 & \underline{92.2} & \underline{88.8} & 26.0 & 55.1 & 68.9 & 25.2 & 64.9 & 54.9 \\
\hdashline
\multicolumn{12}{l}{\textit{\textbf{Clipping-optimized RL Method}}} \\
\quad + DAPO                 & 20.0 & 23.3 & \textbf{80.4} & 91.0 & \underline{88.8} & 24.0 & 57.7 & 68.4 & 28.6 & 65.5 & 54.8 \\
\quad + GPPO & 26.7 & 23.3 & 76.2 & 91.6 & 87.6 & \underline{31.0} & \underline{60.7} &74.2 & \textbf{31.5} & \underline{72.4} & 57.5 \\
\quad + CISPO & 26.7 & \textbf{30.0} & 77.8 & 91.4 & 86.2 & 29.0 & 59.3 & 72.1 & 29.1 & 70.2 & 57.2 \\
\hdashline
\multicolumn{12}{l}{\textit{\textbf{Agentic RL Method}}} \\
\quad + GIGPO & \underline{30.0} & 20.0& 78.4 & 91.6 &87.6 & 30.5 & 58.1&73.5 & \underline{31.1} & 70.1 & 57.1 \\
\quad + ARPO                 & \underline{30.0} & \textbf{30.0} & \underline{78.8} & \underline{92.2} & \underline{88.8} & 26.0 & 58.8 & \underline{76.1} & \underline{31.1} & 71.5 & \underline{58.3} \\

\rowcolor[HTML]{f0edff}\;\;\;   + AEPO (Ours) & \textbf{33.3} & \textbf{30.0} & \textbf{80.4} & \underline{92.2} & \textbf{90.0} & \textbf{31.5} & \textbf{62.5} & \textbf{77.1} & \underline{31.1} & \textbf{73.4} & \textbf{60.1}\\
\bottomrule
\vspace{1em}
\end{tabular}
\end{table*}

\section{Experiment Results}
\subsection{Main Result on Deep Information Seeking}
To validate the effectiveness of AEPO in challenging deep web information seeking tasks, we train the Qwen3 series models combined with AEPO using $1K$ open-source samples and compared them with advanced web agents and RL algorithms. As shown in Table~\ref{tab:deep_search}, we derived the following insights:

(1) \textbf{Limitations of Advanced Large Models:} Both advanced closed-source LLMs and large-parameter open-source LLMs (e.g. GPT-4o and DeepSeek-R1-671B) perform poorly in challenging deep information seeking scenarios, particularly on the GAIA (<30\%) and HLE (<10\%) benchmarks. This indicates that relying solely on model internal knowledge is insufficient for complex agentic search tasks.

(2) \textbf{Strong Generalization Ability of AEPO in Deep Information Seeking:} Compared to robust web agents and advanced RL algorithms, the Qwen3-8B and 14B models combined with AEPO demonstrate exceptional performance, achieving pass@1 scores of 11.2\%, 47.6\% and 43\% on the HLE, GAIA and WebWalkerQA benchmarks, respectively. Notably, our model is trained solely on 1k samples from an open-source web search dataset, without any data synthesis or filtering, showcasing its efficiency in training web agents.

(3) \textbf{Importance of Dual Entropy Balancing Optimization:} AEPO consistently outperforms ARPO in both average performance and individual benchmarks, with Qwen3-8B showing a significant 6\% improvement on the GAIA benchmark and WebWalkerQA. This highlights the importance of AEPO's algorithmic design, which implements dual entropy balancing in both the Rollout and policy update phases, effectively facilitating LLMs' exploratory tool behavior and addressing two high-entropy challenges. This is crucial for deep information seeking scenarios involving frequent tool invocation.

\begin{figure}[t]
    \centering
\setlength{\abovecaptionskip}{0.2cm}
\setlength{\belowcaptionskip}{0.2cm}
    \includegraphics[width=1\linewidth]{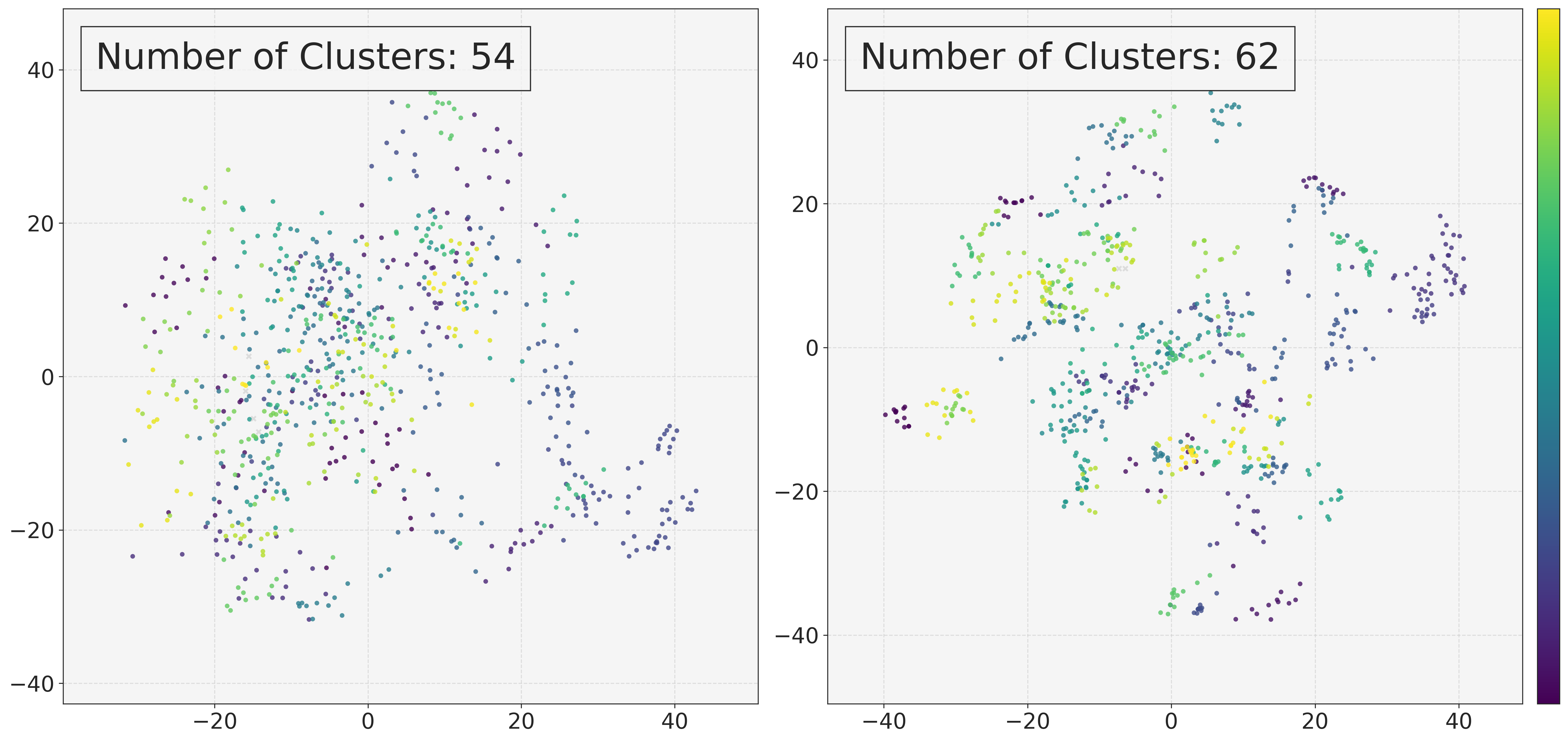}
    \caption{Visualization of Rollout diversity: ARPO (left) and AEPO (right)}
    \label{fig:cluster}
\end{figure}

\subsection{Main Result on Generalized Reasoning}

To further validate the effectiveness of AEPO in web agent training, we conduct a comparision of AEPO with 7 RL algorithms across 10 challenging reasoning tasks. As shown in Table~\ref{tab:general_reasoning}, our key insights are as follows:

(1) \textbf{Instability of Clipping RL Algorithms in Web Agent Training:} Using GRPO as a baseline, clipping-optimized RL algorithms perform well on Qwen 2.5-7B-instruct, with GPPO and CISPO achieving average scores above 57\%. However, in Llama3-8B, they do not show significant improvement over GRPO. Furthermore, practical experiments reveal that clipping-optimized RL algorithms often lead to entropy collapse, disrupting training performance. This indicates that clipping-optimized RL algorithms are sensitive to the architecture of the backbone model and often show instability during web agent training.

(2) \textbf{Generalization Ability of Agentic RL Algorithms:} Agentic RL algorithms, represented by ARPO and GIGPO, demonstrate stable and robust performance across both backbone models, with ARPO achieving average performance consistency above 55\%. Notably, these methods attempt tree-structured rollout during the rollout phase, further confirming the effectiveness of branching exploration in high-entropy tool-call steps.

(3) \textbf{Effectiveness of AEPO}: AEPO consistently outperforms other reinforcement learning algorithms across 10 datasets and backbone models, achieving an average accuracy improvement of nearly 5\% over GRPO while maintaining competitiveness across fine-grained domains. These results highlight AEPO's efficiency and strong adaptability across different model architectures and tasks, making it more suitable than other RL algorithms for training multi-turn web agents.

\subsection{Pass@K Sampling Analysis} 
Due to the dynamic multi-turn interactions and complexity of tool environments in web agent training, we conduct a sampling analysis of the model's Pass@3 and Pass@5 to accurately assess its true problem-solving abilities. 

As illustrated in Figure~\ref{fig:passk}, AEPO demonstrates significant performance improvements with larger-scale sampling. Notably, the Qwen3-14B model combined with AEPO achieves remarkable results: GAIA at 65\%, HLE at 26\%, and XBench-DR at 65\%. Compared to the robust agentic RL algorithm ARPO, AEPO consistently excels across five datasets. This stable improvement in Pass@K can be primarily attributed to AEPO's entropy balancing optimizations, which allows the model to explore fine-grained tool usage behaviors more efficiently, thereby enhancing reasoning and sampling efficiency.

\begin{figure}[t]
    \centering
    \setlength{\abovecaptionskip}{0.1cm}
\setlength{\belowcaptionskip}{0.2cm}
    \includegraphics[width=1\linewidth]{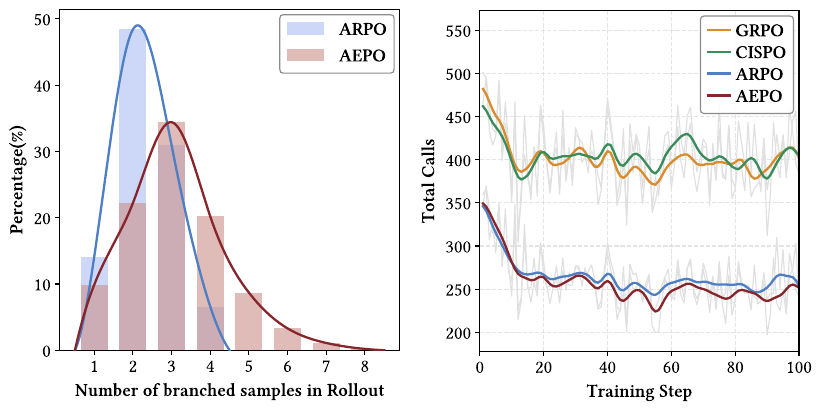}
    \caption{The comparison of branch sampling distribution in rollout (left); The comparison of tool-call efficiency across four RL algorithms (right).}
    \label{fig:rollout_zhuzhuang}
\end{figure}

\subsection{Does AEPO Mitigate Rollout Collapse?}
\paragraph{\textbf{(1) Diversity Analysis}} To investigate whether AEPO's dynamic entropy-balanced rollout improves sampling diversity, we follow the setup of the preliminary experiment (§\ref{sec:pre-exp}) and randomly selected samples from 10 rollout steps, encompassing 640 distinct problems and approximately 7.6$k$ trajectories. We further employ BGEM3~\citep{chen2024bge} as the semantic embedding model, applied the PCA method for dimensionality reduction, and used DBSCAN~\citep{dbscan} for clustering to visualize the representation of rollout sampling.

As shown in Figure~\ref{fig:cluster}, the results indicate that compared to ARPO, AEPO's sampling trajectories form more distinct cluster centers (54 vs. 62) and exhibit tighter intra-cluster distances with larger inter-cluster gaps. This demonstrates that AEPO improves the scope of rollout diversity and provides clearer differentiation in the sampling path distribution. We attribute this to AEPO's entropy pre-monitoring and continuous entropy penalty branches, which effectively address the continuity of high-entropy branches to achieve comprehensive coverage of the problem-solving space.

\paragraph{\textbf{(2) Statistics Analysis}} To quantitatively analyze AEPO's effectiveness in addressing rollout collapse, we measure the branch distribution of ARPO and AEPO over 10 steps during rollout. As shown in Figure~\ref{fig:rollout_zhuzhuang} (left), with both the global and partial branch sampling budgets set to 8, ARPO typically branches into 2-3 trajectories. In contrast, AEPO exhibits a more diverse branching pattern, potentially covering all 8 paths with different branches. This highlights AEPO's dynamic resource allocation and continuous branch penalty mechanism enable the model to explore potential high-entropy tool-call steps across different trajectories, effectively mitigating bias in specific path branches.

\subsection{Does AEPO Achieve Entropy-Balanced and Efficient RL Training?}

\paragraph{\textbf{(1) Tool-call Efficiency Analysis}} In agentic RL training, effectively controlling the frequency of tool usage can significantly reduce financial costs. To confirm the efficiency of AEPO's tool usage, we quantify the tool consumption of AEPO compared to other RL algorithms in the deep information seeking task. As shown in Figure~\ref{fig:rollout_zhuzhuang} (right), AEPO requires only about half the number of tool calls to achieve superior performance compared to vanilla and clipping-optimized RL algorithms. Additionally, compared to the agentic RL algorithm ARPO, AEPO consistently reduces the number of tool calls. We attribute this enhanced efficiency to the entropy pre-monitoring phase, which balances the allocation of rollout exploration resources based on the information gain from the problem and tool usage. This ensures that AEPO not only broadens the rollout exploration space but also achieves efficient web agent training.

\paragraph{\textbf{(2) Entropy Stability Analysis}} To better quantify the effectiveness of entropy-balanced policy optimization during policy updates, we present the RL training curves for 10 reasoning tasks. Figure~\ref{fig:dynamic} illustrates the dynamic visualization of entropy loss and validation set accuracy across 10 reasoning tasks throughout the training steps. We observe that using clipping-optimized RL often encounters entropy instability during training, leading to performance collapse. In contrast, AEPO demonstrates a more stable entropy curve compared to other reinforcement learning algorithms. Interestingly, sharp fluctuations in entropy loss do not improve training stability and effectiveness. Instead, maintaining a consistently high and stable entropy dynamic is generally advantageous for ongoing performance enhancement. This observation supports our initial motivation, as AEPO employs entropy-balanced policy optimizations to foster more reasonable and stable entropy dynamics.


\begin{figure}[t]
    \centering
    \setlength{\abovecaptionskip}{0.2cm}
\setlength{\belowcaptionskip}{0.2cm}
    \includegraphics[width=1\linewidth]{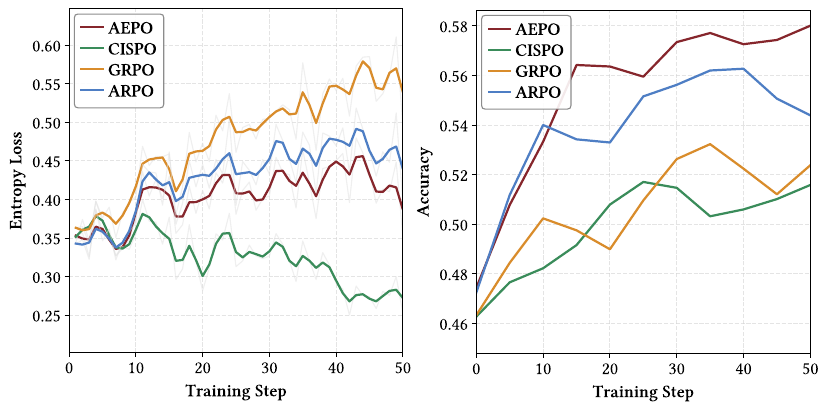}
    \caption{Visualization of training dynamics, including entropy loss(left) and accuracy (right) across training steps}
    \label{fig:dynamic}
\end{figure}

\section{Related Work}
\subsection{Reinforcement Learning for Web Agent.}

The emergence of agent reinforcement learning (RL)\citep{zhang2025landscape} has set the stage for the development of general-purpose web agents, a pursuit shared by both academia and industry. Initial efforts\citep{searchr1,r1searcher,chen2025research,retool,torl} established a foundation by enabling models to autonomously interact with search engines or code interpreters using rule-based RL. Building on this groundwork, subsequent innovations have emerged: Tool-star~\citep{dong2025toolstar} incorporates multi-tool usage within agentic RL, while other studies~\citep{wang2025otc,Tool-Use,r1_searcher++,chen2025Toollight,ReinforcementLearningwithRubricAnchors,Atom-Searcher,Fin-PRM,tan2025hiersearch,xiao2025limi} enhance efficiency and stability through redesigned reward functions. MemAgent introduces memory mechanisms during the RL phase to better manage contextual information~\citep{MemAgent}. Additionally, recent research~\citep{xue2025simpletir,jiang2025verltool,asearcher,Chain-of-Agents} explores comprehensive asynchronous training frameworks for web agents. Building on these advancements, Tongyi Deep Research~\citep{wu2025resumun,su2025ScalingAgentsViaContinualPre-training,fang2025TowardsGeneralAgenticIntelligenceViaEnvironmentScaling,li2025websailor-v2,qiao2025webresearcher,WebShaper,WebDancer,EvolveSearch} aims to fully leverage the post-training paradigm. This includes data synthesis, RL algorithm optimization, and report generation, thereby broadening the scope of web agent training. To minimize resource consumption during training, another line of research seeks to simulate search engines using the generative capabilities of large models for self-alignment~\citep{SSRL,zerosearch}.

Recently, agentic RL methods~\citep{TreeRL,dong2025arpo,GIGPO,li2025treepobridginggappolicy} have focused on optimizing foundational RL algorithms for web agents, employing tree-structured rollouts for autonomous branch sampling under high entropy. While these methods have advanced web agent training, they often overlook the challenges posed by high-entropy tokens. Several single-turn RL studies~\citep{su2025gppo,zhaoxin_entropy,zheng2025returnentropyelicitingexplore,liu2025part} have emphasized that stable entropy training is crucial for enhancing model performance. However, this aspect remains largely unexplored in multi-turn agentic RL. In this paper, we introduce AEPO to achieve entropy-balanced web agent RL training.

\subsection{Agentic Reinforcement Learning.}

Reinforcement learning (RL) plays a crucial role in helping large language model (LLM) agents adapt to dynamic and open environments~\citep{AgentRewardBench,shridhar2020alfworld,GAIA}. Foundational studies such as DQN~\citep{dqn} and AlphaZero~\citep{alphazero} have shown that self-play-based RL can endow agents with skills ranging from natural language understanding to strategic gameplay~\citep{textgame_rl}. Building on these foundations, value-based RL methods have been applied to improve embodied intelligence in hardware control and complex gaming tasks~\citep{agentrl_2,agentrl_3,agent_rl_4,agent_rl_5,ppo,Advantage-Weighted-Regression}. Recent advancements, like RAGEN~\citep{ragen,zhou2024archer}, incorporate reasoning states and environmental interactions into turn-level responses using trajectory-level RL. To enhance tool-integrated reasoning, several studies~\citep{searchr1,retool,r1searcher,searchr1,chen2025research,retool,torl,zerosearch,webthinker,2505_ARTIST} utilize rule-based RL to enable LLMs to autonomously invoke external tools (e.g., search engines, Python compilers) to improve reasoning accuracy. Further research, including ToolRL~\citep{qian2025toolrl}, Tool-Star~\citep{dong2025toolstar}, and OTC~\citep{wang2025otc}, explores the integration of multiple tools and enhances tool-use efficiency. Efforts by Kimi Deepresearcher~\footnote{\url{https://moonshotai.github.io/Kimi-Researcher/}} and Websailor~\citep{li2025websailor} focus on optimizing RL algorithms to better handle deepsearch's long context scenarios. With the surge in reasoning capabilities of Multimodal large language models (MLLMs), several works have effectively broadened the scope of this field by combining agentic RL in the multimodal domain with external tools~\citep{qin2025ui,wang2025ui,Qwen2.5-VL,2411_llava_o1,wemath,We-Math2.0,qiao2025v}.

Although many studies enhance tool invocation through reward shaping and rollout mechanisms, trajectory-level RL alone often struggles to effectively capture the multi-turn, long-horizon characteristics of LLM-based agent behavior. This challenge has led to the development of ARPO, which aims to learn step-level tool-use behavior patterns.

\section{Conclusion}
In this paper, we introduce Agentic Entropy-Balanced Policy Optimization (AEPO), an agentic RL algorithm that effectively balances entropy during both rollout and policy update phases. Initially, we quantify two inherent entropy-driven challenges in preliminary experiments. AEPO comprises two core components: (1) a dynamic entropy-balanced rollout mechanism that adaptively allocates the sampling budget between global and branch sampling through entropy pre-monitoring, while imposing a branch penalty on consecutive high-entropy tool-call steps to prevent oversampling; (2) Entropy-Balanced Policy Optimization, which incorporates a stop-gradient operation in the high-entropy clipping term to preserve and rescale gradients on high-entropy tokens, alongside entropy-aware advantage estimation to focus learning on high-uncertainty tokens. Experiments across 14 benchmarks demonstrate that AEPO consistently outperforms seven mainstream agentic RL algorithms. Quantitative analyses confirm its scalability and stability, offering valuable insights for training general web agents.

\bibliographystyle{ACM-Reference-Format}
\bibliography{sample-base}


\begin{thebibliography}{130}


\ifx \showCODEN    \undefined \def \showCODEN     #1{\unskip}     \fi
\ifx \showDOI      \undefined \def \showDOI       #1{#1}\fi
\ifx \showISBNx    \undefined \def \showISBNx     #1{\unskip}     \fi
\ifx \showISBNxiii \undefined \def \showISBNxiii  #1{\unskip}     \fi
\ifx \showISSN     \undefined \def \showISSN      #1{\unskip}     \fi
\ifx \showLCCN     \undefined \def \showLCCN      #1{\unskip}     \fi
\ifx \shownote     \undefined \def \shownote      #1{#1}          \fi
\ifx \showarticletitle \undefined \def \showarticletitle #1{#1}   \fi
\ifx \showURL      \undefined \def \showURL       {\relax}        \fi
\providecommand\bibfield[2]{#2}
\providecommand\bibinfo[2]{#2}
\providecommand\natexlab[1]{#1}
\providecommand\showeprint[2][]{arXiv:#2}

\bibitem[\protect\citeauthoryear{Bai, Zhou, Pan, Cemri, Suhr, Levine, and Kumar}{Bai et~al\mbox{.}}{2024}]%
        {agent_rl_4}
\bibfield{author}{\bibinfo{person}{Hao Bai}, \bibinfo{person}{Yifei Zhou}, \bibinfo{person}{Jiayi Pan}, \bibinfo{person}{Mert Cemri}, \bibinfo{person}{Alane Suhr}, \bibinfo{person}{Sergey Levine}, {and} \bibinfo{person}{Aviral Kumar}.} \bibinfo{year}{2024}\natexlab{}.
\newblock \showarticletitle{DigiRL: Training In-The-Wild Device-Control Agents with Autonomous Reinforcement Learning}. In \bibinfo{booktitle}{\emph{Advances in Neural Information Processing Systems 38: Annual Conference on Neural Information Processing Systems 2024, NeurIPS 2024, Vancouver, BC, Canada, December 10 - 15, 2024}}, \bibfield{editor}{\bibinfo{person}{Amir Globersons}, \bibinfo{person}{Lester Mackey}, \bibinfo{person}{Danielle Belgrave}, \bibinfo{person}{Angela Fan}, \bibinfo{person}{Ulrich Paquet}, \bibinfo{person}{Jakub~M. Tomczak}, {and} \bibinfo{person}{Cheng Zhang}} (Eds.).
\newblock
\urldef\tempurl%
\url{http://papers.nips.cc/paper\_files/paper/2024/hash/1704ddd0bb89f159dfe609b32c889995-Abstract-Conference.html}
\showURL{%
\tempurl}


\bibitem[\protect\citeauthoryear{Bai, Chen, Liu, Wang, Ge, Song, Dang, Wang, Wang, Tang, Zhong, Zhu, Yang, Li, Wan, Wang, Ding, Fu, Xu, Ye, Zhang, Xie, Cheng, Zhang, Yang, Xu, and Lin}{Bai et~al\mbox{.}}{2025}]%
        {Qwen2.5-VL}
\bibfield{author}{\bibinfo{person}{Shuai Bai}, \bibinfo{person}{Keqin Chen}, \bibinfo{person}{Xuejing Liu}, \bibinfo{person}{Jialin Wang}, \bibinfo{person}{Wenbin Ge}, \bibinfo{person}{Sibo Song}, \bibinfo{person}{Kai Dang}, \bibinfo{person}{Peng Wang}, \bibinfo{person}{Shijie Wang}, \bibinfo{person}{Jun Tang}, \bibinfo{person}{Humen Zhong}, \bibinfo{person}{Yuanzhi Zhu}, \bibinfo{person}{Mingkun Yang}, \bibinfo{person}{Zhaohai Li}, \bibinfo{person}{Jianqiang Wan}, \bibinfo{person}{Pengfei Wang}, \bibinfo{person}{Wei Ding}, \bibinfo{person}{Zheren Fu}, \bibinfo{person}{Yiheng Xu}, \bibinfo{person}{Jiabo Ye}, \bibinfo{person}{Xi Zhang}, \bibinfo{person}{Tianbao Xie}, \bibinfo{person}{Zesen Cheng}, \bibinfo{person}{Hang Zhang}, \bibinfo{person}{Zhibo Yang}, \bibinfo{person}{Haiyang Xu}, {and} \bibinfo{person}{Junyang Lin}.} \bibinfo{year}{2025}\natexlab{}.
\newblock \showarticletitle{Qwen2.5-VL Technical Report}.
\newblock \bibinfo{journal}{\emph{arXiv preprint arXiv:2502.13923}} (\bibinfo{year}{2025}).
\newblock


\bibitem[\protect\citeauthoryear{Chen, Li, Gong, Jiang, Fei, Yang, Shan, Yu, Wang, Zhu, Xiao, Du, Zhang, Qiao, Zhang, Du, Guo, Chen, Ding, Sun, Li, Jiao, Zhou, Zhang, Ding, Sun, Feng, Cai, Zhu, Sun, Zhuang, Cai, Song, Zhu, Li, Tian, Liu, Xu, Yan, Liu, He, Feng, Yang, Xiao, Han, Wang, Yu, Feng, Li, Zheng, Du, Yang, Zeng, Yu, Tao, Chi, Zhang, Lin, Hu, Di, Gao, Li, Zhao, Ren, Xu, Li, Wang, Tian, Leng, Chen, Chen, Shi, Weng, Guan, Yu, Li, Zhu, Li, Cai, Liang, Cheng, Kong, Li, Chen, Song, Luo, Su, Li, Han, Hou, Lu, Zou, Shen, Gong, Ma, Wang, Shi, Zhong, and Duan}{Chen et~al\mbox{.}}{2025c}]%
        {MiniMax-M1}
\bibfield{author}{\bibinfo{person}{Aili Chen}, \bibinfo{person}{Aonian Li}, \bibinfo{person}{Bangwei Gong}, \bibinfo{person}{Binyang Jiang}, \bibinfo{person}{Bo Fei}, \bibinfo{person}{Bo Yang}, \bibinfo{person}{Boji Shan}, \bibinfo{person}{Changqing Yu}, \bibinfo{person}{Chao Wang}, \bibinfo{person}{Cheng Zhu}, \bibinfo{person}{Chengjun Xiao}, \bibinfo{person}{Chengyu Du}, \bibinfo{person}{Chi Zhang}, \bibinfo{person}{Chu Qiao}, \bibinfo{person}{Chunhao Zhang}, \bibinfo{person}{Chunhui Du}, \bibinfo{person}{Congchao Guo}, \bibinfo{person}{Da Chen}, \bibinfo{person}{Deming Ding}, \bibinfo{person}{Dianjun Sun}, \bibinfo{person}{Dong Li}, \bibinfo{person}{Enwei Jiao}, \bibinfo{person}{Haigang Zhou}, \bibinfo{person}{Haimo Zhang}, \bibinfo{person}{Han Ding}, \bibinfo{person}{Haohai Sun}, \bibinfo{person}{Haoyu Feng}, \bibinfo{person}{Huaiguang Cai}, \bibinfo{person}{Haichao Zhu}, \bibinfo{person}{Jian Sun}, \bibinfo{person}{Jiaqi Zhuang}, \bibinfo{person}{Jiaren Cai}, \bibinfo{person}{Jiayuan Song},
  \bibinfo{person}{Jin Zhu}, \bibinfo{person}{Jingyang Li}, \bibinfo{person}{Jinhao Tian}, \bibinfo{person}{Jinli Liu}, \bibinfo{person}{Junhao Xu}, \bibinfo{person}{Junjie Yan}, \bibinfo{person}{Junteng Liu}, \bibinfo{person}{Junxian He}, \bibinfo{person}{Kaiyi Feng}, \bibinfo{person}{Ke Yang}, \bibinfo{person}{Kecheng Xiao}, \bibinfo{person}{Le Han}, \bibinfo{person}{Leyang Wang}, \bibinfo{person}{Lianfei Yu}, \bibinfo{person}{Liheng Feng}, \bibinfo{person}{Lin Li}, \bibinfo{person}{Lin Zheng}, \bibinfo{person}{Linge Du}, \bibinfo{person}{Lingyu Yang}, \bibinfo{person}{Lunbin Zeng}, \bibinfo{person}{Minghui Yu}, \bibinfo{person}{Mingliang Tao}, \bibinfo{person}{Mingyuan Chi}, \bibinfo{person}{Mozhi Zhang}, \bibinfo{person}{Mujie Lin}, \bibinfo{person}{Nan Hu}, \bibinfo{person}{Nongyu Di}, \bibinfo{person}{Peng Gao}, \bibinfo{person}{Pengfei Li}, \bibinfo{person}{Pengyu Zhao}, \bibinfo{person}{Qibing Ren}, \bibinfo{person}{Qidi Xu}, \bibinfo{person}{Qile Li}, \bibinfo{person}{Qin Wang}, \bibinfo{person}{Rong
  Tian}, \bibinfo{person}{Ruitao Leng}, \bibinfo{person}{Shaoxiang Chen}, \bibinfo{person}{Shaoyu Chen}, \bibinfo{person}{Shengmin Shi}, \bibinfo{person}{Shitong Weng}, \bibinfo{person}{Shuchang Guan}, \bibinfo{person}{Shuqi Yu}, \bibinfo{person}{Sichen Li}, \bibinfo{person}{Songquan Zhu}, \bibinfo{person}{Tengfei Li}, \bibinfo{person}{Tianchi Cai}, \bibinfo{person}{Tianrun Liang}, \bibinfo{person}{Weiyu Cheng}, \bibinfo{person}{Weize Kong}, \bibinfo{person}{Wenkai Li}, \bibinfo{person}{Xiancai Chen}, \bibinfo{person}{Xiangjun Song}, \bibinfo{person}{Xiao Luo}, \bibinfo{person}{Xiao Su}, \bibinfo{person}{Xiaobo Li}, \bibinfo{person}{Xiaodong Han}, \bibinfo{person}{Xinzhu Hou}, \bibinfo{person}{Xuan Lu}, \bibinfo{person}{Xun Zou}, \bibinfo{person}{Xuyang Shen}, \bibinfo{person}{Yan Gong}, \bibinfo{person}{Yan Ma}, \bibinfo{person}{Yang Wang}, \bibinfo{person}{Yiqi Shi}, \bibinfo{person}{Yiran Zhong}, {and} \bibinfo{person}{Yonghong Duan}.} \bibinfo{year}{2025}\natexlab{c}.
\newblock \showarticletitle{MiniMax-M1: Scaling Test-Time Compute Efficiently with Lightning Attention}.
\newblock \bibinfo{journal}{\emph{CoRR}}  \bibinfo{volume}{abs/2506.13585} (\bibinfo{year}{2025}).
\newblock
\urldef\tempurl%
\url{https://doi.org/10.48550/ARXIV.2506.13585}
\showDOI{\tempurl}
\showeprint[arXiv]{2506.13585}


\bibitem[\protect\citeauthoryear{Chen, Xiao, Zhang, Luo, Lian, and Liu}{Chen et~al\mbox{.}}{2024}]%
        {chen2024bge}
\bibfield{author}{\bibinfo{person}{Jianlv Chen}, \bibinfo{person}{Shitao Xiao}, \bibinfo{person}{Peitian Zhang}, \bibinfo{person}{Kun Luo}, \bibinfo{person}{Defu Lian}, {and} \bibinfo{person}{Zheng Liu}.} \bibinfo{year}{2024}\natexlab{}.
\newblock \showarticletitle{{BGE} M3-Embedding: Multi-Lingual, Multi-Functionality, Multi-Granularity Text Embeddings Through Self-Knowledge Distillation}.
\newblock \bibinfo{journal}{\emph{CoRR}}  \bibinfo{volume}{abs/2402.03216} (\bibinfo{year}{2024}).
\newblock
\urldef\tempurl%
\url{https://doi.org/10.48550/ARXIV.2402.03216}
\showDOI{\tempurl}
\showeprint[arXiv]{2402.03216}


\bibitem[\protect\citeauthoryear{Chen, Ren, Liu, Hu, Tian, Xie, Liu, Zhang, Liu, Gong, et~al\mbox{.}}{Chen et~al\mbox{.}}{2025e}]%
        {chen2025xbench}
\bibfield{author}{\bibinfo{person}{Kaiyuan Chen}, \bibinfo{person}{Yixin Ren}, \bibinfo{person}{Yang Liu}, \bibinfo{person}{Xiaobo Hu}, \bibinfo{person}{Haotong Tian}, \bibinfo{person}{Tianbao Xie}, \bibinfo{person}{Fangfu Liu}, \bibinfo{person}{Haoye Zhang}, \bibinfo{person}{Hongzhang Liu}, \bibinfo{person}{Yuan Gong}, {et~al\mbox{.}}} \bibinfo{year}{2025}\natexlab{e}.
\newblock \showarticletitle{xbench: Tracking Agents Productivity Scaling with Profession-Aligned Real-World Evaluations}.
\newblock \bibinfo{journal}{\emph{arXiv preprint arXiv:2506.13651}} (\bibinfo{year}{2025}).
\newblock


\bibitem[\protect\citeauthoryear{Chen, Li, Sun, Zhou, Zhu, Wang, Pan, Zhang, Chen, Yang, Zhou, and Chen}{Chen et~al\mbox{.}}{2025d}]%
        {chen2025research}
\bibfield{author}{\bibinfo{person}{Mingyang Chen}, \bibinfo{person}{Tianpeng Li}, \bibinfo{person}{Haoze Sun}, \bibinfo{person}{Yijie Zhou}, \bibinfo{person}{Chenzheng Zhu}, \bibinfo{person}{Haofen Wang}, \bibinfo{person}{Jeff~Z. Pan}, \bibinfo{person}{Wen Zhang}, \bibinfo{person}{Huajun Chen}, \bibinfo{person}{Fan Yang}, \bibinfo{person}{Zenan Zhou}, {and} \bibinfo{person}{Weipeng Chen}.} \bibinfo{year}{2025}\natexlab{d}.
\newblock \bibinfo{title}{ReSearch: Learning to Reason with Search for LLMs via Reinforcement Learning}.
\newblock
\newblock
\showeprint[arxiv]{cs.AI/2503.19470}
\urldef\tempurl%
\url{https://arxiv.org/abs/2503.19470}
\showURL{%
\tempurl}


\bibitem[\protect\citeauthoryear{Chen, Dong, and Dou}{Chen et~al\mbox{.}}{2025a}]%
        {chen2025Toollight}
\bibfield{author}{\bibinfo{person}{Yifei Chen}, \bibinfo{person}{Guanting Dong}, {and} \bibinfo{person}{Zhicheng Dou}.} \bibinfo{year}{2025}\natexlab{a}.
\newblock \showarticletitle{Toward Effective Tool-Integrated Reasoning via Self-Evolved Preference Learning}.
\newblock \bibinfo{journal}{\emph{arXiv preprint arXiv:2509.23285}} (\bibinfo{year}{2025}).
\newblock


\bibitem[\protect\citeauthoryear{Chen, Dong, Zhu, and Dou}{Chen et~al\mbox{.}}{2025b}]%
        {chen2025revisiting}
\bibfield{author}{\bibinfo{person}{Yifei Chen}, \bibinfo{person}{Guanting Dong}, \bibinfo{person}{Yutao Zhu}, {and} \bibinfo{person}{Zhicheng Dou}.} \bibinfo{year}{2025}\natexlab{b}.
\newblock \showarticletitle{Revisiting RAG Ensemble: A Theoretical and Mechanistic Analysis of Multi-RAG System Collaboration}.
\newblock \bibinfo{journal}{\emph{arXiv preprint arXiv:2508.13828}} (\bibinfo{year}{2025}).
\newblock


\bibitem[\protect\citeauthoryear{Cheng, Huang, Zhu, Dai, Zhao, Zhang, and Wei}{Cheng et~al\mbox{.}}{2025}]%
        {zhaoxin_entropy}
\bibfield{author}{\bibinfo{person}{Daixuan Cheng}, \bibinfo{person}{Shaohan Huang}, \bibinfo{person}{Xuekai Zhu}, \bibinfo{person}{Bo Dai}, \bibinfo{person}{Wayne~Xin Zhao}, \bibinfo{person}{Zhenliang Zhang}, {and} \bibinfo{person}{Furu Wei}.} \bibinfo{year}{2025}\natexlab{}.
\newblock \showarticletitle{Reasoning with Exploration: An Entropy Perspective}.
\newblock \bibinfo{journal}{\emph{CoRR}}  \bibinfo{volume}{abs/2506.14758} (\bibinfo{year}{2025}).
\newblock
\urldef\tempurl%
\url{https://doi.org/10.48550/ARXIV.2506.14758}
\showDOI{\tempurl}
\showeprint[arXiv]{2506.14758}


\bibitem[\protect\citeauthoryear{Chu, Zhai, Yang, Tong, Xie, Schuurmans, Le, Levine, and Ma}{Chu et~al\mbox{.}}{2025}]%
        {sft_memory}
\bibfield{author}{\bibinfo{person}{Tianzhe Chu}, \bibinfo{person}{Yuexiang Zhai}, \bibinfo{person}{Jihan Yang}, \bibinfo{person}{Shengbang Tong}, \bibinfo{person}{Saining Xie}, \bibinfo{person}{Dale Schuurmans}, \bibinfo{person}{Quoc~V. Le}, \bibinfo{person}{Sergey Levine}, {and} \bibinfo{person}{Yi Ma}.} \bibinfo{year}{2025}\natexlab{}.
\newblock \showarticletitle{{SFT} Memorizes, {RL} Generalizes: {A} Comparative Study of Foundation Model Post-training}.
\newblock \bibinfo{journal}{\emph{CoRR}}  \bibinfo{volume}{abs/2501.17161} (\bibinfo{year}{2025}).
\newblock
\urldef\tempurl%
\url{https://doi.org/10.48550/ARXIV.2501.17161}
\showDOI{\tempurl}
\showeprint[arXiv]{2501.17161}


\bibitem[\protect\citeauthoryear{Cobbe, Kosaraju, Bavarian, Chen, Jun, Kaiser, Plappert, Tworek, Hilton, Nakano, Hesse, and Schulman}{Cobbe et~al\mbox{.}}{2021}]%
        {cobbe2021gsm8k}
\bibfield{author}{\bibinfo{person}{Karl Cobbe}, \bibinfo{person}{Vineet Kosaraju}, \bibinfo{person}{Mohammad Bavarian}, \bibinfo{person}{Mark Chen}, \bibinfo{person}{Heewoo Jun}, \bibinfo{person}{Lukasz Kaiser}, \bibinfo{person}{Matthias Plappert}, \bibinfo{person}{Jerry Tworek}, \bibinfo{person}{Jacob Hilton}, \bibinfo{person}{Reiichiro Nakano}, \bibinfo{person}{Christopher Hesse}, {and} \bibinfo{person}{John Schulman}.} \bibinfo{year}{2021}\natexlab{}.
\newblock \showarticletitle{Training Verifiers to Solve Math Word Problems}.
\newblock \bibinfo{journal}{\emph{arXiv preprint arXiv:2110.14168}} (\bibinfo{year}{2021}).
\newblock


\bibitem[\protect\citeauthoryear{Deng, Wang, Ying, Wu, Lin, Xiong, Dai, Yang, Zhang, Wang, Qin, Wang, Zha, Dai, and Meng}{Deng et~al\mbox{.}}{2025}]%
        {Atom-Searcher}
\bibfield{author}{\bibinfo{person}{Yong Deng}, \bibinfo{person}{Guoqing Wang}, \bibinfo{person}{Zhenzhe Ying}, \bibinfo{person}{Xiaofeng Wu}, \bibinfo{person}{Jinzhen Lin}, \bibinfo{person}{Wenwen Xiong}, \bibinfo{person}{Yuqin Dai}, \bibinfo{person}{Shuo Yang}, \bibinfo{person}{Zhanwei Zhang}, \bibinfo{person}{Qiwen Wang}, \bibinfo{person}{Yang Qin}, \bibinfo{person}{Yuan Wang}, \bibinfo{person}{Quanxing Zha}, \bibinfo{person}{Sunhao Dai}, {and} \bibinfo{person}{Changhua Meng}.} \bibinfo{year}{2025}\natexlab{}.
\newblock \showarticletitle{Atom-Searcher: Enhancing Agentic Deep Research via Fine-Grained Atomic Thought Reward}.
\newblock \bibinfo{journal}{\emph{CoRR}}  \bibinfo{volume}{abs/2508.12800} (\bibinfo{year}{2025}).
\newblock
\urldef\tempurl%
\url{https://doi.org/10.48550/ARXIV.2508.12800}
\showDOI{\tempurl}
\showeprint[arXiv]{2508.12800}


\bibitem[\protect\citeauthoryear{Dong, Chen, Li, Jin, Qian, Zhu, Mao, Zhou, Dou, and Wen}{Dong et~al\mbox{.}}{2025a}]%
        {dong2025toolstar}
\bibfield{author}{\bibinfo{person}{Guanting Dong}, \bibinfo{person}{Yifei Chen}, \bibinfo{person}{Xiaoxi Li}, \bibinfo{person}{Jiajie Jin}, \bibinfo{person}{Hongjin Qian}, \bibinfo{person}{Yutao Zhu}, \bibinfo{person}{Hangyu Mao}, \bibinfo{person}{Guorui Zhou}, \bibinfo{person}{Zhicheng Dou}, {and} \bibinfo{person}{Ji{-}Rong Wen}.} \bibinfo{year}{2025}\natexlab{a}.
\newblock \showarticletitle{Tool-Star: Empowering LLM-Brained Multi-Tool Reasoner via Reinforcement Learning}.
\newblock \bibinfo{journal}{\emph{CoRR}}  \bibinfo{volume}{abs/2505.16410} (\bibinfo{year}{2025}).
\newblock
\urldef\tempurl%
\url{https://doi.org/10.48550/ARXIV.2505.16410}
\showDOI{\tempurl}
\showeprint[arXiv]{2505.16410}


\bibitem[\protect\citeauthoryear{Dong, Mao, Ma, Bao, Chen, Wang, Chen, Du, Wang, Zhang, Zhou, Zhu, Wen, and Dou}{Dong et~al\mbox{.}}{2025b}]%
        {dong2025arpo}
\bibfield{author}{\bibinfo{person}{Guanting Dong}, \bibinfo{person}{Hangyu Mao}, \bibinfo{person}{Kai Ma}, \bibinfo{person}{Licheng Bao}, \bibinfo{person}{Yifei Chen}, \bibinfo{person}{Zhongyuan Wang}, \bibinfo{person}{Zhongxia Chen}, \bibinfo{person}{Jiazhen Du}, \bibinfo{person}{Huiyang Wang}, \bibinfo{person}{Fuzheng Zhang}, \bibinfo{person}{Guorui Zhou}, \bibinfo{person}{Yutao Zhu}, \bibinfo{person}{Ji{-}Rong Wen}, {and} \bibinfo{person}{Zhicheng Dou}.} \bibinfo{year}{2025}\natexlab{b}.
\newblock \showarticletitle{Agentic Reinforced Policy Optimization}.
\newblock \bibinfo{journal}{\emph{CoRR}}  \bibinfo{volume}{abs/2507.19849} (\bibinfo{year}{2025}).
\newblock
\urldef\tempurl%
\url{https://doi.org/10.48550/ARXIV.2507.19849}
\showDOI{\tempurl}
\showeprint[arXiv]{2507.19849}


\bibitem[\protect\citeauthoryear{Dong, Song, Zhu, Qiao, Dou, and Wen}{Dong et~al\mbox{.}}{2025c}]%
        {vifrag}
\bibfield{author}{\bibinfo{person}{Guanting Dong}, \bibinfo{person}{Xiaoshuai Song}, \bibinfo{person}{Yutao Zhu}, \bibinfo{person}{Runqi Qiao}, \bibinfo{person}{Zhicheng Dou}, {and} \bibinfo{person}{Ji{-}Rong Wen}.} \bibinfo{year}{2025}\natexlab{c}.
\newblock \showarticletitle{Toward Verifiable Instruction-Following Alignment for Retrieval Augmented Generation}. In \bibinfo{booktitle}{\emph{AAAI-25, Sponsored by the Association for the Advancement of Artificial Intelligence, February 25 - March 4, 2025, Philadelphia, PA, {USA}}}, \bibfield{editor}{\bibinfo{person}{Toby Walsh}, \bibinfo{person}{Julie Shah}, {and} \bibinfo{person}{Zico Kolter}} (Eds.). \bibinfo{publisher}{{AAAI} Press}, \bibinfo{pages}{23796--23804}.
\newblock
\urldef\tempurl%
\url{https://doi.org/10.1609/AAAI.V39I22.34551}
\showDOI{\tempurl}


\bibitem[\protect\citeauthoryear{Dong, Zhang, Deng, Zhu, Dou, and Wen}{Dong et~al\mbox{.}}{2024a}]%
        {2412_AR_MCTS}
\bibfield{author}{\bibinfo{person}{Guanting Dong}, \bibinfo{person}{Chenghao Zhang}, \bibinfo{person}{Mengjie Deng}, \bibinfo{person}{Yutao Zhu}, \bibinfo{person}{Zhicheng Dou}, {and} \bibinfo{person}{Ji-Rong Wen}.} \bibinfo{year}{2024}\natexlab{a}.
\newblock \showarticletitle{Progressive Multimodal Reasoning via Active Retrieval}.
\newblock \bibinfo{journal}{\emph{arXiv preprint arXiv:2412.14835}} (\bibinfo{year}{2024}).
\newblock


\bibitem[\protect\citeauthoryear{Dong, Zhu, Zhang, Wang, Dou, and Wen}{Dong et~al\mbox{.}}{2024b}]%
        {dparag}
\bibfield{author}{\bibinfo{person}{Guanting Dong}, \bibinfo{person}{Yutao Zhu}, \bibinfo{person}{Chenghao Zhang}, \bibinfo{person}{Zechen Wang}, \bibinfo{person}{Zhicheng Dou}, {and} \bibinfo{person}{Ji{-}Rong Wen}.} \bibinfo{year}{2024}\natexlab{b}.
\newblock \showarticletitle{Understand What {LLM} Needs: Dual Preference Alignment for Retrieval-Augmented Generation}.
\newblock \bibinfo{journal}{\emph{CoRR}}  \bibinfo{volume}{abs/2406.18676} (\bibinfo{year}{2024}).
\newblock
\urldef\tempurl%
\url{https://doi.org/10.48550/ARXIV.2406.18676}
\showDOI{\tempurl}
\showeprint[arXiv]{2406.18676}


\bibitem[\protect\citeauthoryear{Dubey, Jauhri, Pandey, Kadian, Al-Dahle, Letman, Mathur, Schelten, Yang, Fan, et~al\mbox{.}}{Dubey et~al\mbox{.}}{2024}]%
        {llama3}
\bibfield{author}{\bibinfo{person}{Abhimanyu Dubey}, \bibinfo{person}{Abhinav Jauhri}, \bibinfo{person}{Abhinav Pandey}, \bibinfo{person}{Abhishek Kadian}, \bibinfo{person}{Ahmad Al-Dahle}, \bibinfo{person}{Aiesha Letman}, \bibinfo{person}{Akhil Mathur}, \bibinfo{person}{Alan Schelten}, \bibinfo{person}{Amy Yang}, \bibinfo{person}{Angela Fan}, {et~al\mbox{.}}} \bibinfo{year}{2024}\natexlab{}.
\newblock \showarticletitle{The llama 3 herd of models}.
\newblock \bibinfo{journal}{\emph{arXiv preprint arXiv:2407.21783}} (\bibinfo{year}{2024}).
\newblock


\bibitem[\protect\citeauthoryear{Ester, Kriegel, Sander, and Xu}{Ester et~al\mbox{.}}{1996}]%
        {dbscan}
\bibfield{author}{\bibinfo{person}{Martin Ester}, \bibinfo{person}{Hans-Peter Kriegel}, \bibinfo{person}{J\"{o}rg Sander}, {and} \bibinfo{person}{Xiaowei Xu}.} \bibinfo{year}{1996}\natexlab{}.
\newblock \showarticletitle{A density-based algorithm for discovering clusters in large spatial databases with noise}. In \bibinfo{booktitle}{\emph{Proceedings of the Second International Conference on Knowledge Discovery and Data Mining}} \emph{(\bibinfo{series}{KDD'96})}. \bibinfo{publisher}{AAAI Press}, \bibinfo{pages}{226–231}.
\newblock


\bibitem[\protect\citeauthoryear{Fan, Zhang, Zhou, Zuo, Chen, Fu, Long, Zhu, Jiang, Zhang, Kang, Chen, Huang, He, Wang, Bai, Ding, and Zhou}{Fan et~al\mbox{.}}{2025}]%
        {SSRL}
\bibfield{author}{\bibinfo{person}{Yuchen Fan}, \bibinfo{person}{Kaiyan Zhang}, \bibinfo{person}{Heng Zhou}, \bibinfo{person}{Yuxin Zuo}, \bibinfo{person}{Yanxu Chen}, \bibinfo{person}{Yu Fu}, \bibinfo{person}{Xinwei Long}, \bibinfo{person}{Xuekai Zhu}, \bibinfo{person}{Che Jiang}, \bibinfo{person}{Yuchen Zhang}, \bibinfo{person}{Li Kang}, \bibinfo{person}{Gang Chen}, \bibinfo{person}{Cheng Huang}, \bibinfo{person}{Zhizhou He}, \bibinfo{person}{Bingning Wang}, \bibinfo{person}{Lei Bai}, \bibinfo{person}{Ning Ding}, {and} \bibinfo{person}{Bowen Zhou}.} \bibinfo{year}{2025}\natexlab{}.
\newblock \showarticletitle{{SSRL:} Self-Search Reinforcement Learning}.
\newblock \bibinfo{journal}{\emph{CoRR}}  \bibinfo{volume}{abs/2508.10874} (\bibinfo{year}{2025}).
\newblock
\urldef\tempurl%
\url{https://doi.org/10.48550/ARXIV.2508.10874}
\showDOI{\tempurl}
\showeprint[arXiv]{2508.10874}


\bibitem[\protect\citeauthoryear{Fang, Cai, Li, Wu, Li, Yin, Wang, Wang, Su, Zhang, et~al\mbox{.}}{Fang et~al\mbox{.}}{2025a}]%
        {fang2025TowardsGeneralAgenticIntelligenceViaEnvironmentScaling}
\bibfield{author}{\bibinfo{person}{Runnan Fang}, \bibinfo{person}{Shihao Cai}, \bibinfo{person}{Baixuan Li}, \bibinfo{person}{Jialong Wu}, \bibinfo{person}{Guangyu Li}, \bibinfo{person}{Wenbiao Yin}, \bibinfo{person}{Xinyu Wang}, \bibinfo{person}{Xiaobin Wang}, \bibinfo{person}{Liangcai Su}, \bibinfo{person}{Zhen Zhang}, {et~al\mbox{.}}} \bibinfo{year}{2025}\natexlab{a}.
\newblock \showarticletitle{Towards General Agentic Intelligence via Environment Scaling}.
\newblock \bibinfo{journal}{\emph{arXiv preprint arXiv:2509.13311}} (\bibinfo{year}{2025}).
\newblock


\bibitem[\protect\citeauthoryear{Fang, Zhang, Zhang, Ma, Yu, Mi, and Yu}{Fang et~al\mbox{.}}{2025c}]%
        {WebEvolver}
\bibfield{author}{\bibinfo{person}{Tianqing Fang}, \bibinfo{person}{Hongming Zhang}, \bibinfo{person}{Zhisong Zhang}, \bibinfo{person}{Kaixin Ma}, \bibinfo{person}{Wenhao Yu}, \bibinfo{person}{Haitao Mi}, {and} \bibinfo{person}{Dong Yu}.} \bibinfo{year}{2025}\natexlab{c}.
\newblock \showarticletitle{WebEvolver: Enhancing Web Agent Self-Improvement with Coevolving World Model}.
\newblock \bibinfo{journal}{\emph{CoRR}}  \bibinfo{volume}{abs/2504.21024} (\bibinfo{year}{2025}).
\newblock
\urldef\tempurl%
\url{https://doi.org/10.48550/ARXIV.2504.21024}
\showDOI{\tempurl}
\showeprint[arXiv]{2504.21024}


\bibitem[\protect\citeauthoryear{Fang, Zhang, Wang, Wang, Qin, Wan, Ma, Zhang, Chen, Li, Zhang, Mi, and Yu}{Fang et~al\mbox{.}}{2025b}]%
        {CognitiveKernelPro}
\bibfield{author}{\bibinfo{person}{Tianqing Fang}, \bibinfo{person}{Zhisong Zhang}, \bibinfo{person}{Xiaoyang Wang}, \bibinfo{person}{Rui Wang}, \bibinfo{person}{Can Qin}, \bibinfo{person}{Yuxuan Wan}, \bibinfo{person}{Jun{-}Yu Ma}, \bibinfo{person}{Ce Zhang}, \bibinfo{person}{Jiaqi Chen}, \bibinfo{person}{Xiyun Li}, \bibinfo{person}{Hongming Zhang}, \bibinfo{person}{Haitao Mi}, {and} \bibinfo{person}{Dong Yu}.} \bibinfo{year}{2025}\natexlab{b}.
\newblock \showarticletitle{Cognitive Kernel-Pro: {A} Framework for Deep Research Agents and Agent Foundation Models Training}.
\newblock \bibinfo{journal}{\emph{CoRR}}  \bibinfo{volume}{abs/2508.00414} (\bibinfo{year}{2025}).
\newblock
\urldef\tempurl%
\url{https://doi.org/10.48550/ARXIV.2508.00414}
\showDOI{\tempurl}
\showeprint[arXiv]{2508.00414}


\bibitem[\protect\citeauthoryear{Feng, Huang, Qu, Zhang, Qin, Zhong, Jiang, Chi, and Zhong}{Feng et~al\mbox{.}}{2025a}]%
        {retool}
\bibfield{author}{\bibinfo{person}{Jiazhan Feng}, \bibinfo{person}{Shijue Huang}, \bibinfo{person}{Xingwei Qu}, \bibinfo{person}{Ge Zhang}, \bibinfo{person}{Yujia Qin}, \bibinfo{person}{Baoquan Zhong}, \bibinfo{person}{Chengquan Jiang}, \bibinfo{person}{Jinxin Chi}, {and} \bibinfo{person}{Wanjun Zhong}.} \bibinfo{year}{2025}\natexlab{a}.
\newblock \bibinfo{title}{ReTool: Reinforcement Learning for Strategic Tool Use in LLMs}.
\newblock
\newblock
\showeprint[arxiv]{cs.CL/2504.11536}
\urldef\tempurl%
\url{https://arxiv.org/abs/2504.11536}
\showURL{%
\tempurl}


\bibitem[\protect\citeauthoryear{Feng, Xue, Liu, and An}{Feng et~al\mbox{.}}{2025b}]%
        {Group-in-Group}
\bibfield{author}{\bibinfo{person}{Lang Feng}, \bibinfo{person}{Zhenghai Xue}, \bibinfo{person}{Tingcong Liu}, {and} \bibinfo{person}{Bo An}.} \bibinfo{year}{2025}\natexlab{b}.
\newblock \showarticletitle{Group-in-Group Policy Optimization for {LLM} Agent Training}.
\newblock \bibinfo{journal}{\emph{CoRR}}  \bibinfo{volume}{abs/2505.10978} (\bibinfo{year}{2025}).
\newblock
\urldef\tempurl%
\url{https://doi.org/10.48550/ARXIV.2505.10978}
\showDOI{\tempurl}
\showeprint[arXiv]{2505.10978}


\bibitem[\protect\citeauthoryear{Feng, Xue, Liu, and An}{Feng et~al\mbox{.}}{2025c}]%
        {GIGPO}
\bibfield{author}{\bibinfo{person}{Lang Feng}, \bibinfo{person}{Zhenghai Xue}, \bibinfo{person}{Tingcong Liu}, {and} \bibinfo{person}{Bo An}.} \bibinfo{year}{2025}\natexlab{c}.
\newblock \showarticletitle{Group-in-Group Policy Optimization for LLM Agent Training}.
\newblock \bibinfo{journal}{\emph{arXiv preprint arXiv:2505.10978}} (\bibinfo{year}{2025}).
\newblock


\bibitem[\protect\citeauthoryear{Gao, Fu, Xie, Xu, He, Mei, Zhu, and Wu}{Gao et~al\mbox{.}}{2025}]%
        {asearcher}
\bibfield{author}{\bibinfo{person}{Jiaxuan Gao}, \bibinfo{person}{Wei Fu}, \bibinfo{person}{Minyang Xie}, \bibinfo{person}{Shusheng Xu}, \bibinfo{person}{Chuyi He}, \bibinfo{person}{Zhiyu Mei}, \bibinfo{person}{Banghua Zhu}, {and} \bibinfo{person}{Yi Wu}.} \bibinfo{year}{2025}\natexlab{}.
\newblock \bibinfo{title}{Beyond Ten Turns: Unlocking Long-Horizon Agentic Search with Large-Scale Asynchronous RL}.
\newblock
\newblock
\showeprint[arxiv]{cs.CL/2508.07976}
\urldef\tempurl%
\url{https://arxiv.org/abs/2508.07976}
\showURL{%
\tempurl}


\bibitem[\protect\citeauthoryear{Gou, Shao, Gong, Shen, Yang, Huang, Duan, and Chen}{Gou et~al\mbox{.}}{2024}]%
        {tora}
\bibfield{author}{\bibinfo{person}{Zhibin Gou}, \bibinfo{person}{Zhihong Shao}, \bibinfo{person}{Yeyun Gong}, \bibinfo{person}{Yelong Shen}, \bibinfo{person}{Yujiu Yang}, \bibinfo{person}{Minlie Huang}, \bibinfo{person}{Nan Duan}, {and} \bibinfo{person}{Weizhu Chen}.} \bibinfo{year}{2024}\natexlab{}.
\newblock \showarticletitle{ToRA: {A} Tool-Integrated Reasoning Agent for Mathematical Problem Solving}. In \bibinfo{booktitle}{\emph{The Twelfth International Conference on Learning Representations, {ICLR} 2024, Vienna, Austria, May 7-11, 2024}}. \bibinfo{publisher}{OpenReview.net}.
\newblock
\urldef\tempurl%
\url{https://openreview.net/forum?id=Ep0TtjVoap}
\showURL{%
\tempurl}


\bibitem[\protect\citeauthoryear{Guo, Yang, Zhang, Song, Zhang, Xu, Zhu, Ma, Wang, Bi, et~al\mbox{.}}{Guo et~al\mbox{.}}{2025b}]%
        {deepseek-r1}
\bibfield{author}{\bibinfo{person}{Daya Guo}, \bibinfo{person}{Dejian Yang}, \bibinfo{person}{Haowei Zhang}, \bibinfo{person}{Junxiao Song}, \bibinfo{person}{Ruoyu Zhang}, \bibinfo{person}{Runxin Xu}, \bibinfo{person}{Qihao Zhu}, \bibinfo{person}{Shirong Ma}, \bibinfo{person}{Peiyi Wang}, \bibinfo{person}{Xiao Bi}, {et~al\mbox{.}}} \bibinfo{year}{2025}\natexlab{b}.
\newblock \showarticletitle{Deepseek-r1: Incentivizing reasoning capability in llms via reinforcement learning}.
\newblock \bibinfo{journal}{\emph{arXiv preprint arXiv:2501.12948}} (\bibinfo{year}{2025}).
\newblock


\bibitem[\protect\citeauthoryear{Guo, Guo, Sun, He, Yang, Lu, Zhang, Guo, Zhang, Liu, Duan, Xiao, Wen, Xu, and Dai}{Guo et~al\mbox{.}}{2025a}]%
        {Web-CogReasoner}
\bibfield{author}{\bibinfo{person}{Yuhan Guo}, \bibinfo{person}{Cong Guo}, \bibinfo{person}{Aiwen Sun}, \bibinfo{person}{Hongliang He}, \bibinfo{person}{Xinyu Yang}, \bibinfo{person}{Yue Lu}, \bibinfo{person}{Yingji Zhang}, \bibinfo{person}{Xuntao Guo}, \bibinfo{person}{Dong Zhang}, \bibinfo{person}{Jianzhuang Liu}, \bibinfo{person}{Jiang Duan}, \bibinfo{person}{Yijia Xiao}, \bibinfo{person}{Liangjian Wen}, \bibinfo{person}{Hai{-}Ming Xu}, {and} \bibinfo{person}{Yong Dai}.} \bibinfo{year}{2025}\natexlab{a}.
\newblock \showarticletitle{Web-CogReasoner: Towards Knowledge-Induced Cognitive Reasoning for Web Agents}.
\newblock \bibinfo{journal}{\emph{CoRR}}  \bibinfo{volume}{abs/2508.01858} (\bibinfo{year}{2025}).
\newblock
\urldef\tempurl%
\url{https://doi.org/10.48550/ARXIV.2508.01858}
\showDOI{\tempurl}
\showeprint[arXiv]{2508.01858}


\bibitem[\protect\citeauthoryear{Guu, Lee, Tung, Pasupat, and Chang}{Guu et~al\mbox{.}}{2020}]%
        {guu2020realm}
\bibfield{author}{\bibinfo{person}{Kelvin Guu}, \bibinfo{person}{Kenton Lee}, \bibinfo{person}{Zora Tung}, \bibinfo{person}{Panupong Pasupat}, {and} \bibinfo{person}{Ming{-}Wei Chang}.} \bibinfo{year}{2020}\natexlab{}.
\newblock \showarticletitle{{REALM:} Retrieval-Augmented Language Model Pre-Training}.
\newblock \bibinfo{journal}{\emph{CoRR}}  \bibinfo{volume}{abs/2002.08909} (\bibinfo{year}{2020}).
\newblock
\showeprint[arXiv]{2002.08909}
\urldef\tempurl%
\url{https://arxiv.org/abs/2002.08909}
\showURL{%
\tempurl}


\bibitem[\protect\citeauthoryear{Hendrycks, Burns, Kadavath, Arora, Basart, Tang, Song, and Steinhardt}{Hendrycks et~al\mbox{.}}{2021}]%
        {MATH}
\bibfield{author}{\bibinfo{person}{Dan Hendrycks}, \bibinfo{person}{Collin Burns}, \bibinfo{person}{Saurav Kadavath}, \bibinfo{person}{Akul Arora}, \bibinfo{person}{Steven Basart}, \bibinfo{person}{Eric Tang}, \bibinfo{person}{Dawn Song}, {and} \bibinfo{person}{Jacob Steinhardt}.} \bibinfo{year}{2021}\natexlab{}.
\newblock \showarticletitle{Measuring Mathematical Problem Solving With the {MATH} Dataset}. In \bibinfo{booktitle}{\emph{Proceedings of the Neural Information Processing Systems Track on Datasets and Benchmarks 1, NeurIPS Datasets and Benchmarks 2021, December 2021, virtual}}, \bibfield{editor}{\bibinfo{person}{Joaquin Vanschoren} {and} \bibinfo{person}{Sai{-}Kit Yeung}} (Eds.).
\newblock
\urldef\tempurl%
\url{https://datasets-benchmarks-proceedings.neurips.cc/paper/2021/hash/be83ab3ecd0db773eb2dc1b0a17836a1-Abstract-round2.html}
\showURL{%
\tempurl}


\bibitem[\protect\citeauthoryear{Ho, Nguyen, Sugawara, and Aizawa}{Ho et~al\mbox{.}}{2020}]%
        {2wiki}
\bibfield{author}{\bibinfo{person}{Xanh Ho}, \bibinfo{person}{Anh{-}Khoa~Duong Nguyen}, \bibinfo{person}{Saku Sugawara}, {and} \bibinfo{person}{Akiko Aizawa}.} \bibinfo{year}{2020}\natexlab{}.
\newblock \showarticletitle{Constructing {A} Multi-hop {QA} Dataset for Comprehensive Evaluation of Reasoning Steps}. In \bibinfo{booktitle}{\emph{Proceedings of the 28th International Conference on Computational Linguistics, {COLING} 2020, Barcelona, Spain (Online), December 8-13, 2020}}, \bibfield{editor}{\bibinfo{person}{Donia Scott}, \bibinfo{person}{N{\'{u}}ria Bel}, {and} \bibinfo{person}{Chengqing Zong}} (Eds.). \bibinfo{publisher}{International Committee on Computational Linguistics}, \bibinfo{pages}{6609--6625}.
\newblock
\urldef\tempurl%
\url{https://doi.org/10.18653/V1/2020.COLING-MAIN.580}
\showDOI{\tempurl}


\bibitem[\protect\citeauthoryear{Hou, Hu, Li, Lu, Tang, and Dong}{Hou et~al\mbox{.}}{2025}]%
        {TreeRL}
\bibfield{author}{\bibinfo{person}{Zhenyu Hou}, \bibinfo{person}{Ziniu Hu}, \bibinfo{person}{Yujiang Li}, \bibinfo{person}{Rui Lu}, \bibinfo{person}{Jie Tang}, {and} \bibinfo{person}{Yuxiao Dong}.} \bibinfo{year}{2025}\natexlab{}.
\newblock \showarticletitle{TreeRL: {LLM} Reinforcement Learning with On-Policy Tree Search}. In \bibinfo{booktitle}{\emph{Proceedings of the 63rd Annual Meeting of the Association for Computational Linguistics (Volume 1: Long Papers), {ACL} 2025, Vienna, Austria, July 27 - August 1, 2025}}, \bibfield{editor}{\bibinfo{person}{Wanxiang Che}, \bibinfo{person}{Joyce Nabende}, \bibinfo{person}{Ekaterina Shutova}, {and} \bibinfo{person}{Mohammad~Taher Pilehvar}} (Eds.). \bibinfo{publisher}{Association for Computational Linguistics}, \bibinfo{pages}{12355--12369}.
\newblock
\urldef\tempurl%
\url{https://aclanthology.org/2025.acl-long.604/}
\showURL{%
\tempurl}


\bibitem[\protect\citeauthoryear{Hu}{Hu}{2025}]%
        {hu2025reinforce++}
\bibfield{author}{\bibinfo{person}{Jian Hu}.} \bibinfo{year}{2025}\natexlab{}.
\newblock \showarticletitle{Reinforce++: A simple and efficient approach for aligning large language models}.
\newblock \bibinfo{journal}{\emph{arXiv preprint arXiv:2501.03262}} (\bibinfo{year}{2025}).
\newblock


\bibitem[\protect\citeauthoryear{Huang, Zhuang, Lu, Qin, Xu, Zhao, Peng, Hu, Shen, Hu, Gu, Tu, Liu, Chen, Fu, Fan, Gu, Wang, Yang, Li, and Zhao}{Huang et~al\mbox{.}}{2025}]%
        {ReinforcementLearningwithRubricAnchors}
\bibfield{author}{\bibinfo{person}{Zenan Huang}, \bibinfo{person}{Yihong Zhuang}, \bibinfo{person}{Guoshan Lu}, \bibinfo{person}{Zeyu Qin}, \bibinfo{person}{Haokai Xu}, \bibinfo{person}{Tianyu Zhao}, \bibinfo{person}{Ru Peng}, \bibinfo{person}{Jiaqi Hu}, \bibinfo{person}{Zhanming Shen}, \bibinfo{person}{Xiaomeng Hu}, \bibinfo{person}{Xijun Gu}, \bibinfo{person}{Peiyi Tu}, \bibinfo{person}{Jiaxin Liu}, \bibinfo{person}{Wenyu Chen}, \bibinfo{person}{Yuzhuo Fu}, \bibinfo{person}{Zhiting Fan}, \bibinfo{person}{Yanmei Gu}, \bibinfo{person}{Yuanyuan Wang}, \bibinfo{person}{Zhengkai Yang}, \bibinfo{person}{Jianguo Li}, {and} \bibinfo{person}{Junbo Zhao}.} \bibinfo{year}{2025}\natexlab{}.
\newblock \showarticletitle{Reinforcement Learning with Rubric Anchors}.
\newblock \bibinfo{journal}{\emph{CoRR}}  \bibinfo{volume}{abs/2508.12790} (\bibinfo{year}{2025}).
\newblock
\urldef\tempurl%
\url{https://doi.org/10.48550/ARXIV.2508.12790}
\showDOI{\tempurl}
\showeprint[arXiv]{2508.12790}


\bibitem[\protect\citeauthoryear{Hurst, Lerer, Goucher, Perelman, Ramesh, Clark, Ostrow, Welihinda, Hayes, Radford, et~al\mbox{.}}{Hurst et~al\mbox{.}}{2024}]%
        {gpt_4o_system_card}
\bibfield{author}{\bibinfo{person}{Aaron Hurst}, \bibinfo{person}{Adam Lerer}, \bibinfo{person}{Adam~P Goucher}, \bibinfo{person}{Adam Perelman}, \bibinfo{person}{Aditya Ramesh}, \bibinfo{person}{Aidan Clark}, \bibinfo{person}{AJ Ostrow}, \bibinfo{person}{Akila Welihinda}, \bibinfo{person}{Alan Hayes}, \bibinfo{person}{Alec Radford}, {et~al\mbox{.}}} \bibinfo{year}{2024}\natexlab{}.
\newblock \showarticletitle{Gpt-4o system card}.
\newblock \bibinfo{journal}{\emph{arXiv preprint arXiv:2410.21276}} (\bibinfo{year}{2024}).
\newblock


\bibitem[\protect\citeauthoryear{Ji, Ma, Wang, Chen, Chu, and Wu}{Ji et~al\mbox{.}}{2025}]%
        {ji2025treesearchllmagent}
\bibfield{author}{\bibinfo{person}{Yuxiang Ji}, \bibinfo{person}{Ziyu Ma}, \bibinfo{person}{Yong Wang}, \bibinfo{person}{Guanhua Chen}, \bibinfo{person}{Xiangxiang Chu}, {and} \bibinfo{person}{Liaoni Wu}.} \bibinfo{year}{2025}\natexlab{}.
\newblock \bibinfo{title}{Tree Search for LLM Agent Reinforcement Learning}.
\newblock
\newblock
\showeprint[arxiv]{cs.LG/2509.21240}
\urldef\tempurl%
\url{https://arxiv.org/abs/2509.21240}
\showURL{%
\tempurl}


\bibitem[\protect\citeauthoryear{Jiang, Lu, Li, Lyu, Nie, Wang, Su, Chen, Zou, Du, et~al\mbox{.}}{Jiang et~al\mbox{.}}{2025}]%
        {jiang2025verltool}
\bibfield{author}{\bibinfo{person}{Dongfu Jiang}, \bibinfo{person}{Yi Lu}, \bibinfo{person}{Zhuofeng Li}, \bibinfo{person}{Zhiheng Lyu}, \bibinfo{person}{Ping Nie}, \bibinfo{person}{Haozhe Wang}, \bibinfo{person}{Alex Su}, \bibinfo{person}{Hui Chen}, \bibinfo{person}{Kai Zou}, \bibinfo{person}{Chao Du}, {et~al\mbox{.}}} \bibinfo{year}{2025}\natexlab{}.
\newblock \showarticletitle{VerlTool: Towards Holistic Agentic Reinforcement Learning with Tool Use}.
\newblock \bibinfo{journal}{\emph{arXiv preprint arXiv:2509.01055}} (\bibinfo{year}{2025}).
\newblock


\bibitem[\protect\citeauthoryear{Jin, Zeng, Yue, Wang, Zamani, and Han}{Jin et~al\mbox{.}}{2025b}]%
        {searchr1}
\bibfield{author}{\bibinfo{person}{Bowen Jin}, \bibinfo{person}{Hansi Zeng}, \bibinfo{person}{Zhenrui Yue}, \bibinfo{person}{Dong Wang}, \bibinfo{person}{Hamed Zamani}, {and} \bibinfo{person}{Jiawei Han}.} \bibinfo{year}{2025}\natexlab{b}.
\newblock \showarticletitle{Search-R1: Training LLMs to Reason and Leverage Search Engines with Reinforcement Learning}.
\newblock \bibinfo{journal}{\emph{CoRR}}  \bibinfo{volume}{abs/2503.09516} (\bibinfo{year}{2025}).
\newblock
\urldef\tempurl%
\url{https://doi.org/10.48550/ARXIV.2503.09516}
\showDOI{\tempurl}
\showeprint[arXiv]{2503.09516}


\bibitem[\protect\citeauthoryear{Jin, Li, Dong, Zhang, Zhu, Zhao, Qian, and Dou}{Jin et~al\mbox{.}}{2025a}]%
        {hira}
\bibfield{author}{\bibinfo{person}{Jiajie Jin}, \bibinfo{person}{Xiaoxi Li}, \bibinfo{person}{Guanting Dong}, \bibinfo{person}{Yuyao Zhang}, \bibinfo{person}{Yutao Zhu}, \bibinfo{person}{Yang Zhao}, \bibinfo{person}{Hongjin Qian}, {and} \bibinfo{person}{Zhicheng Dou}.} \bibinfo{year}{2025}\natexlab{a}.
\newblock \bibinfo{title}{Decoupled Planning and Execution: A Hierarchical Reasoning Framework for Deep Search}.
\newblock
\newblock
\showeprint[arxiv]{cs.AI/2507.02652}
\urldef\tempurl%
\url{https://arxiv.org/abs/2507.02652}
\showURL{%
\tempurl}


\bibitem[\protect\citeauthoryear{Jin, Zhu, Yang, Zhang, and Dou}{Jin et~al\mbox{.}}{2024}]%
        {flashrag}
\bibfield{author}{\bibinfo{person}{Jiajie Jin}, \bibinfo{person}{Yutao Zhu}, \bibinfo{person}{Xinyu Yang}, \bibinfo{person}{Chenghao Zhang}, {and} \bibinfo{person}{Zhicheng Dou}.} \bibinfo{year}{2024}\natexlab{}.
\newblock \showarticletitle{FlashRAG: {A} Modular Toolkit for Efficient Retrieval-Augmented Generation Research}.
\newblock \bibinfo{journal}{\emph{CoRR}}  \bibinfo{volume}{abs/2405.13576} (\bibinfo{year}{2024}).
\newblock
\urldef\tempurl%
\url{https://doi.org/10.48550/ARXIV.2405.13576}
\showDOI{\tempurl}
\showeprint[arXiv]{2405.13576}


\bibitem[\protect\citeauthoryear{Krishna, Krishna, Mohananey, Schwarcz, Stambler, Upadhyay, and Faruqui}{Krishna et~al\mbox{.}}{2024}]%
        {frames}
\bibfield{author}{\bibinfo{person}{Satyapriya Krishna}, \bibinfo{person}{Kalpesh Krishna}, \bibinfo{person}{Anhad Mohananey}, \bibinfo{person}{Steven Schwarcz}, \bibinfo{person}{Adam Stambler}, \bibinfo{person}{Shyam Upadhyay}, {and} \bibinfo{person}{Manaal Faruqui}.} \bibinfo{year}{2024}\natexlab{}.
\newblock \bibinfo{title}{Fact, Fetch, and Reason: A Unified Evaluation of Retrieval-Augmented Generation}.
\newblock
\newblock
\showeprint[arxiv]{cs.CL/2409.12941}
\urldef\tempurl%
\url{https://arxiv.org/abs/2409.12941}
\showURL{%
\tempurl}


\bibitem[\protect\citeauthoryear{Lewis, Perez, Piktus, Petroni, Karpukhin, Goyal, K{\"{u}}ttler, Lewis, Yih, Rockt{\"{a}}schel, Riedel, and Kiela}{Lewis et~al\mbox{.}}{2020a}]%
        {rag_lewis}
\bibfield{author}{\bibinfo{person}{Patrick Lewis}, \bibinfo{person}{Ethan Perez}, \bibinfo{person}{Aleksandra Piktus}, \bibinfo{person}{Fabio Petroni}, \bibinfo{person}{Vladimir Karpukhin}, \bibinfo{person}{Naman Goyal}, \bibinfo{person}{Heinrich K{\"{u}}ttler}, \bibinfo{person}{Mike Lewis}, \bibinfo{person}{Wen{-}tau Yih}, \bibinfo{person}{Tim Rockt{\"{a}}schel}, \bibinfo{person}{Sebastian Riedel}, {and} \bibinfo{person}{Douwe Kiela}.} \bibinfo{year}{2020}\natexlab{a}.
\newblock \showarticletitle{Retrieval-Augmented Generation for Knowledge-Intensive {NLP} Tasks}. In \bibinfo{booktitle}{\emph{Advances in Neural Information Processing Systems 33: Annual Conference on Neural Information Processing Systems 2020, NeurIPS 2020, December 6-12, 2020, virtual}}, \bibfield{editor}{\bibinfo{person}{Hugo Larochelle}, \bibinfo{person}{Marc'Aurelio Ranzato}, \bibinfo{person}{Raia Hadsell}, \bibinfo{person}{Maria{-}Florina Balcan}, {and} \bibinfo{person}{Hsuan{-}Tien Lin}} (Eds.).
\newblock
\urldef\tempurl%
\url{https://proceedings.neurips.cc/paper/2020/hash/6b493230205f780e1bc26945df7481e5-Abstract.html}
\showURL{%
\tempurl}


\bibitem[\protect\citeauthoryear{Lewis, Perez, Piktus, Petroni, Karpukhin, Goyal, K{\"{u}}ttler, Lewis, Yih, Rockt{\"{a}}schel, Riedel, and Kiela}{Lewis et~al\mbox{.}}{2020b}]%
        {lewis2021retrievalaugmented}
\bibfield{author}{\bibinfo{person}{Patrick S.~H. Lewis}, \bibinfo{person}{Ethan Perez}, \bibinfo{person}{Aleksandra Piktus}, \bibinfo{person}{Fabio Petroni}, \bibinfo{person}{Vladimir Karpukhin}, \bibinfo{person}{Naman Goyal}, \bibinfo{person}{Heinrich K{\"{u}}ttler}, \bibinfo{person}{Mike Lewis}, \bibinfo{person}{Wen{-}tau Yih}, \bibinfo{person}{Tim Rockt{\"{a}}schel}, \bibinfo{person}{Sebastian Riedel}, {and} \bibinfo{person}{Douwe Kiela}.} \bibinfo{year}{2020}\natexlab{b}.
\newblock \showarticletitle{Retrieval-Augmented Generation for Knowledge-Intensive {NLP} Tasks}. In \bibinfo{booktitle}{\emph{Advances in Neural Information Processing Systems 33: Annual Conference on Neural Information Processing Systems 2020, NeurIPS 2020, December 6-12, 2020, virtual}}, \bibfield{editor}{\bibinfo{person}{Hugo Larochelle}, \bibinfo{person}{Marc'Aurelio Ranzato}, \bibinfo{person}{Raia Hadsell}, \bibinfo{person}{Maria{-}Florina Balcan}, {and} \bibinfo{person}{Hsuan{-}Tien Lin}} (Eds.).
\newblock
\urldef\tempurl%
\url{https://proceedings.neurips.cc/paper/2020/hash/6b493230205f780e1bc26945df7481e5-Abstract.html}
\showURL{%
\tempurl}


\bibitem[\protect\citeauthoryear{Li, Dong, Xue, Peng, Wang, and Liu}{Li et~al\mbox{.}}{2024a}]%
        {dotamath}
\bibfield{author}{\bibinfo{person}{Chengpeng Li}, \bibinfo{person}{Guanting Dong}, \bibinfo{person}{Mingfeng Xue}, \bibinfo{person}{Ru Peng}, \bibinfo{person}{Xiang Wang}, {and} \bibinfo{person}{Dayiheng Liu}.} \bibinfo{year}{2024}\natexlab{a}.
\newblock \showarticletitle{DotaMath: Decomposition of Thought with Code Assistance and Self-correction for Mathematical Reasoning}.
\newblock \bibinfo{journal}{\emph{CoRR}}  \bibinfo{volume}{abs/2407.04078} (\bibinfo{year}{2024}).
\newblock
\urldef\tempurl%
\url{https://doi.org/10.48550/ARXIV.2407.04078}
\showDOI{\tempurl}
\showeprint[arXiv]{2407.04078}


\bibitem[\protect\citeauthoryear{Li, Zhang, Yin, Ye, Zhao, Zhang, Ou, Zhang, Wu, Wu, et~al\mbox{.}}{Li et~al\mbox{.}}{2025f}]%
        {li2025websailor-v2}
\bibfield{author}{\bibinfo{person}{Kuan Li}, \bibinfo{person}{Zhongwang Zhang}, \bibinfo{person}{Huifeng Yin}, \bibinfo{person}{Rui Ye}, \bibinfo{person}{Yida Zhao}, \bibinfo{person}{Liwen Zhang}, \bibinfo{person}{Litu Ou}, \bibinfo{person}{Dingchu Zhang}, \bibinfo{person}{Xixi Wu}, \bibinfo{person}{Jialong Wu}, {et~al\mbox{.}}} \bibinfo{year}{2025}\natexlab{f}.
\newblock \showarticletitle{WebSailor-V2: Bridging the Chasm to Proprietary Agents via Synthetic Data and Scalable Reinforcement Learning}.
\newblock \bibinfo{journal}{\emph{arXiv preprint arXiv:2509.13305}} (\bibinfo{year}{2025}).
\newblock


\bibitem[\protect\citeauthoryear{Li, Zhang, Yin, Zhang, Ou, Wu, Yin, Li, Tao, Wang, Shen, Zhang, Zhang, Wu, Jiang, Yan, Xie, Huang, and Zhou}{Li et~al\mbox{.}}{2025g}]%
        {li2025websailor}
\bibfield{author}{\bibinfo{person}{Kuan Li}, \bibinfo{person}{Zhongwang Zhang}, \bibinfo{person}{Huifeng Yin}, \bibinfo{person}{Liwen Zhang}, \bibinfo{person}{Litu Ou}, \bibinfo{person}{Jialong Wu}, \bibinfo{person}{Wenbiao Yin}, \bibinfo{person}{Baixuan Li}, \bibinfo{person}{Zhengwei Tao}, \bibinfo{person}{Xinyu Wang}, \bibinfo{person}{Weizhou Shen}, \bibinfo{person}{Junkai Zhang}, \bibinfo{person}{Dingchu Zhang}, \bibinfo{person}{Xixi Wu}, \bibinfo{person}{Yong Jiang}, \bibinfo{person}{Ming Yan}, \bibinfo{person}{Pengjun Xie}, \bibinfo{person}{Fei Huang}, {and} \bibinfo{person}{Jingren Zhou}.} \bibinfo{year}{2025}\natexlab{g}.
\newblock \bibinfo{title}{WebSailor: Navigating Super-human Reasoning for Web Agent}.
\newblock
\newblock
\showeprint[arxiv]{cs.CL/2507.02592}
\urldef\tempurl%
\url{https://arxiv.org/abs/2507.02592}
\showURL{%
\tempurl}


\bibitem[\protect\citeauthoryear{Li, Lin, Jiang, Cao, Liu, Zhang, Huang, Chen, Sun, Wang, Lu, Qin, Zhu, Yao, Fan, Li, Wang, Liu, Zhu, Zhu, Shi, Wang, Guan, Tang, Liu, Jiang, Yang, Liu, Zhang, and Zhou}{Li et~al\mbox{.}}{2025e}]%
        {Chain-of-Agents}
\bibfield{author}{\bibinfo{person}{Weizhen Li}, \bibinfo{person}{Jianbo Lin}, \bibinfo{person}{Zhuosong Jiang}, \bibinfo{person}{Jingyi Cao}, \bibinfo{person}{Xinpeng Liu}, \bibinfo{person}{Jiayu Zhang}, \bibinfo{person}{Zhenqiang Huang}, \bibinfo{person}{Qianben Chen}, \bibinfo{person}{Weichen Sun}, \bibinfo{person}{Qiexiang Wang}, \bibinfo{person}{Hongxuan Lu}, \bibinfo{person}{Tianrui Qin}, \bibinfo{person}{Chenghao Zhu}, \bibinfo{person}{Yi Yao}, \bibinfo{person}{Shuying Fan}, \bibinfo{person}{Xiaowan Li}, \bibinfo{person}{Tiannan Wang}, \bibinfo{person}{Pai Liu}, \bibinfo{person}{King Zhu}, \bibinfo{person}{He Zhu}, \bibinfo{person}{Dingfeng Shi}, \bibinfo{person}{Piaohong Wang}, \bibinfo{person}{Yeyi Guan}, \bibinfo{person}{Xiangru Tang}, \bibinfo{person}{Minghao Liu}, \bibinfo{person}{Yuchen~Eleanor Jiang}, \bibinfo{person}{Jian Yang}, \bibinfo{person}{Jiaheng Liu}, \bibinfo{person}{Ge Zhang}, {and} \bibinfo{person}{Wangchunshu Zhou}.} \bibinfo{year}{2025}\natexlab{e}.
\newblock \showarticletitle{Chain-of-Agents: End-to-End Agent Foundation Models via Multi-Agent Distillation and Agentic {RL}}.
\newblock \bibinfo{journal}{\emph{CoRR}}  \bibinfo{volume}{abs/2508.13167} (\bibinfo{year}{2025}).
\newblock
\urldef\tempurl%
\url{https://doi.org/10.48550/ARXIV.2508.13167}
\showDOI{\tempurl}
\showeprint[arXiv]{2508.13167}


\bibitem[\protect\citeauthoryear{Li, Dong, Jin, Zhang, Zhou, Zhu, Zhang, and Dou}{Li et~al\mbox{.}}{2025a}]%
        {searcho1}
\bibfield{author}{\bibinfo{person}{Xiaoxi Li}, \bibinfo{person}{Guanting Dong}, \bibinfo{person}{Jiajie Jin}, \bibinfo{person}{Yuyao Zhang}, \bibinfo{person}{Yujia Zhou}, \bibinfo{person}{Yutao Zhu}, \bibinfo{person}{Peitian Zhang}, {and} \bibinfo{person}{Zhicheng Dou}.} \bibinfo{year}{2025}\natexlab{a}.
\newblock \showarticletitle{Search-o1: Agentic Search-Enhanced Large Reasoning Models}.
\newblock \bibinfo{journal}{\emph{CoRR}}  \bibinfo{volume}{abs/2501.05366} (\bibinfo{year}{2025}).
\newblock
\urldef\tempurl%
\url{https://doi.org/10.48550/ARXIV.2501.05366}
\showDOI{\tempurl}
\showeprint[arXiv]{2501.05366}


\bibitem[\protect\citeauthoryear{Li, Jin, Dong, Qian, Zhu, Wu, Wen, and Dou}{Li et~al\mbox{.}}{2025d}]%
        {webthinker}
\bibfield{author}{\bibinfo{person}{Xiaoxi Li}, \bibinfo{person}{Jiajie Jin}, \bibinfo{person}{Guanting Dong}, \bibinfo{person}{Hongjin Qian}, \bibinfo{person}{Yutao Zhu}, \bibinfo{person}{Yongkang Wu}, \bibinfo{person}{Ji-Rong Wen}, {and} \bibinfo{person}{Zhicheng Dou}.} \bibinfo{year}{2025}\natexlab{d}.
\newblock \showarticletitle{WebThinker: Empowering Large Reasoning Models with Deep Research Capability}.
\newblock \bibinfo{journal}{\emph{arXiv preprint arXiv:2504.21776}} (\bibinfo{year}{2025}).
\newblock


\bibitem[\protect\citeauthoryear{Li, Jin, Zhou, Wu, Li, Ye, and Dou}{Li et~al\mbox{.}}{2024b}]%
        {retrollm}
\bibfield{author}{\bibinfo{person}{Xiaoxi Li}, \bibinfo{person}{Jiajie Jin}, \bibinfo{person}{Yujia Zhou}, \bibinfo{person}{Yongkang Wu}, \bibinfo{person}{Zhonghua Li}, \bibinfo{person}{Qi Ye}, {and} \bibinfo{person}{Zhicheng Dou}.} \bibinfo{year}{2024}\natexlab{b}.
\newblock \showarticletitle{RetroLLM: Empowering Large Language Models to Retrieve Fine-grained Evidence within Generation}.
\newblock \bibinfo{journal}{\emph{CoRR}}  \bibinfo{volume}{abs/2412.11919} (\bibinfo{year}{2024}).
\newblock
\urldef\tempurl%
\url{https://doi.org/10.48550/ARXIV.2412.11919}
\showDOI{\tempurl}
\showeprint[arXiv]{2412.11919}


\bibitem[\protect\citeauthoryear{Li, Zou, and Liu}{Li et~al\mbox{.}}{2025h}]%
        {torl}
\bibfield{author}{\bibinfo{person}{Xuefeng Li}, \bibinfo{person}{Haoyang Zou}, {and} \bibinfo{person}{Pengfei Liu}.} \bibinfo{year}{2025}\natexlab{h}.
\newblock \showarticletitle{ToRL: Scaling Tool-Integrated {RL}}.
\newblock \bibinfo{journal}{\emph{CoRR}}  \bibinfo{volume}{abs/2503.23383} (\bibinfo{year}{2025}).
\newblock
\urldef\tempurl%
\url{https://doi.org/10.48550/ARXIV.2503.23383}
\showDOI{\tempurl}
\showeprint[arXiv]{2503.23383}


\bibitem[\protect\citeauthoryear{Li, Gu, Wen, Li, Xing, Guo, Zheng, Zhou, Qu, Zhou, et~al\mbox{.}}{Li et~al\mbox{.}}{2025b}]%
        {li2025treepo}
\bibfield{author}{\bibinfo{person}{Yizhi Li}, \bibinfo{person}{Qingshui Gu}, \bibinfo{person}{Zhoufutu Wen}, \bibinfo{person}{Ziniu Li}, \bibinfo{person}{Tianshun Xing}, \bibinfo{person}{Shuyue Guo}, \bibinfo{person}{Tianyu Zheng}, \bibinfo{person}{Xin Zhou}, \bibinfo{person}{Xingwei Qu}, \bibinfo{person}{Wangchunshu Zhou}, {et~al\mbox{.}}} \bibinfo{year}{2025}\natexlab{b}.
\newblock \showarticletitle{Treepo: Bridging the gap of policy optimization and efficacy and inference efficiency with heuristic tree-based modeling}.
\newblock \bibinfo{journal}{\emph{arXiv preprint arXiv:2508.17445}} (\bibinfo{year}{2025}).
\newblock


\bibitem[\protect\citeauthoryear{Li, Gu, Wen, Li, Xing, Guo, Zheng, Zhou, Qu, Zhou, et~al\mbox{.}}{Li et~al\mbox{.}}{2025c}]%
        {li2025treepobridginggappolicy}
\bibfield{author}{\bibinfo{person}{Yizhi Li}, \bibinfo{person}{Qingshui Gu}, \bibinfo{person}{Zhoufutu Wen}, \bibinfo{person}{Ziniu Li}, \bibinfo{person}{Tianshun Xing}, \bibinfo{person}{Shuyue Guo}, \bibinfo{person}{Tianyu Zheng}, \bibinfo{person}{Xin Zhou}, \bibinfo{person}{Xingwei Qu}, \bibinfo{person}{Wangchunshu Zhou}, {et~al\mbox{.}}} \bibinfo{year}{2025}\natexlab{c}.
\newblock \showarticletitle{Treepo: Bridging the gap of policy optimization and efficacy and inference efficiency with heuristic tree-based modeling}.
\newblock \bibinfo{journal}{\emph{arXiv preprint arXiv:2508.17445}} (\bibinfo{year}{2025}).
\newblock


\bibitem[\protect\citeauthoryear{Lightman, Kosaraju, Burda, Edwards, Baker, Lee, Leike, Schulman, Sutskever, and Cobbe}{Lightman et~al\mbox{.}}{2024}]%
        {math500}
\bibfield{author}{\bibinfo{person}{Hunter Lightman}, \bibinfo{person}{Vineet Kosaraju}, \bibinfo{person}{Yuri Burda}, \bibinfo{person}{Harrison Edwards}, \bibinfo{person}{Bowen Baker}, \bibinfo{person}{Teddy Lee}, \bibinfo{person}{Jan Leike}, \bibinfo{person}{John Schulman}, \bibinfo{person}{Ilya Sutskever}, {and} \bibinfo{person}{Karl Cobbe}.} \bibinfo{year}{2024}\natexlab{}.
\newblock \showarticletitle{Let's Verify Step by Step}. In \bibinfo{booktitle}{\emph{The Twelfth International Conference on Learning Representations, {ICLR} 2024, Vienna, Austria, May 7-11, 2024}}. \bibinfo{publisher}{OpenReview.net}.
\newblock
\urldef\tempurl%
\url{https://openreview.net/forum?id=v8L0pN6EOi}
\showURL{%
\tempurl}


\bibitem[\protect\citeauthoryear{Liu, He, Lin, Yang, Shen, and Liu}{Liu et~al\mbox{.}}{2025a}]%
        {liu2025ettrl}
\bibfield{author}{\bibinfo{person}{Jia Liu}, \bibinfo{person}{ChangYi He}, \bibinfo{person}{YingQiao Lin}, \bibinfo{person}{MingMin Yang}, \bibinfo{person}{FeiYang Shen}, {and} \bibinfo{person}{ShaoGuo Liu}.} \bibinfo{year}{2025}\natexlab{a}.
\newblock \showarticletitle{Ettrl: Balancing exploration and exploitation in llm test-time reinforcement learning via entropy mechanism}.
\newblock \bibinfo{journal}{\emph{arXiv preprint arXiv:2508.11356}} (\bibinfo{year}{2025}).
\newblock


\bibitem[\protect\citeauthoryear{Liu, Liu, He, Wang, Liu, Pan, Hu, Xiong, Huang, Hu, et~al\mbox{.}}{Liu et~al\mbox{.}}{2025b}]%
        {liu2025part}
\bibfield{author}{\bibinfo{person}{Zihe Liu}, \bibinfo{person}{Jiashun Liu}, \bibinfo{person}{Yancheng He}, \bibinfo{person}{Weixun Wang}, \bibinfo{person}{Jiaheng Liu}, \bibinfo{person}{Ling Pan}, \bibinfo{person}{Xinyu Hu}, \bibinfo{person}{Shaopan Xiong}, \bibinfo{person}{Ju Huang}, \bibinfo{person}{Jian Hu}, {et~al\mbox{.}}} \bibinfo{year}{2025}\natexlab{b}.
\newblock \showarticletitle{Part I: Tricks or traps? A deep dive into RL for LLM reasoning}.
\newblock \bibinfo{journal}{\emph{arXiv preprint arXiv:2508.08221}} (\bibinfo{year}{2025}).
\newblock


\bibitem[\protect\citeauthoryear{L{\`{u}}, Kazemnejad, Meade, Patel, Shin, Zambrano, Stanczak, Shaw, Pal, and Reddy}{L{\`{u}} et~al\mbox{.}}{2025}]%
        {AgentRewardBench}
\bibfield{author}{\bibinfo{person}{Xing~Han L{\`{u}}}, \bibinfo{person}{Amirhossein Kazemnejad}, \bibinfo{person}{Nicholas Meade}, \bibinfo{person}{Arkil Patel}, \bibinfo{person}{Dongchan Shin}, \bibinfo{person}{Alejandra Zambrano}, \bibinfo{person}{Karolina Stanczak}, \bibinfo{person}{Peter Shaw}, \bibinfo{person}{Christopher~J. Pal}, {and} \bibinfo{person}{Siva Reddy}.} \bibinfo{year}{2025}\natexlab{}.
\newblock \showarticletitle{AgentRewardBench: Evaluating Automatic Evaluations of Web Agent Trajectories}.
\newblock \bibinfo{journal}{\emph{CoRR}}  \bibinfo{volume}{abs/2504.08942} (\bibinfo{year}{2025}).
\newblock
\urldef\tempurl%
\url{https://doi.org/10.48550/ARXIV.2504.08942}
\showDOI{\tempurl}
\showeprint[arXiv]{2504.08942}


\bibitem[\protect\citeauthoryear{Mialon, Fourrier, Wolf, LeCun, and Scialom}{Mialon et~al\mbox{.}}{2024}]%
        {GAIA}
\bibfield{author}{\bibinfo{person}{Gr{\'{e}}goire Mialon}, \bibinfo{person}{Cl{\'{e}}mentine Fourrier}, \bibinfo{person}{Thomas Wolf}, \bibinfo{person}{Yann LeCun}, {and} \bibinfo{person}{Thomas Scialom}.} \bibinfo{year}{2024}\natexlab{}.
\newblock \showarticletitle{{GAIA:} a benchmark for General {AI} Assistants}. In \bibinfo{booktitle}{\emph{The Twelfth International Conference on Learning Representations, {ICLR} 2024, Vienna, Austria, May 7-11, 2024}}. \bibinfo{publisher}{OpenReview.net}.
\newblock
\urldef\tempurl%
\url{https://openreview.net/forum?id=fibxvahvs3}
\showURL{%
\tempurl}


\bibitem[\protect\citeauthoryear{MiniMax}{MiniMax}{2025}]%
        {minimax2025minimaxm1scalingtesttimecompute}
\bibfield{author}{\bibinfo{person}{MiniMax}.} \bibinfo{year}{2025}\natexlab{}.
\newblock \bibinfo{title}{MiniMax-M1: Scaling Test-Time Compute Efficiently with Lightning Attention}.
\newblock
\newblock
\showeprint[arxiv]{cs.CL/2506.13585}
\urldef\tempurl%
\url{https://arxiv.org/abs/2506.13585}
\showURL{%
\tempurl}


\bibitem[\protect\citeauthoryear{Mnih, Kavukcuoglu, Silver, Rusu, Veness, Bellemare, Graves, Riedmiller, Fidjeland, Ostrovski, Petersen, Beattie, Sadik, Antonoglou, King, Kumaran, Wierstra, Legg, and Hassabis}{Mnih et~al\mbox{.}}{2015}]%
        {dqn}
\bibfield{author}{\bibinfo{person}{Volodymyr Mnih}, \bibinfo{person}{Koray Kavukcuoglu}, \bibinfo{person}{David Silver}, \bibinfo{person}{Andrei~A. Rusu}, \bibinfo{person}{Joel Veness}, \bibinfo{person}{Marc~G. Bellemare}, \bibinfo{person}{Alex Graves}, \bibinfo{person}{Martin~A. Riedmiller}, \bibinfo{person}{Andreas Fidjeland}, \bibinfo{person}{Georg Ostrovski}, \bibinfo{person}{Stig Petersen}, \bibinfo{person}{Charles Beattie}, \bibinfo{person}{Amir Sadik}, \bibinfo{person}{Ioannis Antonoglou}, \bibinfo{person}{Helen King}, \bibinfo{person}{Dharshan Kumaran}, \bibinfo{person}{Daan Wierstra}, \bibinfo{person}{Shane Legg}, {and} \bibinfo{person}{Demis Hassabis}.} \bibinfo{year}{2015}\natexlab{}.
\newblock \showarticletitle{Human-level control through deep reinforcement learning}.
\newblock \bibinfo{journal}{\emph{Nat.}} \bibinfo{volume}{518}, \bibinfo{number}{7540} (\bibinfo{year}{2015}), \bibinfo{pages}{529--533}.
\newblock
\urldef\tempurl%
\url{https://doi.org/10.1038/NATURE14236}
\showDOI{\tempurl}


\bibitem[\protect\citeauthoryear{Narasimhan, Kulkarni, and Barzilay}{Narasimhan et~al\mbox{.}}{2015}]%
        {textgame_rl}
\bibfield{author}{\bibinfo{person}{Karthik Narasimhan}, \bibinfo{person}{Tejas~D. Kulkarni}, {and} \bibinfo{person}{Regina Barzilay}.} \bibinfo{year}{2015}\natexlab{}.
\newblock \showarticletitle{Language Understanding for Text-based Games using Deep Reinforcement Learning}. In \bibinfo{booktitle}{\emph{Proceedings of the 2015 Conference on Empirical Methods in Natural Language Processing, {EMNLP} 2015, Lisbon, Portugal, September 17-21, 2015}}, \bibfield{editor}{\bibinfo{person}{Llu{\'{\i}}s M{\`{a}}rquez}, \bibinfo{person}{Chris Callison{-}Burch}, \bibinfo{person}{Jian Su}, \bibinfo{person}{Daniele Pighin}, {and} \bibinfo{person}{Yuval Marton}} (Eds.). \bibinfo{publisher}{The Association for Computational Linguistics}, \bibinfo{pages}{1--11}.
\newblock
\urldef\tempurl%
\url{https://doi.org/10.18653/V1/D15-1001}
\showDOI{\tempurl}


\bibitem[\protect\citeauthoryear{OpenAI}{OpenAI}{2024}]%
        {2409_openai_o1}
\bibfield{author}{\bibinfo{person}{OpenAI}.} \bibinfo{year}{2024}\natexlab{}.
\newblock \bibinfo{title}{Learning to Reason with LLMs}.
\newblock
\newblock
\urldef\tempurl%
\url{https://openai.com/index/learning-to-reason-with-llms}
\showURL{%
\tempurl}


\bibitem[\protect\citeauthoryear{Peng, Kumar, Zhang, and Levine}{Peng et~al\mbox{.}}{2019}]%
        {Advantage-Weighted-Regression}
\bibfield{author}{\bibinfo{person}{Xue~Bin Peng}, \bibinfo{person}{Aviral Kumar}, \bibinfo{person}{Grace Zhang}, {and} \bibinfo{person}{Sergey Levine}.} \bibinfo{year}{2019}\natexlab{}.
\newblock \showarticletitle{Advantage-Weighted Regression: Simple and Scalable Off-Policy Reinforcement Learning}.
\newblock \bibinfo{journal}{\emph{CoRR}}  \bibinfo{volume}{abs/1910.00177} (\bibinfo{year}{2019}).
\newblock
\showeprint[arXiv]{1910.00177}
\urldef\tempurl%
\url{http://arxiv.org/abs/1910.00177}
\showURL{%
\tempurl}


\bibitem[\protect\citeauthoryear{Phan, Gatti, Han, Li, Hu, Zhang, Zhang, Shaaban, Ling, Shi, et~al\mbox{.}}{Phan et~al\mbox{.}}{2025}]%
        {HLE}
\bibfield{author}{\bibinfo{person}{Long Phan}, \bibinfo{person}{Alice Gatti}, \bibinfo{person}{Ziwen Han}, \bibinfo{person}{Nathaniel Li}, \bibinfo{person}{Josephina Hu}, \bibinfo{person}{Hugh Zhang}, \bibinfo{person}{Chen Bo~Calvin Zhang}, \bibinfo{person}{Mohamed Shaaban}, \bibinfo{person}{John Ling}, \bibinfo{person}{Sean Shi}, {et~al\mbox{.}}} \bibinfo{year}{2025}\natexlab{}.
\newblock \showarticletitle{Humanity's last exam}.
\newblock \bibinfo{journal}{\emph{arXiv preprint arXiv:2501.14249}} (\bibinfo{year}{2025}).
\newblock


\bibitem[\protect\citeauthoryear{Press, Zhang, Min, Schmidt, Smith, and Lewis}{Press et~al\mbox{.}}{2023}]%
        {bamboogle}
\bibfield{author}{\bibinfo{person}{Ofir Press}, \bibinfo{person}{Muru Zhang}, \bibinfo{person}{Sewon Min}, \bibinfo{person}{Ludwig Schmidt}, \bibinfo{person}{Noah~A. Smith}, {and} \bibinfo{person}{Mike Lewis}.} \bibinfo{year}{2023}\natexlab{}.
\newblock \showarticletitle{Measuring and Narrowing the Compositionality Gap in Language Models}. In \bibinfo{booktitle}{\emph{Findings of the Association for Computational Linguistics: {EMNLP} 2023, Singapore, December 6-10, 2023}}, \bibfield{editor}{\bibinfo{person}{Houda Bouamor}, \bibinfo{person}{Juan Pino}, {and} \bibinfo{person}{Kalika Bali}} (Eds.). \bibinfo{publisher}{Association for Computational Linguistics}, \bibinfo{pages}{5687--5711}.
\newblock
\urldef\tempurl%
\url{https://doi.org/10.18653/V1/2023.FINDINGS-EMNLP.378}
\showDOI{\tempurl}


\bibitem[\protect\citeauthoryear{Qian, Acikgoz, He, Wang, Chen, Hakkani-T{\"u}r, Tur, and Ji}{Qian et~al\mbox{.}}{2025}]%
        {qian2025toolrl}
\bibfield{author}{\bibinfo{person}{Cheng Qian}, \bibinfo{person}{Emre~Can Acikgoz}, \bibinfo{person}{Qi He}, \bibinfo{person}{Hongru Wang}, \bibinfo{person}{Xiusi Chen}, \bibinfo{person}{Dilek Hakkani-T{\"u}r}, \bibinfo{person}{Gokhan Tur}, {and} \bibinfo{person}{Heng Ji}.} \bibinfo{year}{2025}\natexlab{}.
\newblock \showarticletitle{Toolrl: Reward is all tool learning needs}.
\newblock \bibinfo{journal}{\emph{arXiv preprint arXiv:2504.13958}} (\bibinfo{year}{2025}).
\newblock


\bibitem[\protect\citeauthoryear{Qiao, Tan, Dong, MinhuiWu, Wang, Zhang, GongQue, Sun, Xu, Xue, et~al\mbox{.}}{Qiao et~al\mbox{.}}{2025b}]%
        {qiao2025v}
\bibfield{author}{\bibinfo{person}{Runqi Qiao}, \bibinfo{person}{Qiuna Tan}, \bibinfo{person}{Guanting Dong}, \bibinfo{person}{MinhuiWu MinhuiWu}, \bibinfo{person}{Jiapeng Wang}, \bibinfo{person}{Yifan Zhang}, \bibinfo{person}{Zhuoma GongQue}, \bibinfo{person}{Chong Sun}, \bibinfo{person}{Yida Xu}, \bibinfo{person}{Yadong Xue}, {et~al\mbox{.}}} \bibinfo{year}{2025}\natexlab{b}.
\newblock \showarticletitle{V-oracle: Making progressive reasoning in deciphering oracle bones for you and me}. In \bibinfo{booktitle}{\emph{Proceedings of the 63rd Annual Meeting of the Association for Computational Linguistics (Volume 1: Long Papers)}}. \bibinfo{pages}{20124--20150}.
\newblock


\bibitem[\protect\citeauthoryear{Qiao, Tan, Dong, Wu, Sun, Song, Gongque, Lei, Wei, Zhang, Qiao, Zhang, Zong, Xu, Diao, Bao, Li, and Zhang}{Qiao et~al\mbox{.}}{2024}]%
        {wemath}
\bibfield{author}{\bibinfo{person}{Runqi Qiao}, \bibinfo{person}{Qiuna Tan}, \bibinfo{person}{Guanting Dong}, \bibinfo{person}{Minhui Wu}, \bibinfo{person}{Chong Sun}, \bibinfo{person}{Xiaoshuai Song}, \bibinfo{person}{Zhuoma Gongque}, \bibinfo{person}{Shanglin Lei}, \bibinfo{person}{Zhe Wei}, \bibinfo{person}{Miaoxuan Zhang}, \bibinfo{person}{Runfeng Qiao}, \bibinfo{person}{Yifan Zhang}, \bibinfo{person}{Xiao Zong}, \bibinfo{person}{Yida Xu}, \bibinfo{person}{Muxi Diao}, \bibinfo{person}{Zhimin Bao}, \bibinfo{person}{Chen Li}, {and} \bibinfo{person}{Honggang Zhang}.} \bibinfo{year}{2024}\natexlab{}.
\newblock \showarticletitle{We-Math: Does Your Large Multimodal Model Achieve Human-like Mathematical Reasoning?}
\newblock \bibinfo{journal}{\emph{CoRR}}  \bibinfo{volume}{abs/2407.01284} (\bibinfo{year}{2024}).
\newblock
\urldef\tempurl%
\url{https://doi.org/10.48550/ARXIV.2407.01284}
\showDOI{\tempurl}
\showeprint[arXiv]{2407.01284}


\bibitem[\protect\citeauthoryear{Qiao, Tan, Yang, Wang, Wang, Wan, Zhou, Dong, Zeng, Xu, Wang, Sun, Li, and Zhang}{Qiao et~al\mbox{.}}{2025c}]%
        {We-Math2.0}
\bibfield{author}{\bibinfo{person}{Runqi Qiao}, \bibinfo{person}{Qiuna Tan}, \bibinfo{person}{Peiqing Yang}, \bibinfo{person}{Yanzi Wang}, \bibinfo{person}{Xiaowan Wang}, \bibinfo{person}{Enhui Wan}, \bibinfo{person}{Sitong Zhou}, \bibinfo{person}{Guanting Dong}, \bibinfo{person}{Yuchen Zeng}, \bibinfo{person}{Yida Xu}, \bibinfo{person}{Jie Wang}, \bibinfo{person}{Chong Sun}, \bibinfo{person}{Chen Li}, {and} \bibinfo{person}{Honggang Zhang}.} \bibinfo{year}{2025}\natexlab{c}.
\newblock \showarticletitle{We-Math 2.0: {A} Versatile MathBook System for Incentivizing Visual Mathematical Reasoning}.
\newblock \bibinfo{journal}{\emph{CoRR}}  \bibinfo{volume}{abs/2508.10433} (\bibinfo{year}{2025}).
\newblock
\urldef\tempurl%
\url{https://doi.org/10.48550/ARXIV.2508.10433}
\showDOI{\tempurl}
\showeprint[arXiv]{2508.10433}


\bibitem[\protect\citeauthoryear{Qiao, Chen, Chen, Yu, Yin, Wang, Zhang, Li, Yin, Li, et~al\mbox{.}}{Qiao et~al\mbox{.}}{2025a}]%
        {qiao2025webresearcher}
\bibfield{author}{\bibinfo{person}{Zile Qiao}, \bibinfo{person}{Guoxin Chen}, \bibinfo{person}{Xuanzhong Chen}, \bibinfo{person}{Donglei Yu}, \bibinfo{person}{Wenbiao Yin}, \bibinfo{person}{Xinyu Wang}, \bibinfo{person}{Zhen Zhang}, \bibinfo{person}{Baixuan Li}, \bibinfo{person}{Huifeng Yin}, \bibinfo{person}{Kuan Li}, {et~al\mbox{.}}} \bibinfo{year}{2025}\natexlab{a}.
\newblock \showarticletitle{WebResearcher: Unleashing unbounded reasoning capability in Long-Horizon Agents}.
\newblock \bibinfo{journal}{\emph{arXiv preprint arXiv:2509.13309}} (\bibinfo{year}{2025}).
\newblock


\bibitem[\protect\citeauthoryear{Qin, Ye, Fang, Wang, Liang, Tian, Zhang, Li, Li, Huang, et~al\mbox{.}}{Qin et~al\mbox{.}}{2025}]%
        {qin2025ui}
\bibfield{author}{\bibinfo{person}{Yujia Qin}, \bibinfo{person}{Yining Ye}, \bibinfo{person}{Junjie Fang}, \bibinfo{person}{Haoming Wang}, \bibinfo{person}{Shihao Liang}, \bibinfo{person}{Shizuo Tian}, \bibinfo{person}{Junda Zhang}, \bibinfo{person}{Jiahao Li}, \bibinfo{person}{Yunxin Li}, \bibinfo{person}{Shijue Huang}, {et~al\mbox{.}}} \bibinfo{year}{2025}\natexlab{}.
\newblock \showarticletitle{UI-TARS: Pioneering Automated GUI Interaction with Native Agents}.
\newblock \bibinfo{journal}{\emph{arXiv preprint arXiv:2501.12326}} (\bibinfo{year}{2025}).
\newblock


\bibitem[\protect\citeauthoryear{Qwen, :, Yang, Yang, Zhang, Hui, Zheng, Yu, Li, Liu, Huang, Wei, Lin, Yang, Tu, Zhang, Yang, Yang, Zhou, Lin, Dang, Lu, Bao, Yang, Yu, Li, Xue, Zhang, Zhu, Men, Lin, Li, Xia, Ren, Ren, Fan, Su, Zhang, Wan, Liu, Cui, Zhang, and Qiu}{Qwen et~al\mbox{.}}{2024}]%
        {qwen2.5}
\bibfield{author}{\bibinfo{person}{Qwen}, \bibinfo{person}{:}, \bibinfo{person}{An Yang}, \bibinfo{person}{Baosong Yang}, \bibinfo{person}{Beichen Zhang}, \bibinfo{person}{Binyuan Hui}, \bibinfo{person}{Bo Zheng}, \bibinfo{person}{Bowen Yu}, \bibinfo{person}{Chengyuan Li}, \bibinfo{person}{Dayiheng Liu}, \bibinfo{person}{Fei Huang}, \bibinfo{person}{Haoran Wei}, \bibinfo{person}{Huan Lin}, \bibinfo{person}{Jian Yang}, \bibinfo{person}{Jianhong Tu}, \bibinfo{person}{Jianwei Zhang}, \bibinfo{person}{Jianxin Yang}, \bibinfo{person}{Jiaxi Yang}, \bibinfo{person}{Jingren Zhou}, \bibinfo{person}{Junyang Lin}, \bibinfo{person}{Kai Dang}, \bibinfo{person}{Keming Lu}, \bibinfo{person}{Keqin Bao}, \bibinfo{person}{Kexin Yang}, \bibinfo{person}{Le Yu}, \bibinfo{person}{Mei Li}, \bibinfo{person}{Mingfeng Xue}, \bibinfo{person}{Pei Zhang}, \bibinfo{person}{Qin Zhu}, \bibinfo{person}{Rui Men}, \bibinfo{person}{Runji Lin}, \bibinfo{person}{Tianhao Li}, \bibinfo{person}{Tingyu Xia}, \bibinfo{person}{Xingzhang Ren},
  \bibinfo{person}{Xuancheng Ren}, \bibinfo{person}{Yang Fan}, \bibinfo{person}{Yang Su}, \bibinfo{person}{Yichang Zhang}, \bibinfo{person}{Yu Wan}, \bibinfo{person}{Yuqiong Liu}, \bibinfo{person}{Zeyu Cui}, \bibinfo{person}{Zhenru Zhang}, {and} \bibinfo{person}{Zihan Qiu}.} \bibinfo{year}{2024}\natexlab{}.
\newblock \bibinfo{title}{Qwen2.5 Technical Report}.
\newblock
\newblock
\showeprint[arxiv]{cs.CL/2412.15115}
\urldef\tempurl%
\url{https://arxiv.org/abs/2412.15115}
\showURL{%
\tempurl}


\bibitem[\protect\citeauthoryear{Schulman, Wolski, Dhariwal, Radford, and Klimov}{Schulman et~al\mbox{.}}{2017}]%
        {ppo}
\bibfield{author}{\bibinfo{person}{John Schulman}, \bibinfo{person}{Filip Wolski}, \bibinfo{person}{Prafulla Dhariwal}, \bibinfo{person}{Alec Radford}, {and} \bibinfo{person}{Oleg Klimov}.} \bibinfo{year}{2017}\natexlab{}.
\newblock \showarticletitle{Proximal Policy Optimization Algorithms}.
\newblock \bibinfo{journal}{\emph{CoRR}}  \bibinfo{volume}{abs/1707.06347} (\bibinfo{year}{2017}).
\newblock
\showeprint[arXiv]{1707.06347}
\urldef\tempurl%
\url{http://arxiv.org/abs/1707.06347}
\showURL{%
\tempurl}


\bibitem[\protect\citeauthoryear{Shao, Wang, Zhu, Xu, Song, Zhang, Li, Wu, and Guo}{Shao et~al\mbox{.}}{2024}]%
        {deepseekmath}
\bibfield{author}{\bibinfo{person}{Zhihong Shao}, \bibinfo{person}{Peiyi Wang}, \bibinfo{person}{Qihao Zhu}, \bibinfo{person}{Runxin Xu}, \bibinfo{person}{Junxiao Song}, \bibinfo{person}{Mingchuan Zhang}, \bibinfo{person}{Y.~K. Li}, \bibinfo{person}{Y. Wu}, {and} \bibinfo{person}{Daya Guo}.} \bibinfo{year}{2024}\natexlab{}.
\newblock \showarticletitle{DeepSeekMath: Pushing the Limits of Mathematical Reasoning in Open Language Models}.
\newblock \bibinfo{journal}{\emph{CoRR}}  \bibinfo{volume}{abs/2402.03300} (\bibinfo{year}{2024}).
\newblock
\urldef\tempurl%
\url{https://doi.org/10.48550/ARXIV.2402.03300}
\showDOI{\tempurl}
\showeprint[arXiv]{2402.03300}


\bibitem[\protect\citeauthoryear{Sheng, Zhang, Ye, Wu, Zhang, Zhang, Peng, Lin, and Wu}{Sheng et~al\mbox{.}}{2024}]%
        {sheng2024hybridflow}
\bibfield{author}{\bibinfo{person}{Guangming Sheng}, \bibinfo{person}{Chi Zhang}, \bibinfo{person}{Zilingfeng Ye}, \bibinfo{person}{Xibin Wu}, \bibinfo{person}{Wang Zhang}, \bibinfo{person}{Ru Zhang}, \bibinfo{person}{Yanghua Peng}, \bibinfo{person}{Haibin Lin}, {and} \bibinfo{person}{Chuan Wu}.} \bibinfo{year}{2024}\natexlab{}.
\newblock \showarticletitle{HybridFlow: A Flexible and Efficient RLHF Framework}.
\newblock \bibinfo{journal}{\emph{arXiv preprint arXiv: 2409.19256}} (\bibinfo{year}{2024}).
\newblock


\bibitem[\protect\citeauthoryear{Shridhar, Yuan, C{\^o}t{\'e}, Bisk, Trischler, and Hausknecht}{Shridhar et~al\mbox{.}}{2020}]%
        {shridhar2020alfworld}
\bibfield{author}{\bibinfo{person}{Mohit Shridhar}, \bibinfo{person}{Xingdi Yuan}, \bibinfo{person}{Marc-Alexandre C{\^o}t{\'e}}, \bibinfo{person}{Yonatan Bisk}, \bibinfo{person}{Adam Trischler}, {and} \bibinfo{person}{Matthew Hausknecht}.} \bibinfo{year}{2020}\natexlab{}.
\newblock \showarticletitle{Alfworld: Aligning text and embodied environments for interactive learning}.
\newblock \bibinfo{journal}{\emph{arXiv preprint arXiv:2010.03768}} (\bibinfo{year}{2020}).
\newblock


\bibitem[\protect\citeauthoryear{Silver, Hubert, Schrittwieser, Antonoglou, Lai, Guez, Lanctot, Sifre, Kumaran, Graepel, Lillicrap, Simonyan, and Hassabis}{Silver et~al\mbox{.}}{2017}]%
        {alphazero}
\bibfield{author}{\bibinfo{person}{David Silver}, \bibinfo{person}{Thomas Hubert}, \bibinfo{person}{Julian Schrittwieser}, \bibinfo{person}{Ioannis Antonoglou}, \bibinfo{person}{Matthew Lai}, \bibinfo{person}{Arthur Guez}, \bibinfo{person}{Marc Lanctot}, \bibinfo{person}{Laurent Sifre}, \bibinfo{person}{Dharshan Kumaran}, \bibinfo{person}{Thore Graepel}, \bibinfo{person}{Timothy~P. Lillicrap}, \bibinfo{person}{Karen Simonyan}, {and} \bibinfo{person}{Demis Hassabis}.} \bibinfo{year}{2017}\natexlab{}.
\newblock \showarticletitle{Mastering Chess and Shogi by Self-Play with a General Reinforcement Learning Algorithm}.
\newblock \bibinfo{journal}{\emph{CoRR}}  \bibinfo{volume}{abs/1712.01815} (\bibinfo{year}{2017}).
\newblock
\showeprint[arXiv]{1712.01815}
\urldef\tempurl%
\url{http://arxiv.org/abs/1712.01815}
\showURL{%
\tempurl}


\bibitem[\protect\citeauthoryear{Singh, Magazine, Pandya, and Nambi}{Singh et~al\mbox{.}}{2025}]%
        {2505_ARTIST}
\bibfield{author}{\bibinfo{person}{Joykirat Singh}, \bibinfo{person}{Raghav Magazine}, \bibinfo{person}{Yash Pandya}, {and} \bibinfo{person}{Akshay Nambi}.} \bibinfo{year}{2025}\natexlab{}.
\newblock \showarticletitle{Agentic Reasoning and Tool Integration for LLMs via Reinforcement Learning}.
\newblock \bibinfo{journal}{\emph{arXiv preprint arXiv:2505.01441}} (\bibinfo{year}{2025}).
\newblock


\bibitem[\protect\citeauthoryear{Song, Jiang, Min, Chen, Chen, Zhao, Fang, and Wen}{Song et~al\mbox{.}}{2025a}]%
        {r1searcher}
\bibfield{author}{\bibinfo{person}{Huatong Song}, \bibinfo{person}{Jinhao Jiang}, \bibinfo{person}{Yingqian Min}, \bibinfo{person}{Jie Chen}, \bibinfo{person}{Zhipeng Chen}, \bibinfo{person}{Wayne~Xin Zhao}, \bibinfo{person}{Lei Fang}, {and} \bibinfo{person}{Ji{-}Rong Wen}.} \bibinfo{year}{2025}\natexlab{a}.
\newblock \showarticletitle{R1-Searcher: Incentivizing the Search Capability in LLMs via Reinforcement Learning}.
\newblock \bibinfo{journal}{\emph{CoRR}}  \bibinfo{volume}{abs/2503.05592} (\bibinfo{year}{2025}).
\newblock
\urldef\tempurl%
\url{https://doi.org/10.48550/ARXIV.2503.05592}
\showDOI{\tempurl}
\showeprint[arXiv]{2503.05592}


\bibitem[\protect\citeauthoryear{Song, Jiang, Tian, Chen, Wu, Zhao, Min, Zhao, Fang, and Wen}{Song et~al\mbox{.}}{2025b}]%
        {r1_searcher++}
\bibfield{author}{\bibinfo{person}{Huatong Song}, \bibinfo{person}{Jinhao Jiang}, \bibinfo{person}{Wenqing Tian}, \bibinfo{person}{Zhipeng Chen}, \bibinfo{person}{Yuhuan Wu}, \bibinfo{person}{Jiahao Zhao}, \bibinfo{person}{Yingqian Min}, \bibinfo{person}{Wayne~Xin Zhao}, \bibinfo{person}{Lei Fang}, {and} \bibinfo{person}{Ji{-}Rong Wen}.} \bibinfo{year}{2025}\natexlab{b}.
\newblock \showarticletitle{R1-Searcher++: Incentivizing the Dynamic Knowledge Acquisition of LLMs via Reinforcement Learning}.
\newblock \bibinfo{journal}{\emph{CoRR}}  \bibinfo{volume}{abs/2505.17005} (\bibinfo{year}{2025}).
\newblock
\urldef\tempurl%
\url{https://doi.org/10.48550/ARXIV.2505.17005}
\showDOI{\tempurl}
\showeprint[arXiv]{2505.17005}


\bibitem[\protect\citeauthoryear{Su, Zhang, Li, Chen, Wang, Song, Wang, Li, Wu, Chen, et~al\mbox{.}}{Su et~al\mbox{.}}{2025c}]%
        {su2025ScalingAgentsViaContinualPre-training}
\bibfield{author}{\bibinfo{person}{Liangcai Su}, \bibinfo{person}{Zhen Zhang}, \bibinfo{person}{Guangyu Li}, \bibinfo{person}{Zhuo Chen}, \bibinfo{person}{Chenxi Wang}, \bibinfo{person}{Maojia Song}, \bibinfo{person}{Xinyu Wang}, \bibinfo{person}{Kuan Li}, \bibinfo{person}{Jialong Wu}, \bibinfo{person}{Xuanzhong Chen}, {et~al\mbox{.}}} \bibinfo{year}{2025}\natexlab{c}.
\newblock \showarticletitle{Scaling agents via continual pre-training}.
\newblock \bibinfo{journal}{\emph{arXiv preprint arXiv:2509.13310}} (\bibinfo{year}{2025}).
\newblock


\bibitem[\protect\citeauthoryear{Su, Pan, Bai, Liu, Dong, Huang, Hu, Zhang, Gai, and Zhou}{Su et~al\mbox{.}}{2025a}]%
        {klear}
\bibfield{author}{\bibinfo{person}{Zhenpeng Su}, \bibinfo{person}{Leiyu Pan}, \bibinfo{person}{Xue Bai}, \bibinfo{person}{Dening Liu}, \bibinfo{person}{Guanting Dong}, \bibinfo{person}{Jiaming Huang}, \bibinfo{person}{Wenping Hu}, \bibinfo{person}{Fuzheng Zhang}, \bibinfo{person}{Kun Gai}, {and} \bibinfo{person}{Guorui Zhou}.} \bibinfo{year}{2025}\natexlab{a}.
\newblock \showarticletitle{Klear-Reasoner: Advancing Reasoning Capability via Gradient-Preserving Clipping Policy Optimization}.
\newblock \bibinfo{journal}{\emph{CoRR}}  \bibinfo{volume}{abs/2508.07629} (\bibinfo{year}{2025}).
\newblock
\urldef\tempurl%
\url{https://doi.org/10.48550/ARXIV.2508.07629}
\showDOI{\tempurl}
\showeprint[arXiv]{2508.07629}


\bibitem[\protect\citeauthoryear{Su, Pan, Lv, Li, Hu, Zhang, Gai, and Zhou}{Su et~al\mbox{.}}{2025b}]%
        {su2025gppo}
\bibfield{author}{\bibinfo{person}{Zhenpeng Su}, \bibinfo{person}{Leiyu Pan}, \bibinfo{person}{Minxuan Lv}, \bibinfo{person}{Yuntao Li}, \bibinfo{person}{Wenping Hu}, \bibinfo{person}{Fuzheng Zhang}, \bibinfo{person}{Kun Gai}, {and} \bibinfo{person}{Guorui Zhou}.} \bibinfo{year}{2025}\natexlab{b}.
\newblock \showarticletitle{CE-GPPO: Controlling Entropy via Gradient-Preserving Clipping Policy Optimization in Reinforcement Learning}.
\newblock \bibinfo{journal}{\emph{arXiv preprint arXiv:2509.20712}} (\bibinfo{year}{2025}).
\newblock


\bibitem[\protect\citeauthoryear{Sun, Qiao, Guo, Fan, Hou, Jiang, Xie, Zhang, Huang, and Zhou}{Sun et~al\mbox{.}}{2025}]%
        {zerosearch}
\bibfield{author}{\bibinfo{person}{Hao Sun}, \bibinfo{person}{Zile Qiao}, \bibinfo{person}{Jiayan Guo}, \bibinfo{person}{Xuanbo Fan}, \bibinfo{person}{Yingyan Hou}, \bibinfo{person}{Yong Jiang}, \bibinfo{person}{Pengjun Xie}, \bibinfo{person}{Yan Zhang}, \bibinfo{person}{Fei Huang}, {and} \bibinfo{person}{Jingren Zhou}.} \bibinfo{year}{2025}\natexlab{}.
\newblock \bibinfo{title}{ZeroSearch: Incentivize the Search Capability of LLMs without Searching}.
\newblock
\newblock
\showeprint[arxiv]{cs.CL/2505.04588}
\urldef\tempurl%
\url{https://arxiv.org/abs/2505.04588}
\showURL{%
\tempurl}


\bibitem[\protect\citeauthoryear{Tan, Dou, Wang, Wang, Chen, and Wen}{Tan et~al\mbox{.}}{2024a}]%
        {2411_htmlrag}
\bibfield{author}{\bibinfo{person}{Jiejun Tan}, \bibinfo{person}{Zhicheng Dou}, \bibinfo{person}{Wen Wang}, \bibinfo{person}{Mang Wang}, \bibinfo{person}{Weipeng Chen}, {and} \bibinfo{person}{Ji{-}Rong Wen}.} \bibinfo{year}{2024}\natexlab{a}.
\newblock \showarticletitle{HtmlRAG: {HTML} is Better Than Plain Text for Modeling Retrieved Knowledge in {RAG} Systems}.
\newblock \bibinfo{journal}{\emph{CoRR}}  \bibinfo{volume}{abs/2411.02959} (\bibinfo{year}{2024}).
\newblock
\urldef\tempurl%
\url{https://doi.org/10.48550/ARXIV.2411.02959}
\showDOI{\tempurl}
\showeprint[arXiv]{2411.02959}


\bibitem[\protect\citeauthoryear{Tan, Dou, Yu, Cheng, Ju, Xie, and Wen}{Tan et~al\mbox{.}}{2025}]%
        {tan2025hiersearch}
\bibfield{author}{\bibinfo{person}{Jiejun Tan}, \bibinfo{person}{Zhicheng Dou}, \bibinfo{person}{Yan Yu}, \bibinfo{person}{Jiehan Cheng}, \bibinfo{person}{Qiang Ju}, \bibinfo{person}{Jian Xie}, {and} \bibinfo{person}{Ji-Rong Wen}.} \bibinfo{year}{2025}\natexlab{}.
\newblock \showarticletitle{HierSearch: A Hierarchical Enterprise Deep Search Framework Integrating Local and Web Searches}.
\newblock \bibinfo{journal}{\emph{arXiv preprint arXiv:2508.08088}} (\bibinfo{year}{2025}).
\newblock


\bibitem[\protect\citeauthoryear{Tan, Zhang, Liu, Zheng, Wang, and An}{Tan et~al\mbox{.}}{2024b}]%
        {agentrl_2}
\bibfield{author}{\bibinfo{person}{Weihao Tan}, \bibinfo{person}{Wentao Zhang}, \bibinfo{person}{Shanqi Liu}, \bibinfo{person}{Longtao Zheng}, \bibinfo{person}{Xinrun Wang}, {and} \bibinfo{person}{Bo An}.} \bibinfo{year}{2024}\natexlab{b}.
\newblock \showarticletitle{True Knowledge Comes from Practice: Aligning Large Language Models with Embodied Environments via Reinforcement Learning}. In \bibinfo{booktitle}{\emph{The Twelfth International Conference on Learning Representations, {ICLR} 2024, Vienna, Austria, May 7-11, 2024}}. \bibinfo{publisher}{OpenReview.net}.
\newblock
\urldef\tempurl%
\url{https://openreview.net/forum?id=hILVmJ4Uvu}
\showURL{%
\tempurl}


\bibitem[\protect\citeauthoryear{Tao, Wu, Yin, Zhang, Li, Shen, Li, Zhang, Wang, Jiang, Xie, Huang, and Zhou}{Tao et~al\mbox{.}}{2025}]%
        {WebShaper}
\bibfield{author}{\bibinfo{person}{Zhengwei Tao}, \bibinfo{person}{Jialong Wu}, \bibinfo{person}{Wenbiao Yin}, \bibinfo{person}{Junkai Zhang}, \bibinfo{person}{Baixuan Li}, \bibinfo{person}{Haiyang Shen}, \bibinfo{person}{Kuan Li}, \bibinfo{person}{Liwen Zhang}, \bibinfo{person}{Xinyu Wang}, \bibinfo{person}{Yong Jiang}, \bibinfo{person}{Pengjun Xie}, \bibinfo{person}{Fei Huang}, {and} \bibinfo{person}{Jingren Zhou}.} \bibinfo{year}{2025}\natexlab{}.
\newblock \showarticletitle{WebShaper: Agentically Data Synthesizing via Information-Seeking Formalization}.
\newblock \bibinfo{journal}{\emph{CoRR}}  \bibinfo{volume}{abs/2507.15061} (\bibinfo{year}{2025}).
\newblock
\urldef\tempurl%
\url{https://doi.org/10.48550/ARXIV.2507.15061}
\showDOI{\tempurl}
\showeprint[arXiv]{2507.15061}


\bibitem[\protect\citeauthoryear{Team, Bai, Bao, Chen, Chen, Chen, Chen, Chen, Chen, Chen, et~al\mbox{.}}{Team et~al\mbox{.}}{2025a}]%
        {team2025kimi2}
\bibfield{author}{\bibinfo{person}{Kimi Team}, \bibinfo{person}{Yifan Bai}, \bibinfo{person}{Yiping Bao}, \bibinfo{person}{Guanduo Chen}, \bibinfo{person}{Jiahao Chen}, \bibinfo{person}{Ningxin Chen}, \bibinfo{person}{Ruijue Chen}, \bibinfo{person}{Yanru Chen}, \bibinfo{person}{Yuankun Chen}, \bibinfo{person}{Yutian Chen}, {et~al\mbox{.}}} \bibinfo{year}{2025}\natexlab{a}.
\newblock \showarticletitle{Kimi k2: Open agentic intelligence}.
\newblock \bibinfo{journal}{\emph{arXiv preprint arXiv:2507.20534}} (\bibinfo{year}{2025}).
\newblock


\bibitem[\protect\citeauthoryear{Team, Du, Gao, Xing, Jiang, Chen, Li, Xiao, Du, Liao, et~al\mbox{.}}{Team et~al\mbox{.}}{2025b}]%
        {team2025kimi}
\bibfield{author}{\bibinfo{person}{Kimi Team}, \bibinfo{person}{Angang Du}, \bibinfo{person}{Bofei Gao}, \bibinfo{person}{Bowei Xing}, \bibinfo{person}{Changjiu Jiang}, \bibinfo{person}{Cheng Chen}, \bibinfo{person}{Cheng Li}, \bibinfo{person}{Chenjun Xiao}, \bibinfo{person}{Chenzhuang Du}, \bibinfo{person}{Chonghua Liao}, {et~al\mbox{.}}} \bibinfo{year}{2025}\natexlab{b}.
\newblock \showarticletitle{Kimi k1. 5: Scaling reinforcement learning with llms}.
\newblock \bibinfo{journal}{\emph{arXiv preprint arXiv:2501.12599}} (\bibinfo{year}{2025}).
\newblock


\bibitem[\protect\citeauthoryear{Team, Li, Lei, Wang, Rong, Wang, Zhang, Gao, Zhang, Sun, et~al\mbox{.}}{Team et~al\mbox{.}}{2025c}]%
        {team2025longcat}
\bibfield{author}{\bibinfo{person}{Meituan~LongCat Team}, \bibinfo{person}{Bei Li}, \bibinfo{person}{Bingye Lei}, \bibinfo{person}{Bo Wang}, \bibinfo{person}{Bolin Rong}, \bibinfo{person}{Chao Wang}, \bibinfo{person}{Chao Zhang}, \bibinfo{person}{Chen Gao}, \bibinfo{person}{Chen Zhang}, \bibinfo{person}{Cheng Sun}, {et~al\mbox{.}}} \bibinfo{year}{2025}\natexlab{c}.
\newblock \showarticletitle{LongCat-Flash Technical Report}.
\newblock \bibinfo{journal}{\emph{arXiv preprint arXiv:2509.01322}} (\bibinfo{year}{2025}).
\newblock


\bibitem[\protect\citeauthoryear{Team}{Team}{2024a}]%
        {qwq-32b-preview}
\bibfield{author}{\bibinfo{person}{Qwen Team}.} \bibinfo{year}{2024}\natexlab{a}.
\newblock \bibinfo{title}{QwQ: Reflect Deeply on the Boundaries of the Unknown}.
\newblock
\newblock
\urldef\tempurl%
\url{https://qwenlm.github.io/blog/qwq-32b-preview/}
\showURL{%
\tempurl}


\bibitem[\protect\citeauthoryear{Team}{Team}{2024b}]%
        {qwen_qwq}
\bibfield{author}{\bibinfo{person}{Qwen Team}.} \bibinfo{year}{2024}\natexlab{b}.
\newblock \showarticletitle{Qwq: Reflect deeply on the boundaries of the unknown}.
\newblock \bibinfo{journal}{\emph{Hugging Face}} (\bibinfo{year}{2024}).
\newblock


\bibitem[\protect\citeauthoryear{Tishby, Pereira, and Bialek}{Tishby et~al\mbox{.}}{2000}]%
        {tishby2000information}
\bibfield{author}{\bibinfo{person}{Naftali Tishby}, \bibinfo{person}{Fernando~C Pereira}, {and} \bibinfo{person}{William Bialek}.} \bibinfo{year}{2000}\natexlab{}.
\newblock \showarticletitle{The information bottleneck method}.
\newblock \bibinfo{journal}{\emph{arXiv preprint physics/0004057}} (\bibinfo{year}{2000}).
\newblock


\bibitem[\protect\citeauthoryear{Trivedi, Balasubramanian, Khot, and Sabharwal}{Trivedi et~al\mbox{.}}{2022}]%
        {musique}
\bibfield{author}{\bibinfo{person}{Harsh Trivedi}, \bibinfo{person}{Niranjan Balasubramanian}, \bibinfo{person}{Tushar Khot}, {and} \bibinfo{person}{Ashish Sabharwal}.} \bibinfo{year}{2022}\natexlab{}.
\newblock \showarticletitle{MuSiQue: Multihop Questions via Single-hop Question Composition}.
\newblock \bibinfo{journal}{\emph{Transactions of the Association for Computational Linguistics}}  \bibinfo{volume}{10} (\bibinfo{year}{2022}), \bibinfo{pages}{539--554}.
\newblock


\bibitem[\protect\citeauthoryear{Wang, Qian, Li, Qiu, Xue, Wang, Ji, and Wong}{Wang et~al\mbox{.}}{2025a}]%
        {Tool-Use}
\bibfield{author}{\bibinfo{person}{Hongru Wang}, \bibinfo{person}{Cheng Qian}, \bibinfo{person}{Manling Li}, \bibinfo{person}{Jiahao Qiu}, \bibinfo{person}{Boyang Xue}, \bibinfo{person}{Mengdi Wang}, \bibinfo{person}{Heng Ji}, {and} \bibinfo{person}{Kam{-}Fai Wong}.} \bibinfo{year}{2025}\natexlab{a}.
\newblock \showarticletitle{Toward a Theory of Agents as Tool-Use Decision-Makers}.
\newblock \bibinfo{journal}{\emph{CoRR}}  \bibinfo{volume}{abs/2506.00886} (\bibinfo{year}{2025}).
\newblock
\urldef\tempurl%
\url{https://doi.org/10.48550/ARXIV.2506.00886}
\showDOI{\tempurl}
\showeprint[arXiv]{2506.00886}


\bibitem[\protect\citeauthoryear{Wang, Qian, Zhong, Chen, Qiu, Huang, Jin, Wang, Wong, and Ji}{Wang et~al\mbox{.}}{2025b}]%
        {wang2025otc}
\bibfield{author}{\bibinfo{person}{Hongru Wang}, \bibinfo{person}{Cheng Qian}, \bibinfo{person}{Wanjun Zhong}, \bibinfo{person}{Xiusi Chen}, \bibinfo{person}{Jiahao Qiu}, \bibinfo{person}{Shijue Huang}, \bibinfo{person}{Bowen Jin}, \bibinfo{person}{Mengdi Wang}, \bibinfo{person}{Kam-Fai Wong}, {and} \bibinfo{person}{Heng Ji}.} \bibinfo{year}{2025}\natexlab{b}.
\newblock \showarticletitle{OTC: Optimal Tool Calls via Reinforcement Learning}.
\newblock \bibinfo{journal}{\emph{arXiv preprint arXiv:2504.14870}} (\bibinfo{year}{2025}).
\newblock


\bibitem[\protect\citeauthoryear{Wang, Zou, Song, Feng, Fang, Lu, Liu, Luo, Liang, Huang, et~al\mbox{.}}{Wang et~al\mbox{.}}{2025f}]%
        {wang2025ui}
\bibfield{author}{\bibinfo{person}{Haoming Wang}, \bibinfo{person}{Haoyang Zou}, \bibinfo{person}{Huatong Song}, \bibinfo{person}{Jiazhan Feng}, \bibinfo{person}{Junjie Fang}, \bibinfo{person}{Junting Lu}, \bibinfo{person}{Longxiang Liu}, \bibinfo{person}{Qinyu Luo}, \bibinfo{person}{Shihao Liang}, \bibinfo{person}{Shijue Huang}, {et~al\mbox{.}}} \bibinfo{year}{2025}\natexlab{f}.
\newblock \showarticletitle{Ui-tars-2 technical report: Advancing gui agent with multi-turn reinforcement learning}.
\newblock \bibinfo{journal}{\emph{arXiv preprint arXiv:2509.02544}} (\bibinfo{year}{2025}).
\newblock


\bibitem[\protect\citeauthoryear{Wang, Yu, Gao, Zheng, Liu, Lu, Dang, Chen, Yang, Zhang, Liu, Yang, Zhao, Yue, Song, Yu, Huang, and Lin}{Wang et~al\mbox{.}}{2025e}]%
        {20/80}
\bibfield{author}{\bibinfo{person}{Shenzhi Wang}, \bibinfo{person}{Le Yu}, \bibinfo{person}{Chang Gao}, \bibinfo{person}{Chujie Zheng}, \bibinfo{person}{Shixuan Liu}, \bibinfo{person}{Rui Lu}, \bibinfo{person}{Kai Dang}, \bibinfo{person}{Xionghui Chen}, \bibinfo{person}{Jianxin Yang}, \bibinfo{person}{Zhenru Zhang}, \bibinfo{person}{Yuqiong Liu}, \bibinfo{person}{An Yang}, \bibinfo{person}{Andrew Zhao}, \bibinfo{person}{Yang Yue}, \bibinfo{person}{Shiji Song}, \bibinfo{person}{Bowen Yu}, \bibinfo{person}{Gao Huang}, {and} \bibinfo{person}{Junyang Lin}.} \bibinfo{year}{2025}\natexlab{e}.
\newblock \showarticletitle{Beyond the 80/20 Rule: High-Entropy Minority Tokens Drive Effective Reinforcement Learning for {LLM} Reasoning}.
\newblock \bibinfo{journal}{\emph{CoRR}}  \bibinfo{volume}{abs/2506.01939} (\bibinfo{year}{2025}).
\newblock
\urldef\tempurl%
\url{https://doi.org/10.48550/ARXIV.2506.01939}
\showDOI{\tempurl}
\showeprint[arXiv]{2506.01939}


\bibitem[\protect\citeauthoryear{Wang, Wu, Liu, Hao, Wang, and Shao}{Wang et~al\mbox{.}}{2024}]%
        {agent_rl_5}
\bibfield{author}{\bibinfo{person}{Taiyi Wang}, \bibinfo{person}{Zhihao Wu}, \bibinfo{person}{Jianheng Liu}, \bibinfo{person}{Jianye Hao}, \bibinfo{person}{Jun Wang}, {and} \bibinfo{person}{Kun Shao}.} \bibinfo{year}{2024}\natexlab{}.
\newblock \showarticletitle{DistRL: An Asynchronous Distributed Reinforcement Learning Framework for On-Device Control Agents}.
\newblock \bibinfo{journal}{\emph{CoRR}}  \bibinfo{volume}{abs/2410.14803} (\bibinfo{year}{2024}).
\newblock
\urldef\tempurl%
\url{https://doi.org/10.48550/ARXIV.2410.14803}
\showDOI{\tempurl}
\showeprint[arXiv]{2410.14803}


\bibitem[\protect\citeauthoryear{Wang, Yang, Zeng, Ren, Liu, Peng, Cheng, He, Wang, Gao, Chen, Wang, Du, and Shen}{Wang et~al\mbox{.}}{2025d}]%
        {wang2025reinforcement}
\bibfield{author}{\bibinfo{person}{Yiping Wang}, \bibinfo{person}{Qing Yang}, \bibinfo{person}{Zhiyuan Zeng}, \bibinfo{person}{Liliang Ren}, \bibinfo{person}{Lucas Liu}, \bibinfo{person}{Baolin Peng}, \bibinfo{person}{Hao Cheng}, \bibinfo{person}{Xuehai He}, \bibinfo{person}{Kuan Wang}, \bibinfo{person}{Jianfeng Gao}, \bibinfo{person}{Weizhu Chen}, \bibinfo{person}{Shuohang Wang}, \bibinfo{person}{Simon~Shaolei Du}, {and} \bibinfo{person}{Yelong Shen}.} \bibinfo{year}{2025}\natexlab{d}.
\newblock \showarticletitle{Reinforcement Learning for Reasoning in Large Language Models with One Training Example}.
\newblock \bibinfo{journal}{\emph{arXiv preprint arXiv:2504.20571}} (\bibinfo{year}{2025}).
\newblock


\bibitem[\protect\citeauthoryear{Wang, Wang, Wang, Zhang, Li, Yang, Yu, Nguyen, Liu, Gottlieb, Lam, Lu, Cho, Wu, Fei-Fei, Wang, Choi, and Li}{Wang et~al\mbox{.}}{2025c}]%
        {ragen}
\bibfield{author}{\bibinfo{person}{Zihan Wang}, \bibinfo{person}{Kangrui Wang}, \bibinfo{person}{Qineng Wang}, \bibinfo{person}{Pingyue Zhang}, \bibinfo{person}{Linjie Li}, \bibinfo{person}{Zhengyuan Yang}, \bibinfo{person}{Kefan Yu}, \bibinfo{person}{Minh~Nhat Nguyen}, \bibinfo{person}{Licheng Liu}, \bibinfo{person}{Eli Gottlieb}, \bibinfo{person}{Monica Lam}, \bibinfo{person}{Yiping Lu}, \bibinfo{person}{Kyunghyun Cho}, \bibinfo{person}{Jiajun Wu}, \bibinfo{person}{Li Fei-Fei}, \bibinfo{person}{Lijuan Wang}, \bibinfo{person}{Yejin Choi}, {and} \bibinfo{person}{Manling Li}.} \bibinfo{year}{2025}\natexlab{c}.
\newblock \bibinfo{title}{RAGEN: Understanding Self-Evolution in LLM Agents via Multi-Turn Reinforcement Learning}.
\newblock
\newblock
\showeprint[arxiv]{cs.LG/2504.20073}
\urldef\tempurl%
\url{https://arxiv.org/abs/2504.20073}
\showURL{%
\tempurl}


\bibitem[\protect\citeauthoryear{Wu, Li, Fang, Yin, Zhang, Tao, Zhang, Xi, Jiang, Xie, Huang, and Zhou}{Wu et~al\mbox{.}}{2025a}]%
        {wu2025webdancer}
\bibfield{author}{\bibinfo{person}{Jialong Wu}, \bibinfo{person}{Baixuan Li}, \bibinfo{person}{Runnan Fang}, \bibinfo{person}{Wenbiao Yin}, \bibinfo{person}{Liwen Zhang}, \bibinfo{person}{Zhengwei Tao}, \bibinfo{person}{Dingchu Zhang}, \bibinfo{person}{Zekun Xi}, \bibinfo{person}{Yong Jiang}, \bibinfo{person}{Pengjun Xie}, \bibinfo{person}{Fei Huang}, {and} \bibinfo{person}{Jingren Zhou}.} \bibinfo{year}{2025}\natexlab{a}.
\newblock \bibinfo{title}{WebDancer: Towards Autonomous Information Seeking Agency}.
\newblock
\newblock
\showeprint[arxiv]{cs.CL/2505.22648}
\urldef\tempurl%
\url{https://arxiv.org/abs/2505.22648}
\showURL{%
\tempurl}


\bibitem[\protect\citeauthoryear{Wu, Li, Fang, Yin, Zhang, Tao, Zhang, Xi, Jiang, Xie, Huang, and Zhou}{Wu et~al\mbox{.}}{2025b}]%
        {WebDancer}
\bibfield{author}{\bibinfo{person}{Jialong Wu}, \bibinfo{person}{Baixuan Li}, \bibinfo{person}{Runnan Fang}, \bibinfo{person}{Wenbiao Yin}, \bibinfo{person}{Liwen Zhang}, \bibinfo{person}{Zhengwei Tao}, \bibinfo{person}{Dingchu Zhang}, \bibinfo{person}{Zekun Xi}, \bibinfo{person}{Yong Jiang}, \bibinfo{person}{Pengjun Xie}, \bibinfo{person}{Fei Huang}, {and} \bibinfo{person}{Jingren Zhou}.} \bibinfo{year}{2025}\natexlab{b}.
\newblock \showarticletitle{WebDancer: Towards Autonomous Information Seeking Agency}.
\newblock \bibinfo{journal}{\emph{CoRR}}  \bibinfo{volume}{abs/2505.22648} (\bibinfo{year}{2025}).
\newblock
\urldef\tempurl%
\url{https://doi.org/10.48550/ARXIV.2505.22648}
\showDOI{\tempurl}
\showeprint[arXiv]{2505.22648}


\bibitem[\protect\citeauthoryear{Wu, Yin, Jiang, Wang, Xi, Fang, Zhang, He, Zhou, Xie, and Huang}{Wu et~al\mbox{.}}{2025d}]%
        {2501_WebWalker}
\bibfield{author}{\bibinfo{person}{Jialong Wu}, \bibinfo{person}{Wenbiao Yin}, \bibinfo{person}{Yong Jiang}, \bibinfo{person}{Zhenglin Wang}, \bibinfo{person}{Zekun Xi}, \bibinfo{person}{Runnan Fang}, \bibinfo{person}{Linhai Zhang}, \bibinfo{person}{Yulan He}, \bibinfo{person}{Deyu Zhou}, \bibinfo{person}{Pengjun Xie}, {and} \bibinfo{person}{Fei Huang}.} \bibinfo{year}{2025}\natexlab{d}.
\newblock \showarticletitle{WebWalker: Benchmarking LLMs in Web Traversal}.
\newblock \bibinfo{journal}{\emph{CoRR}}  \bibinfo{volume}{abs/2501.07572} (\bibinfo{year}{2025}).
\newblock
\urldef\tempurl%
\url{https://doi.org/10.48550/ARXIV.2501.07572}
\showDOI{\tempurl}
\showeprint[arXiv]{2501.07572}


\bibitem[\protect\citeauthoryear{Wu, Yin, Jiang, Wang, Xi, Fang, Zhou, Xie, and Huang}{Wu et~al\mbox{.}}{2025e}]%
        {wu2025webwalker}
\bibfield{author}{\bibinfo{person}{Jialong Wu}, \bibinfo{person}{Wenbiao Yin}, \bibinfo{person}{Yong Jiang}, \bibinfo{person}{Zhenglin Wang}, \bibinfo{person}{Zekun Xi}, \bibinfo{person}{Runnan Fang}, \bibinfo{person}{Deyu Zhou}, \bibinfo{person}{Pengjun Xie}, {and} \bibinfo{person}{Fei Huang}.} \bibinfo{year}{2025}\natexlab{e}.
\newblock \bibinfo{title}{WebWalker: Benchmarking LLMs in Web Traversal}.
\newblock
\newblock
\showeprint[arxiv]{cs.CL/2501.07572}
\urldef\tempurl%
\url{https://arxiv.org/abs/2501.07572}
\showURL{%
\tempurl}


\bibitem[\protect\citeauthoryear{Wu, Li, Zhao, Zhang, Ou, Yin, Zhang, Jiang, Xie, Huang, Cheng, Wang, Cheng, and Zhou}{Wu et~al\mbox{.}}{2025c}]%
        {wu2025resumun}
\bibfield{author}{\bibinfo{person}{Xixi Wu}, \bibinfo{person}{Kuan Li}, \bibinfo{person}{Yida Zhao}, \bibinfo{person}{Liwen Zhang}, \bibinfo{person}{Litu Ou}, \bibinfo{person}{Huifeng Yin}, \bibinfo{person}{Zhongwang Zhang}, \bibinfo{person}{Yong Jiang}, \bibinfo{person}{Pengjun Xie}, \bibinfo{person}{Fei Huang}, \bibinfo{person}{Minhao Cheng}, \bibinfo{person}{Shuai Wang}, \bibinfo{person}{Hong Cheng}, {and} \bibinfo{person}{Jingren Zhou}.} \bibinfo{year}{2025}\natexlab{c}.
\newblock \showarticletitle{ReSum: Unlocking Long-Horizon Search Intelligence via Context Summarization}.
\newblock \bibinfo{journal}{\emph{arXiv preprint arXiv:2509.13313}} (\bibinfo{year}{2025}).
\newblock


\bibitem[\protect\citeauthoryear{Xiao, Jiang, Sun, Li, Lin, Zhuang, Zeng, Xia, Hua, Li, et~al\mbox{.}}{Xiao et~al\mbox{.}}{2025}]%
        {xiao2025limi}
\bibfield{author}{\bibinfo{person}{Yang Xiao}, \bibinfo{person}{Mohan Jiang}, \bibinfo{person}{Jie Sun}, \bibinfo{person}{Keyu Li}, \bibinfo{person}{Jifan Lin}, \bibinfo{person}{Yumin Zhuang}, \bibinfo{person}{Ji Zeng}, \bibinfo{person}{Shijie Xia}, \bibinfo{person}{Qishuo Hua}, \bibinfo{person}{Xuefeng Li}, {et~al\mbox{.}}} \bibinfo{year}{2025}\natexlab{}.
\newblock \showarticletitle{LIMI: Less is More for Agency}.
\newblock \bibinfo{journal}{\emph{arXiv preprint arXiv:2509.17567}} (\bibinfo{year}{2025}).
\newblock


\bibitem[\protect\citeauthoryear{Xu, Jin, Hao, Song, Sun, and Yuan}{Xu et~al\mbox{.}}{2024}]%
        {2411_llava_o1}
\bibfield{author}{\bibinfo{person}{Guowei Xu}, \bibinfo{person}{Peng Jin}, \bibinfo{person}{Li Hao}, \bibinfo{person}{Yibing Song}, \bibinfo{person}{Lichao Sun}, {and} \bibinfo{person}{Li Yuan}.} \bibinfo{year}{2024}\natexlab{}.
\newblock \showarticletitle{LLaVA-o1: Let Vision Language Models Reason Step-by-Step}.
\newblock \bibinfo{journal}{\emph{arXiv preprint arXiv:2411.10440}} (\bibinfo{year}{2024}).
\newblock


\bibitem[\protect\citeauthoryear{Xue, Zheng, Liu, Li, Zheng, Ma, and An}{Xue et~al\mbox{.}}{2025}]%
        {xue2025simpletir}
\bibfield{author}{\bibinfo{person}{Zhenghai Xue}, \bibinfo{person}{Longtao Zheng}, \bibinfo{person}{Qian Liu}, \bibinfo{person}{Yingru Li}, \bibinfo{person}{Xiaosen Zheng}, \bibinfo{person}{Zejun Ma}, {and} \bibinfo{person}{Bo An}.} \bibinfo{year}{2025}\natexlab{}.
\newblock \showarticletitle{SimpleTIR: End-to-End Reinforcement Learning for Multi-Turn Tool-Integrated Reasoning}.
\newblock \bibinfo{journal}{\emph{arXiv preprint arXiv:2509.02479}} (\bibinfo{year}{2025}).
\newblock


\bibitem[\protect\citeauthoryear{Yang, Li, Yang, Zhang, Hui, Zheng, Yu, Gao, Huang, Lv, Zheng, Liu, Zhou, Huang, Hu, Ge, Wei, Lin, Tang, Yang, Tu, Zhang, Yang, Yang, Zhou, Zhou, Lin, Dang, Bao, Yang, Yu, Deng, Li, Xue, Li, Zhang, Wang, Zhu, Men, Gao, Liu, Luo, Li, Tang, Yin, Ren, Wang, Zhang, Ren, Fan, Su, Zhang, Zhang, Wan, Liu, Wang, Cui, Zhang, Zhou, and Qiu}{Yang et~al\mbox{.}}{2025b}]%
        {qwen3}
\bibfield{author}{\bibinfo{person}{An Yang}, \bibinfo{person}{Anfeng Li}, \bibinfo{person}{Baosong Yang}, \bibinfo{person}{Beichen Zhang}, \bibinfo{person}{Binyuan Hui}, \bibinfo{person}{Bo Zheng}, \bibinfo{person}{Bowen Yu}, \bibinfo{person}{Chang Gao}, \bibinfo{person}{Chengen Huang}, \bibinfo{person}{Chenxu Lv}, \bibinfo{person}{Chujie Zheng}, \bibinfo{person}{Dayiheng Liu}, \bibinfo{person}{Fan Zhou}, \bibinfo{person}{Fei Huang}, \bibinfo{person}{Feng Hu}, \bibinfo{person}{Hao Ge}, \bibinfo{person}{Haoran Wei}, \bibinfo{person}{Huan Lin}, \bibinfo{person}{Jialong Tang}, \bibinfo{person}{Jian Yang}, \bibinfo{person}{Jianhong Tu}, \bibinfo{person}{Jianwei Zhang}, \bibinfo{person}{Jian Yang}, \bibinfo{person}{Jiaxi Yang}, \bibinfo{person}{Jingren Zhou}, \bibinfo{person}{Jingren Zhou}, \bibinfo{person}{Junyang Lin}, \bibinfo{person}{Kai Dang}, \bibinfo{person}{Keqin Bao}, \bibinfo{person}{Kexin Yang}, \bibinfo{person}{Le Yu}, \bibinfo{person}{Lianghao Deng}, \bibinfo{person}{Mei Li}, \bibinfo{person}{Mingfeng
  Xue}, \bibinfo{person}{Mingze Li}, \bibinfo{person}{Pei Zhang}, \bibinfo{person}{Peng Wang}, \bibinfo{person}{Qin Zhu}, \bibinfo{person}{Rui Men}, \bibinfo{person}{Ruize Gao}, \bibinfo{person}{Shixuan Liu}, \bibinfo{person}{Shuang Luo}, \bibinfo{person}{Tianhao Li}, \bibinfo{person}{Tianyi Tang}, \bibinfo{person}{Wenbiao Yin}, \bibinfo{person}{Xingzhang Ren}, \bibinfo{person}{Xinyu Wang}, \bibinfo{person}{Xinyu Zhang}, \bibinfo{person}{Xuancheng Ren}, \bibinfo{person}{Yang Fan}, \bibinfo{person}{Yang Su}, \bibinfo{person}{Yichang Zhang}, \bibinfo{person}{Yinger Zhang}, \bibinfo{person}{Yu Wan}, \bibinfo{person}{Yuqiong Liu}, \bibinfo{person}{Zekun Wang}, \bibinfo{person}{Zeyu Cui}, \bibinfo{person}{Zhenru Zhang}, \bibinfo{person}{Zhipeng Zhou}, {and} \bibinfo{person}{Zihan Qiu}.} \bibinfo{year}{2025}\natexlab{b}.
\newblock \showarticletitle{Qwen3 Technical Report}.
\newblock \bibinfo{journal}{\emph{CoRR}}  \bibinfo{volume}{abs/2505.09388} (\bibinfo{year}{2025}).
\newblock
\urldef\tempurl%
\url{https://doi.org/10.48550/ARXIV.2505.09388}
\showDOI{\tempurl}
\showeprint[arXiv]{2505.09388}


\bibitem[\protect\citeauthoryear{Yang, Guo, Huang, Liang, Wang, and Tang}{Yang et~al\mbox{.}}{2025a}]%
        {yang2025treerpo}
\bibfield{author}{\bibinfo{person}{Zhicheng Yang}, \bibinfo{person}{Zhijiang Guo}, \bibinfo{person}{Yinya Huang}, \bibinfo{person}{Xiaodan Liang}, \bibinfo{person}{Yiwei Wang}, {and} \bibinfo{person}{Jing Tang}.} \bibinfo{year}{2025}\natexlab{a}.
\newblock \showarticletitle{TreeRPO: Tree Relative Policy Optimization}.
\newblock \bibinfo{journal}{\emph{arXiv preprint arXiv:2506.05183}} (\bibinfo{year}{2025}).
\newblock


\bibitem[\protect\citeauthoryear{Yao, Zhao, Yu, Du, Shafran, Narasimhan, and Cao}{Yao et~al\mbox{.}}{2022}]%
        {yao2022react}
\bibfield{author}{\bibinfo{person}{Shunyu Yao}, \bibinfo{person}{Jeffrey Zhao}, \bibinfo{person}{Dian Yu}, \bibinfo{person}{Nan Du}, \bibinfo{person}{Izhak Shafran}, \bibinfo{person}{Karthik Narasimhan}, {and} \bibinfo{person}{Yuan Cao}.} \bibinfo{year}{2022}\natexlab{}.
\newblock \showarticletitle{React: Synergizing reasoning and acting in language models}.
\newblock \bibinfo{journal}{\emph{arXiv preprint arXiv:2210.03629}} (\bibinfo{year}{2022}).
\newblock


\bibitem[\protect\citeauthoryear{Yu, Peng, Tian, Song, Mi, and Yu}{Yu et~al\mbox{.}}{2024}]%
        {siam}
\bibfield{author}{\bibinfo{person}{Dian Yu}, \bibinfo{person}{Baolin Peng}, \bibinfo{person}{Ye Tian}, \bibinfo{person}{Linfeng Song}, \bibinfo{person}{Haitao Mi}, {and} \bibinfo{person}{Dong Yu}.} \bibinfo{year}{2024}\natexlab{}.
\newblock \showarticletitle{SIaM: Self-Improving Code-Assisted Mathematical Reasoning of Large Language Models}.
\newblock \bibinfo{journal}{\emph{CoRR}}  \bibinfo{volume}{abs/2408.15565} (\bibinfo{year}{2024}).
\newblock
\urldef\tempurl%
\url{https://doi.org/10.48550/ARXIV.2408.15565}
\showDOI{\tempurl}
\showeprint[arXiv]{2408.15565}


\bibitem[\protect\citeauthoryear{Yu, Chen, Feng, Chen, Dai, Yu, Zhang, Ma, Liu, Wang, and Zhou}{Yu et~al\mbox{.}}{2025a}]%
        {MemAgent}
\bibfield{author}{\bibinfo{person}{Hongli Yu}, \bibinfo{person}{Tinghong Chen}, \bibinfo{person}{Jiangtao Feng}, \bibinfo{person}{Jiangjie Chen}, \bibinfo{person}{Weinan Dai}, \bibinfo{person}{Qiying Yu}, \bibinfo{person}{Ya{-}Qin Zhang}, \bibinfo{person}{Wei{-}Ying Ma}, \bibinfo{person}{Jingjing Liu}, \bibinfo{person}{Mingxuan Wang}, {and} \bibinfo{person}{Hao Zhou}.} \bibinfo{year}{2025}\natexlab{a}.
\newblock \showarticletitle{MemAgent: Reshaping Long-Context {LLM} with Multi-Conv RL-based Memory Agent}.
\newblock \bibinfo{journal}{\emph{CoRR}}  \bibinfo{volume}{abs/2507.02259} (\bibinfo{year}{2025}).
\newblock
\urldef\tempurl%
\url{https://doi.org/10.48550/ARXIV.2507.02259}
\showDOI{\tempurl}
\showeprint[arXiv]{2507.02259}


\bibitem[\protect\citeauthoryear{Yu, Zhang, Zhu, Yuan, Zuo, Yue, Fan, Liu, Liu, Liu, Lin, Lin, Ma, Sheng, Tong, Zhang, Zhang, Zhang, Zhu, Zhu, Chen, Chen, Wang, Yu, Dai, Song, Wei, Zhou, Liu, Ma, Zhang, Yan, Qiao, Wu, and Wang}{Yu et~al\mbox{.}}{2025b}]%
        {DAPO}
\bibfield{author}{\bibinfo{person}{Qiying Yu}, \bibinfo{person}{Zheng Zhang}, \bibinfo{person}{Ruofei Zhu}, \bibinfo{person}{Yufeng Yuan}, \bibinfo{person}{Xiaochen Zuo}, \bibinfo{person}{Yu Yue}, \bibinfo{person}{Tiantian Fan}, \bibinfo{person}{Gaohong Liu}, \bibinfo{person}{Lingjun Liu}, \bibinfo{person}{Xin Liu}, \bibinfo{person}{Haibin Lin}, \bibinfo{person}{Zhiqi Lin}, \bibinfo{person}{Bole Ma}, \bibinfo{person}{Guangming Sheng}, \bibinfo{person}{Yuxuan Tong}, \bibinfo{person}{Chi Zhang}, \bibinfo{person}{Mofan Zhang}, \bibinfo{person}{Wang Zhang}, \bibinfo{person}{Hang Zhu}, \bibinfo{person}{Jinhua Zhu}, \bibinfo{person}{Jiaze Chen}, \bibinfo{person}{Jiangjie Chen}, \bibinfo{person}{Chengyi Wang}, \bibinfo{person}{Hongli Yu}, \bibinfo{person}{Weinan Dai}, \bibinfo{person}{Yuxuan Song}, \bibinfo{person}{Xiangpeng Wei}, \bibinfo{person}{Hao Zhou}, \bibinfo{person}{Jingjing Liu}, \bibinfo{person}{Wei{-}Ying Ma}, \bibinfo{person}{Ya{-}Qin Zhang}, \bibinfo{person}{Lin Yan}, \bibinfo{person}{Mu Qiao},
  \bibinfo{person}{Yonghui Wu}, {and} \bibinfo{person}{Mingxuan Wang}.} \bibinfo{year}{2025}\natexlab{b}.
\newblock \showarticletitle{{DAPO:} An Open-Source {LLM} Reinforcement Learning System at Scale}.
\newblock \bibinfo{journal}{\emph{CoRR}}  \bibinfo{volume}{abs/2503.14476} (\bibinfo{year}{2025}).
\newblock
\urldef\tempurl%
\url{https://doi.org/10.48550/ARXIV.2503.14476}
\showDOI{\tempurl}
\showeprint[arXiv]{2503.14476}


\bibitem[\protect\citeauthoryear{Zeng, Lv, Zheng, Hou, Chen, Xie, Wang, Yin, Zeng, Zhang, Wang, Zhong, Liu, Lu, Cao, Zhang, Huang, Wei, Cheng, An, Niu, Wen, Bai, Du, Wang, Zhu, Zhang, Wen, Wu, Xu, Huang, Zhao, Cai, Yu, Li, Ge, Huang, Zhang, Xu, Zhu, Li, Yin, Lin, Yang, Jiang, Ai, Zhu, Wang, Pan, Wang, Sun, Li, Li, Hu, Zhang, Peng, Tai, Zhang, Wang, Yang, Liu, Zhao, Liu, Yan, Liu, Chen, Li, Zhao, Ren, Jiao, Zhao, Yan, Wang, Gui, Zhao, Liu, Li, Li, Lu, Wang, Yuan, Li, Du, Du, Liu, Zhi, Gao, Wang, Yang, Xu, Fan, Wu, Ding, Wang, Zhang, Li, Xu, Zhao, and Zhai}{Zeng et~al\mbox{.}}{2025}]%
        {GLM-4.5}
\bibfield{author}{\bibinfo{person}{Aohan Zeng}, \bibinfo{person}{Xin Lv}, \bibinfo{person}{Qinkai Zheng}, \bibinfo{person}{Zhenyu Hou}, \bibinfo{person}{Bin Chen}, \bibinfo{person}{Chengxing Xie}, \bibinfo{person}{Cunxiang Wang}, \bibinfo{person}{Da Yin}, \bibinfo{person}{Hao Zeng}, \bibinfo{person}{Jiajie Zhang}, \bibinfo{person}{Kedong Wang}, \bibinfo{person}{Lucen Zhong}, \bibinfo{person}{Mingdao Liu}, \bibinfo{person}{Rui Lu}, \bibinfo{person}{Shulin Cao}, \bibinfo{person}{Xiaohan Zhang}, \bibinfo{person}{Xuancheng Huang}, \bibinfo{person}{Yao Wei}, \bibinfo{person}{Yean Cheng}, \bibinfo{person}{Yifan An}, \bibinfo{person}{Yilin Niu}, \bibinfo{person}{Yuanhao Wen}, \bibinfo{person}{Yushi Bai}, \bibinfo{person}{Zhengxiao Du}, \bibinfo{person}{Zihan Wang}, \bibinfo{person}{Zilin Zhu}, \bibinfo{person}{Bohan Zhang}, \bibinfo{person}{Bosi Wen}, \bibinfo{person}{Bowen Wu}, \bibinfo{person}{Bowen Xu}, \bibinfo{person}{Can Huang}, \bibinfo{person}{Casey Zhao}, \bibinfo{person}{Changpeng Cai},
  \bibinfo{person}{Chao Yu}, \bibinfo{person}{Chen Li}, \bibinfo{person}{Chendi Ge}, \bibinfo{person}{Chenghua Huang}, \bibinfo{person}{Chenhui Zhang}, \bibinfo{person}{Chenxi Xu}, \bibinfo{person}{Chenzheng Zhu}, \bibinfo{person}{Chuang Li}, \bibinfo{person}{Congfeng Yin}, \bibinfo{person}{Daoyan Lin}, \bibinfo{person}{Dayong Yang}, \bibinfo{person}{Dazhi Jiang}, \bibinfo{person}{Ding Ai}, \bibinfo{person}{Erle Zhu}, \bibinfo{person}{Fei Wang}, \bibinfo{person}{Gengzheng Pan}, \bibinfo{person}{Guo Wang}, \bibinfo{person}{Hailong Sun}, \bibinfo{person}{Haitao Li}, \bibinfo{person}{Haiyang Li}, \bibinfo{person}{Haiyi Hu}, \bibinfo{person}{Hanyu Zhang}, \bibinfo{person}{Hao Peng}, \bibinfo{person}{Hao Tai}, \bibinfo{person}{Haoke Zhang}, \bibinfo{person}{Haoran Wang}, \bibinfo{person}{Haoyu Yang}, \bibinfo{person}{He Liu}, \bibinfo{person}{He Zhao}, \bibinfo{person}{Hongwei Liu}, \bibinfo{person}{Hongxi Yan}, \bibinfo{person}{Huan Liu}, \bibinfo{person}{Huilong Chen}, \bibinfo{person}{Ji Li},
  \bibinfo{person}{Jiajing Zhao}, \bibinfo{person}{Jiamin Ren}, \bibinfo{person}{Jian Jiao}, \bibinfo{person}{Jiani Zhao}, \bibinfo{person}{Jianyang Yan}, \bibinfo{person}{Jiaqi Wang}, \bibinfo{person}{Jiayi Gui}, \bibinfo{person}{Jiayue Zhao}, \bibinfo{person}{Jie Liu}, \bibinfo{person}{Jijie Li}, \bibinfo{person}{Jing Li}, \bibinfo{person}{Jing Lu}, \bibinfo{person}{Jingsen Wang}, \bibinfo{person}{Jingwei Yuan}, \bibinfo{person}{Jingxuan Li}, \bibinfo{person}{Jingzhao Du}, \bibinfo{person}{Jinhua Du}, \bibinfo{person}{Jinxin Liu}, \bibinfo{person}{Junkai Zhi}, \bibinfo{person}{Junli Gao}, \bibinfo{person}{Ke Wang}, \bibinfo{person}{Lekang Yang}, \bibinfo{person}{Liang Xu}, \bibinfo{person}{Lin Fan}, \bibinfo{person}{Lindong Wu}, \bibinfo{person}{Lintao Ding}, \bibinfo{person}{Lu Wang}, \bibinfo{person}{Man Zhang}, \bibinfo{person}{Minghao Li}, \bibinfo{person}{Minghuan Xu}, \bibinfo{person}{Mingming Zhao}, {and} \bibinfo{person}{Mingshu Zhai}.} \bibinfo{year}{2025}\natexlab{}.
\newblock \showarticletitle{{GLM-4.5:} Agentic, Reasoning, and Coding {(ARC)} Foundation Models}.
\newblock \bibinfo{journal}{\emph{CoRR}}  \bibinfo{volume}{abs/2508.06471} (\bibinfo{year}{2025}).
\newblock
\urldef\tempurl%
\url{https://doi.org/10.48550/ARXIV.2508.06471}
\showDOI{\tempurl}
\showeprint[arXiv]{2508.06471}


\bibitem[\protect\citeauthoryear{Zhai, Bai, Lin, Pan, Tong, Zhou, Suhr, Xie, LeCun, Ma, and Levine}{Zhai et~al\mbox{.}}{2024}]%
        {agentrl_3}
\bibfield{author}{\bibinfo{person}{Simon Zhai}, \bibinfo{person}{Hao Bai}, \bibinfo{person}{Zipeng Lin}, \bibinfo{person}{Jiayi Pan}, \bibinfo{person}{Peter Tong}, \bibinfo{person}{Yifei Zhou}, \bibinfo{person}{Alane Suhr}, \bibinfo{person}{Saining Xie}, \bibinfo{person}{Yann LeCun}, \bibinfo{person}{Yi Ma}, {and} \bibinfo{person}{Sergey Levine}.} \bibinfo{year}{2024}\natexlab{}.
\newblock \showarticletitle{Fine-Tuning Large Vision-Language Models as Decision-Making Agents via Reinforcement Learning}. In \bibinfo{booktitle}{\emph{Advances in Neural Information Processing Systems 38: Annual Conference on Neural Information Processing Systems 2024, NeurIPS 2024, Vancouver, BC, Canada, December 10 - 15, 2024}}, \bibfield{editor}{\bibinfo{person}{Amir Globersons}, \bibinfo{person}{Lester Mackey}, \bibinfo{person}{Danielle Belgrave}, \bibinfo{person}{Angela Fan}, \bibinfo{person}{Ulrich Paquet}, \bibinfo{person}{Jakub~M. Tomczak}, {and} \bibinfo{person}{Cheng Zhang}} (Eds.).
\newblock
\urldef\tempurl%
\url{http://papers.nips.cc/paper\_files/paper/2024/hash/c848b7d3adc08fcd0bf1df3101ba6728-Abstract-Conference.html}
\showURL{%
\tempurl}


\bibitem[\protect\citeauthoryear{Zhang, Zhao, Wu, Li, Yin, Zhang, Jiang, Li, Tu, Xie, and Huang}{Zhang et~al\mbox{.}}{2025b}]%
        {EvolveSearch}
\bibfield{author}{\bibinfo{person}{Dingchu Zhang}, \bibinfo{person}{Yida Zhao}, \bibinfo{person}{Jialong Wu}, \bibinfo{person}{Baixuan Li}, \bibinfo{person}{Wenbiao Yin}, \bibinfo{person}{Liwen Zhang}, \bibinfo{person}{Yong Jiang}, \bibinfo{person}{Yufeng Li}, \bibinfo{person}{Kewei Tu}, \bibinfo{person}{Pengjun Xie}, {and} \bibinfo{person}{Fei Huang}.} \bibinfo{year}{2025}\natexlab{b}.
\newblock \showarticletitle{EvolveSearch: An Iterative Self-Evolving Search Agent}.
\newblock \bibinfo{journal}{\emph{CoRR}}  \bibinfo{volume}{abs/2505.22501} (\bibinfo{year}{2025}).
\newblock
\urldef\tempurl%
\url{https://doi.org/10.48550/ARXIV.2505.22501}
\showDOI{\tempurl}
\showeprint[arXiv]{2505.22501}


\bibitem[\protect\citeauthoryear{Zhang, Geng, Yu, Yin, Zhang, Tan, Zhou, Li, Xue, Li, et~al\mbox{.}}{Zhang et~al\mbox{.}}{2025a}]%
        {zhang2025landscape}
\bibfield{author}{\bibinfo{person}{Guibin Zhang}, \bibinfo{person}{Hejia Geng}, \bibinfo{person}{Xiaohang Yu}, \bibinfo{person}{Zhenfei Yin}, \bibinfo{person}{Zaibin Zhang}, \bibinfo{person}{Zelin Tan}, \bibinfo{person}{Heng Zhou}, \bibinfo{person}{Zhongzhi Li}, \bibinfo{person}{Xiangyuan Xue}, \bibinfo{person}{Yijiang Li}, {et~al\mbox{.}}} \bibinfo{year}{2025}\natexlab{a}.
\newblock \showarticletitle{The Landscape of Agentic Reinforcement Learning for LLMs: A Survey}.
\newblock \bibinfo{journal}{\emph{arXiv preprint arXiv:2509.02547}} (\bibinfo{year}{2025}).
\newblock


\bibitem[\protect\citeauthoryear{Zhang, Li, Cui, Cai, Liu, Fu, Huang, Zhao, Zhang, Chen, Wang, Luu, Bi, Shi, and Shi}{Zhang et~al\mbox{.}}{2023}]%
        {zhang2023sirens}
\bibfield{author}{\bibinfo{person}{Yue Zhang}, \bibinfo{person}{Yafu Li}, \bibinfo{person}{Leyang Cui}, \bibinfo{person}{Deng Cai}, \bibinfo{person}{Lemao Liu}, \bibinfo{person}{Tingchen Fu}, \bibinfo{person}{Xinting Huang}, \bibinfo{person}{Enbo Zhao}, \bibinfo{person}{Yu Zhang}, \bibinfo{person}{Yulong Chen}, \bibinfo{person}{Longyue Wang}, \bibinfo{person}{Anh~Tuan Luu}, \bibinfo{person}{Wei Bi}, \bibinfo{person}{Freda Shi}, {and} \bibinfo{person}{Shuming Shi}.} \bibinfo{year}{2023}\natexlab{}.
\newblock \showarticletitle{Siren's Song in the {AI} Ocean: {A} Survey on Hallucination in Large Language Models}.
\newblock \bibinfo{journal}{\emph{CoRR}}  \bibinfo{volume}{abs/2309.01219} (\bibinfo{year}{2023}).
\newblock
\urldef\tempurl%
\url{https://doi.org/10.48550/ARXIV.2309.01219}
\showDOI{\tempurl}
\showeprint[arXiv]{2309.01219}


\bibitem[\protect\citeauthoryear{Zheng, Liu, Li, Chen, Yu, Gao, Dang, Liu, Men, Yang, Zhou, and Lin}{Zheng et~al\mbox{.}}{2025a}]%
        {GSPO}
\bibfield{author}{\bibinfo{person}{Chujie Zheng}, \bibinfo{person}{Shixuan Liu}, \bibinfo{person}{Mingze Li}, \bibinfo{person}{Xiong{-}Hui Chen}, \bibinfo{person}{Bowen Yu}, \bibinfo{person}{Chang Gao}, \bibinfo{person}{Kai Dang}, \bibinfo{person}{Yuqiong Liu}, \bibinfo{person}{Rui Men}, \bibinfo{person}{An Yang}, \bibinfo{person}{Jingren Zhou}, {and} \bibinfo{person}{Junyang Lin}.} \bibinfo{year}{2025}\natexlab{a}.
\newblock \showarticletitle{Group Sequence Policy Optimization}.
\newblock \bibinfo{journal}{\emph{CoRR}}  \bibinfo{volume}{abs/2507.18071} (\bibinfo{year}{2025}).
\newblock
\urldef\tempurl%
\url{https://doi.org/10.48550/ARXIV.2507.18071}
\showDOI{\tempurl}
\showeprint[arXiv]{2507.18071}


\bibitem[\protect\citeauthoryear{Zheng, Xing, Gu, Liang, Qu, Zhou, Li, Wen, Lin, Huang, et~al\mbox{.}}{Zheng et~al\mbox{.}}{2025b}]%
        {zheng2025first}
\bibfield{author}{\bibinfo{person}{Tianyu Zheng}, \bibinfo{person}{Tianshun Xing}, \bibinfo{person}{Qingshui Gu}, \bibinfo{person}{Taoran Liang}, \bibinfo{person}{Xingwei Qu}, \bibinfo{person}{Xin Zhou}, \bibinfo{person}{Yizhi Li}, \bibinfo{person}{Zhoufutu Wen}, \bibinfo{person}{Chenghua Lin}, \bibinfo{person}{Wenhao Huang}, {et~al\mbox{.}}} \bibinfo{year}{2025}\natexlab{b}.
\newblock \showarticletitle{First Return, Entropy-Eliciting Explore}.
\newblock \bibinfo{journal}{\emph{arXiv preprint arXiv:2507.07017}} (\bibinfo{year}{2025}).
\newblock


\bibitem[\protect\citeauthoryear{Zheng, Xing, Gu, Liang, Qu, Zhou, Li, Wen, Lin, Huang, et~al\mbox{.}}{Zheng et~al\mbox{.}}{2025c}]%
        {zheng2025returnentropyelicitingexplore}
\bibfield{author}{\bibinfo{person}{Tianyu Zheng}, \bibinfo{person}{Tianshun Xing}, \bibinfo{person}{Qingshui Gu}, \bibinfo{person}{Taoran Liang}, \bibinfo{person}{Xingwei Qu}, \bibinfo{person}{Xin Zhou}, \bibinfo{person}{Yizhi Li}, \bibinfo{person}{Zhoufutu Wen}, \bibinfo{person}{Chenghua Lin}, \bibinfo{person}{Wenhao Huang}, {et~al\mbox{.}}} \bibinfo{year}{2025}\natexlab{c}.
\newblock \showarticletitle{First return, entropy-eliciting explore}.
\newblock \bibinfo{journal}{\emph{arXiv preprint arXiv:2507.07017}} (\bibinfo{year}{2025}).
\newblock


\bibitem[\protect\citeauthoryear{Zhou, Chen, Guo, Yan, Lee, Wang, Lee, Zhang, Shao, Yang, et~al\mbox{.}}{Zhou et~al\mbox{.}}{2025a}]%
        {zhou2025agentfly}
\bibfield{author}{\bibinfo{person}{Huichi Zhou}, \bibinfo{person}{Yihang Chen}, \bibinfo{person}{Siyuan Guo}, \bibinfo{person}{Xue Yan}, \bibinfo{person}{Kin~Hei Lee}, \bibinfo{person}{Zihan Wang}, \bibinfo{person}{Ka~Yiu Lee}, \bibinfo{person}{Guchun Zhang}, \bibinfo{person}{Kun Shao}, \bibinfo{person}{Linyi Yang}, {et~al\mbox{.}}} \bibinfo{year}{2025}\natexlab{a}.
\newblock \showarticletitle{Agentfly: Fine-tuning llm agents without fine-tuning llms}.
\newblock \bibinfo{journal}{\emph{arXiv preprint arXiv:2508.16153}} (\bibinfo{year}{2025}).
\newblock


\bibitem[\protect\citeauthoryear{Zhou, Jiang, Zhu, Li, Guo, Chen, and Zhang}{Zhou et~al\mbox{.}}{2025b}]%
        {Fin-PRM}
\bibfield{author}{\bibinfo{person}{Yuanchen Zhou}, \bibinfo{person}{Shuo Jiang}, \bibinfo{person}{Jie Zhu}, \bibinfo{person}{Junhui Li}, \bibinfo{person}{Lifan Guo}, \bibinfo{person}{Feng Chen}, {and} \bibinfo{person}{Chi Zhang}.} \bibinfo{year}{2025}\natexlab{b}.
\newblock \showarticletitle{Fin-PRM: {A} Domain-Specialized Process Reward Model for Financial Reasoning in Large Language Models}.
\newblock \bibinfo{journal}{\emph{CoRR}}  \bibinfo{volume}{abs/2508.15202} (\bibinfo{year}{2025}).
\newblock
\urldef\tempurl%
\url{https://doi.org/10.48550/ARXIV.2508.15202}
\showDOI{\tempurl}
\showeprint[arXiv]{2508.15202}


\bibitem[\protect\citeauthoryear{Zhou, Zanette, Pan, Levine, and Kumar}{Zhou et~al\mbox{.}}{2024}]%
        {zhou2024archer}
\bibfield{author}{\bibinfo{person}{Yifei Zhou}, \bibinfo{person}{Andrea Zanette}, \bibinfo{person}{Jiayi Pan}, \bibinfo{person}{Sergey Levine}, {and} \bibinfo{person}{Aviral Kumar}.} \bibinfo{year}{2024}\natexlab{}.
\newblock \showarticletitle{Archer: Training language model agents via hierarchical multi-turn rl}.
\newblock \bibinfo{journal}{\emph{arXiv preprint arXiv:2402.19446}} (\bibinfo{year}{2024}).
\newblock


\bibitem[\protect\citeauthoryear{Zhu, Feng, Du, Gu, Yu, Wang, Chen, Chu, Chen, and Qin}{Zhu et~al\mbox{.}}{2024}]%
        {zhu2024information}
\bibfield{author}{\bibinfo{person}{Kun Zhu}, \bibinfo{person}{Xiaocheng Feng}, \bibinfo{person}{Xiyuan Du}, \bibinfo{person}{Yuxuan Gu}, \bibinfo{person}{Weijiang Yu}, \bibinfo{person}{Haotian Wang}, \bibinfo{person}{Qianglong Chen}, \bibinfo{person}{Zheng Chu}, \bibinfo{person}{Jingchang Chen}, {and} \bibinfo{person}{Bing Qin}.} \bibinfo{year}{2024}\natexlab{}.
\newblock \showarticletitle{An information bottleneck perspective for effective noise filtering on retrieval-augmented generation}.
\newblock \bibinfo{journal}{\emph{arXiv preprint arXiv:2406.01549}} (\bibinfo{year}{2024}).
\newblock


\end{thebibliography}

\clearpage
\appendix

\begin{center}
{\Large \textbf{Appendix}}
\end{center}


\setcounter{section}{0}
\renewcommand{\thesection}{\Alph{section}}

\section{Proof of the Gradient of AEPO}
\small
\label{app:proof}

In this section, we will comprehensively detail the theoretical derivation of AEPO's forward propagation formulas and how they lead to the backward propagation formulas. Specifically:

We begin with the loss function:

\begin{equation}
\mathcal{L} = \mathbb{E}_{x \sim \mathcal{D}} \left[ \frac{1}{\sum_{j=1}^{G} T_{j}} \sum_{j=1}^{G}\sum_{t=1}^{T_{j}} \min \left( \delta \tilde{A}^{(t)}, \operatorname{clip} \left( \delta, 1-\epsilon_{l}, \frac{1+\epsilon_{h}}{\operatorname{sg}(\delta)}\delta \right) \tilde{A}^{(t)} \right) \right],
\end{equation}

where $\delta = r_t^{(j)}(\theta)$ represents the importance ratio, and $\operatorname{sg}(\cdot)$ is the stop-gradient operator. Given that $\tilde{A}^{(t)}$ is a constant and $\nabla_\theta \delta = \delta \, \phi_\theta(a_{j,t}, s_{j,t})$, the gradient of $f(\delta)$ can be expressed as:

\begin{equation}
\nabla_\theta f(\delta) = \tilde{A}^{(t)} \, s(\delta) \, \delta \, \phi_\theta(a_{j,t}, s_{j,t}),
\end{equation}

where $s(\delta)$ depends on the range of $\delta$. Therefore, we consider three scenarios:

\textbf{(1) If $\tilde{A}^{(t)}>0$ and $\delta>1+\epsilon_h$: }
The upper clipping boundary is active, so $\partial f/\partial \delta = (1+\epsilon_h)/\operatorname{sg}(\delta)$, effectively simplifying to $(1+\epsilon_h)$.

\textbf{(2) If $\tilde{A}^{(t)}<0$ and $\delta<1-\epsilon_\ell$: } 
The lower clipping boundary dominates, leading to $\partial f/\partial \delta = 0$, causing the gradient to vanish.

The region is unclipped, resulting in $\partial f/\partial \delta = \delta$.

By combining all cases, we derive:

\begin{equation}
\mathcal{F}_{j,t}(\theta)=
\begin{cases}
1+\epsilon_h, & \tilde{A}^{(t)}>0,~\delta>1+\epsilon_h,\\[3pt]
0, & \tilde{A}^{(t)}<0,~\delta<1-\epsilon_\ell,\\[3pt]
\delta, & \text{otherwise.}
\end{cases}
\label{eq:F_definition}
\end{equation}

Thus, the gradient update is given by:
\begin{equation}
\nabla_{\theta}\mathcal{L}
=
\mathbb{E}_{x \sim \mathcal{D}}\!\left[
\frac{1}{\sum_{j=1}^{G} T_{j}}
\sum_{j=1}^{G}\sum_{t=1}^{T_{j}}
\mathcal{F}_{j,t}(\theta)
\cdot \phi_{\theta}(a_{j,t}, s_{j,t})
\cdot \tilde{A}^{(t)}
\right].
\label{eq:gradient_update}
\end{equation}

\section{Discussion of the Gradient Forms in Clipping-optimized RL}
\label{app:discussion}
In this section, we discuss the gradient differences between AEPO and clipping-optimized RL algorithms to gain insight into the differences in their policy update stages~\citep{su2025gppo}.


\subsection{CISPO}
\begin{equation}
\mathcal{L} = 
\mathbb{E}_{x \sim \mathcal{D}}
\left[
\frac{1}{\sum_{j=1}^{G} T_{j}}
\sum_{j=1}^{G}\sum_{t=1}^{T_{j}}
\delta \, \tilde{A}^{(t)} \log \pi_{\theta}(a_{t}^{(j)} \mid s_{t}^{(j)})
\right].
\end{equation}

By expanding the gradient of the loss function, we obtain:
\begin{equation}
\nabla_{\theta}\mathcal{L} =
\mathbb{E}_{x \sim \mathcal{D}}
\left[
\frac{1}{\sum_{j=1}^{G} T_{j}}
\sum_{j=1}^{G}\sum_{t=1}^{T_{j}}
\mathcal{F}_{j,t}(\theta)\,
\phi_{\theta}(a_{j,t}, s_{j,t})\,
\tilde{A}^{(t)}
\right],
\end{equation}
where
\begin{equation}
\mathcal{F}_{j,t}(\theta) =
\begin{cases}
1 - \epsilon_{\ell}, & \tilde{A}^{(t)} < 0, ~\delta < 1 - \epsilon_{\ell}, \\[3pt]
1 + \epsilon_{h},    & \tilde{A}^{(t)} > 0, ~\delta > 1 + \epsilon_{h}, \\[3pt]
1 - \epsilon_{\ell}, & \tilde{A}^{(t)} > 0, ~\delta < 1 - \epsilon_{\ell}, \\[3pt]
1 + \epsilon_{h},    & \tilde{A}^{(t)} < 0, ~\delta > 1 + \epsilon_{h}, \\[3pt]
\delta,              & \text{otherwise.}
\end{cases}
\end{equation}
As shown in Eq. (14), AEPO modifies the CISPO objective by introducing an asymmetric clipping rule that deactivates gradient flow when both $\tilde{A}^{(t)}<0$ and $\delta<1-\epsilon_\ell$. In CISPO, the gradient factor remains $F_{t}(\theta)=1-\epsilon_\ell$ for this region, propagating a fixed penalty regardless of sample reliability. AEPO, instead, sets $F_{t}(\theta)=0$, effectively filtering out low-confidence negative advantages. This simple but principled change prevents unstable gradient signals from low-likelihood rollouts and reduces the variance introduced by symmetric updates. Consequently, AEPO achieves smoother optimization dynamics and more stable convergence, especially under high-entropy exploration regimes where CISPO often exhibits oscillatory behavior. Figure~\ref{fig:dynamic} provides experimental evidence for this discussion.

\subsection{GPPO}
\begin{equation}
\mathcal{L}^{\mathrm{GPPO}}(\theta)
= \mathbb{E}_{x \sim \mathcal{D}}
\left[
\frac{1}{\sum_{j=1}^{G} T_{j}}
\sum_{j=1}^{G}\sum_{t=1}^{T_{j}}
\ell^{(t)}
\right],
\end{equation}
where
\begin{equation}
\ell^{(t)} =
\begin{cases}
\beta_{1} \cdot \dfrac{1 - \epsilon_{\ell}}{\operatorname{sg}(\delta)} \, \delta \, \tilde{A}^{(t)},
& \tilde{A}^{(t)} < 0, ~\delta < 1 - \epsilon_{\ell}, \\[6pt]
\beta_{2} \cdot \dfrac{1 + \epsilon_{h}}{\operatorname{sg}(\delta)} \, \delta \, \tilde{A}^{(t)},
& \tilde{A}^{(t)} > 0, ~\delta > 1 + \epsilon_{h}, \\[6pt]
\delta \, \tilde{A}^{(t)}, &
\text{otherwise}.
\end{cases}
\end{equation}

By expanding its gradient, we have:
\begin{equation}
\nabla_{\theta}\mathcal{L}^{\mathrm{GPPO}} =
\mathbb{E}_{x \sim \mathcal{D}}
\left[
\frac{1}{\sum_{j=1}^{G} T_{j}}
\sum_{j=1}^{G}\sum_{t=1}^{T_{j}}
\mathcal{F}_{j,t}(\theta)\,
\phi_{\theta}(a_{j,t}, s_{j,t})\,
\tilde{A}^{(t)}
\right],
\end{equation}
where
\begin{equation}
\mathcal{F}_{j,t}(\theta) =
\begin{cases}
\beta_{1}(1 - \epsilon_{\ell}), & \tilde{A}^{(t)} < 0, ~\delta < 1 - \epsilon_{\ell}, \\[3pt]
\beta_{2}(1 + \epsilon_{h}),    & \tilde{A}^{(t)} > 0, ~\delta > 1 + \epsilon_{h}, \\[3pt]
\delta,                         & \text{otherwise.}
\end{cases}
\end{equation}
Compared with GPPO, which retains bounded (non-zero) gradients inside clipped regions via its 
$\beta$-scaled correction terms, AEPO enforces a stricter rule: \textbf{residual gradients in the region 
$\tilde{A}^{(t)} < 0,~\delta < 1 - \epsilon_{\ell}$ are discarded (i.e., $F_{t}(\theta) = 0$). 
Empirically, we find that agentic RL training is particularly sensitive to GPPO-style pessimistic suppression. }
Since our goal is to exploit high-entropy tokens with positive rewards, excessive pessimism can 
hamper effective credit assignment for these tokens. AEPO therefore removes residual negative updates 
while allowing high-entropy, positively rewarded tokens to fully contribute to the gradient, 
improving both stability and credit propagation in long-horizon agentic tasks.

\section{Baselines}
\label{app:baselines}

In this section, we provide a detailed overview of the baseline models involved in all experiments, as follows: 
\subsection{RL algorithms}
\textbf{(1) Classical RL Method:}
\begin{itemize}[leftmargin=1em]
\item \textbf{GRPO~\citep{deepseekmath}}
is a reinforcement learning algorithm for fine-tuning large language models via group-based policy optimization. It optimizes model behaviors by comparing responses within sampled groups and assigning relative rewards, enabling more stable and sample-efficient policy updates.
\item \textbf{Reinforce++~\citep{hu2025reinforce++}}
extends the classic policy-gradient algorithm by incorporating variance reduction and adaptive normalization techniques. It improves training stability and sample efficiency when fine-tuning language models with scalar rewards, while keeping the overall objective aligned with standard REINFORCE.
\end{itemize}

\noindent \textbf{(2) Clipping-optimized RL Method}
\begin{itemize}[leftmargin=1em]
\item \textbf{DAPO~\citep{DAPO}} decouples the clipping operation from the policy update to achieve more stable optimization, and introduces a dynamic sampling strategy that adaptively selects training examples to maintain effective gradient signals. These techniques together improve training efficiency and prevent performance degradation in long-horizon reasoning tasks.

\item \textbf{GPPO~\citep{klear}} extends the PPO framework by decoupling the clipping operation between the forward and backward passes. During optimization, the policy ratio is clipped in the forward computation to ensure bounded updates, while the original, unclipped ratio is used in the backward path to preserve complete gradient information.

\item \textbf{CISPO~\citep{minimax2025minimaxm1scalingtesttimecompute}} reformulates ratio clipping by applying the constraint to importance sampling weights instead of policy ratios. It bounds update magnitudes in expectation while preserving token-level gradient information through unclipped policy ratios.
\end{itemize}

\noindent  \textbf{(3) Agentic RL Method}
\begin{itemize}[leftmargin=1em]
\item \textbf{GIGPO~\citep{GIGPO}}
groups complete trajectories at episode level to compute macro‐relative advantages, and also retroactively groups actions sharing anchor states across trajectories at step level to compute micro-relative advantages. Both levels are combined without using a critic, preserving the critic-free nature while enabling per-step credit signals.

\item \textbf{ARPO~\citep{dong2025arpo}}
is an RL method tailored for multi-turn LLM agents. It introduces an entropy-based adaptive rollout scheme that increases sampling in steps with high uncertainty, and incorporates an advantage attribution mechanism to assign credit across branching tool-use interactions.
\end{itemize}

\begin{algorithm}
\caption{Agentic Entropy-Balanced Policy Optimization}
\label{algo:aepo}
\begin{algorithmic}[1]
\REQUIRE Reasoning model $\pi_\theta$, external tools $T$,
total rollout size $k$, entropy sensitivity $\beta$, branch penalty slope $\gamma$, clipping bounds $\epsilon_l,\epsilon_h$, entropy-aware weight $\alpha$
\STATE \textbf{Input:} Dataset $D$
\STATE Initialize reference model: $\pi_{\theta}^{\text{old}} \gets \pi_\theta$
\FOR{$i = 1$ to $\mathcal{C}$}
    \STATE Sample mini-batch $D_b \subset D$
    \STATE \hspace{\algorithmicindent}\textcolor{blue}{// Dynamic Entropy-Balanced Rollout}
    \FOR{each query $q \in D_b$}      
        \STATE Generate 1 complete trajectory $r$ to obtain $H_{\text{root}}$ and $H_{\text{tool}}^{\text{avg}}$
        \STATE Global rollout size $m \gets k\cdot\sigma\!\bigl(\beta(H_{\text{root}}-H_{\text{tool}}^{\text{avg}})\bigr)$
        \STATE Branch rollout size $b \gets k - m$

        \STATE Initialize rollout pool $\mathcal{P}\gets\varnothing$
        \STATE Consecutive-high-entropy counter $l\gets 0$
        \WHILE{$|\mathcal{P}|<m$}
            \STATE Sample trajectory $r$; add to $\mathcal{P}$
        \ENDWHILE
        \WHILE{$b>0$ \textbf{and} $\exists\, r_j \in \mathcal{P}$ not terminated}
            \STATE Select a trajectory $r\in\mathcal{P}$ at tool-call step $t$
            \STATE $\Delta H_t \gets \text{Normalize}(H_t - H_{\text{initial}})$
            \STATE Consecutive penalty $\hat P(l)\gets \gamma \cdot l$
            \STATE Branch probability $P_t\gets (\alpha+\beta\Delta H_t)(1-\hat P(l))$
            \IF{$P_t>\tau$}
                \STATE Branch $Z$ sub-trajectories; \;$b\gets b-Z$
            \ELSE
                \STATE $l\gets l+1$ if $\Delta H_t>0$
            \ENDIF
        \ENDWHILE
        \IF{$b>0$}
            \STATE Sample $b$ additional trajectories and add to $\mathcal{P}$
        \ENDIF
    \ENDFOR
    \STATE \hspace{\algorithmicindent}\textcolor{blue}{// Entropy-Balanced Policy Optimization}
    \FOR{$\text{step}=1$ to $\mathcal{S}$}
        \STATE Compute standard advantage $\hat A_{\text{Acc}}$ and entropy advantage $\hat A_{\Delta H}$ via Eq.~\eqref{eq:Advantage}
        \STATE Entropy-aware advantage $\hat A \gets \hat A_{\text{Acc}}\cdot(1+\hat A_{\Delta H})^\alpha$
        \FOR{each token $t$ in trajectory $j$}
            \STATE Importance ratio $\delta\gets \pi_\theta/\pi_{\theta_{\text{old}}}$
            \IF{$\delta>1+\epsilon_h$ \textbf{and} $\hat A>0$}
                \STATE Gradient scaler $\mathcal{F}_{j,t}\gets 1+\epsilon_h$
            \ELSIF{$\delta<1-\epsilon_l$ \textbf{and} $\hat A<0$}
                \STATE Gradient scaler $\mathcal{F}_{j,t}\gets 0$
            \ELSE
                \STATE Gradient scaler $\mathcal{F}_{j,t}\gets \delta$
            \ENDIF
        \ENDFOR
        \STATE Update parameters via Eq.~\ref{eq:gradient_update}
    \ENDFOR
\ENDFOR
\STATE \textbf{Output:} Fine-tuned model $\pi_\theta$
\end{algorithmic}
\end{algorithm}

\subsection{Web Search Agent}

\begin{itemize}[leftmargin=1em]
\item \textbf{RAG~\citep{rag_lewis}} (Retrieval-Augmented Generation) combines information retrieval with generative modeling to enhance the accuracy, reliability, and timeliness of outputs. It retrieves relevant information from an external knowledge base before generating responses, addressing internal knowledge gaps and reducing hallucinations.

\item \textbf{Search-o1~\citep{searcho1}} is a framework designed to enhance reasoning by integrating agentic RAG mechanisms with a Reason-in-Documents module. It improves accuracy, coherence, and reliability in reasoning tasks, outperforming native reasoning and traditional RAG methods in complex scenarios.

\item \textbf{WebThinker~\citep{webthinker}} is an open-source framework developed by Renmin University of China, enabling LRMs to autonomously search, explore web pages, and generate research reports. It employs direct preference optimization and iterative synthesis tools to enhance tool utilization capabilities.

\item \textbf{ReAct~\citep{yao2022react}} combines reasoning and action to tackle complex tasks effectively. It allows models to generate reasoning steps and use external tools, such as search engines and databases, during decision-making, optimizing results through iterative processes.
\end{itemize}

\section{The Overall Algorithm Workflow of AEPO}

In this section, we delve into the overall workflow of the Agentic Entropy-Balanced Policy Optimization (AEPO) algorithm, as depicted in Algorithm Diagram \ref{algo:aepo}. The AEPO algorithm integrates dynamic entropy-balanced rollouts with entropy-balanced policy optimization to enhance multi-turn tool-use capabilities in large language models.

\end{document}